\documentclass{article}
\usepackage{authblk}
\usepackage{graphicx}
\usepackage{fullname}
\usepackage{amssymb}
\usepackage{lingmacros}
\usepackage{alltt}
\usepackage{rotating}
\usepackage{verbatim}
\usepackage{float}
\usepackage{url}
\usepackage{color}
\usepackage{array}
\newcolumntype{M}[1]{>{\centering\arraybackslash}m{#1}}
\newcolumntype{N}{@{}m{0pt}@{}}
\usepackage[cmyk]{xcolor}

\usepackage{todonotes}

\usepackage{bussproofs}
\usepackage{boxedminipage}
\usepackage{multicol}
\usepackage{paralist}
\usepackage{amsmath}

\title{Handling irresolvable conflicts in the Semantic Web:\\
an RDF-based conflict-tolerant version of\\ the Deontic Traditional Scheme}

\author[1]{Livio Robaldo}
\author[2]{Gianluca Pozzato}

\affil[1]{School of Law, Swansea University {\tt (livio.robaldo@swansea.ac.uk)}}
\affil[2]{University of Turin {\tt (gianluca.pozzato@unito.it)}}

\date{\today}

\begin{document}

\date{}
\maketitle

\begin{abstract}
\noindent This paper presents a new ontology that implements the well-known Deontic Traditional Scheme in RDFs and SPARQL, fit to handle {\it irresolvable conflicts}, i.e., situations in which two or more statements prescribe conflicting obligations, prohibitions, or permissions, with none of them being ``stronger'' than the other one(s). 

%THIS IS WEAK AND BORING TO READ
%The result is a novel framework for representing and inferring conflicts in the Semantic Web, towards further developments in disciplines aiming at checking compliance on Big Data with respect to the in-force regulations. 

In our view, this paper marks a significant advancement in standard theoretical research in formal Deontic Logic. Most contemporary approaches in this field are confined to the propositional level, mainly focus on the notion of obligation, and lack implementations. The proposed framework is encoded in RDF, which is not only a first-order language but also the most widely used knowledge representation language, as it forms the foundation of the Semantic Web. Moreover, the proposed computational ontology formalizes all deontic modalities defined in the Deontic Traditional Scheme, without specifically focusing on obligations, and offers constructs to model and reason with various types of irresolvable conflicts, violations, and the interaction between deontic modalities and contextual constraints in a given state of affairs. To the best of our knowledge, no existing approach in the literature addresses all these aspects within a unified integrated framework.

%QUESTO PARAGRAPH NO, NON È ABBASTANZA "ACCATTIVANTE". 
%The computational ontology was designed by integrating contributions from three different research strands, which have been investigated almost independently in past literature: (1) LegalTech solutions compatible with the RDF format, in particular the work in \cite{Robaldo-etal:23a}, (2) Natural Language Semantics via reification, in particular the work in \cite{Gordon-Hobbs:17}, and (3) conflict-tolerant accounts in formal Deontic Logic, in particular the work in \cite{Goble:13}. 

All examples presented and discussed in this paper, together with Java code and clear instructions to re-execute them locally, are available at \url{https://github.com/liviorobaldo/conflict-tolerantDeonticTraditionalScheme}.
\end{abstract}

\section{Introduction: contradictions vs conflicts}

\noindent Normative reasoning aims at formalizing norms from existing legislation in formal logic, fit to enable  automated compliance checking and other legal reasoning tasks. In this paper, we are specifically interested in automated compliance checking on data encoded in Resource Description Framework (RDF), {\it the} standard format for data interchange and linking on the World Wide Web.

Recent literature, e.g., \cite{Robaldo-etal:23a}, has highlighted the need of devising automated solutions for compliance checking fully interoperable with the RDF standard, given the increasing availability of Big Data in this format and, consequently, the need to check their compliance with respect to the in-force regulations. Nevertheless, the work in  \cite{Robaldo-etal:23a}, which reviews and compares main currently available automated legal reasoners, showed that only few of these reasoners are able to directly process RDF data; for the others, it is first necessary to translate the RDF triples into their input formats, a solution that is not practical when dealing with Big Data because it would increase the overall computational time beyond an acceptable level. 

In addition, \cite{Robaldo-etal:23a} highlighted the need of defining a common conceptualization of the norms to be shared among the legal reasoners, in order to enable comparisons thus researching optimal solutions. This paper aims at filling this gap, namely at developing a computational ontology compatible with Semantic Web technologies.

To define such an ontology, we reviewed past literature in deontic logic, the research area that investigates how to represent deontic statements such as obligations, prohibitions, permissions, etc. and related concepts in formal logic. Deontic logic has been used since the 1950s as the main formal tool for modelling normative reasoning \cite{Gabbay-etal:13}. From the literature review, which will be summarized, although not exhaustively, below in section \ref{DTSandConflictTolerantDLs}, we decided to ground the design of the proposed computational ontology on the distinction between {\it contradictions} and {\it conflicts}. 

Let us exemplify this distinction by consider the following two obligations:

\enumsentence{\label{conflictingobligations1}
\begin{enumerate}
    \renewcommand{\labelenumi}{\alph{enumi}.}
    \item It is obligatory to leave the building.
    \vspace{5pt}
    \item It is obligatory to not leave the building.
\end{enumerate}
}

\noindent (\ref{conflictingobligations1}.a) could hold, for instance, in a state of affairs where there is some danger {\it inside} the building, e.g., a fire, while (\ref{conflictingobligations1}.b) could hold in a state of affairs where there is some danger {\it outside} the building, e.g., a sand storm. 

The interesting question is of course what to do when there is {\it both} a fire in the building {\it and} a sand storm outside. In such a context, we must necessarily decide which one of the two dangers we want to take the risk with and we must violate the obligation associated with that danger.

In light of this, this paper assumes that all deontic logics that represent the conjunction of (\ref{conflictingobligations1}.a) and 
(\ref{conflictingobligations1}.b) as {\it contradictory}, first of all Standard Deontic Logic \cite{vonWright:51}, are inadequate for modelling normative reasoning. 

Contradictions should be associated with statements that are {\it illogical} in the state of affairs in which they are uttered and, as such, always false. For instance, the statement ``Yoof is a dog and Yoof is a cat'' is contradictory in states of affairs where the set of dogs and the set of cats are disjoint. This would be illogical.

On the contrary, the conjunction of (\ref{conflictingobligations1}.a) and 
(\ref{conflictingobligations1}.b) does not seem to be illogical, not even when there is both a fire in the
building and a sand storm outside. In such a context, we will simply have to violate one of the two obligations; however, as it is
well-known, violations are {\it not} contradictions and so there is no apparent reason to stipulate that, on the contrary, a situation in which it is necessary to violate some obligations is contradictory, i.e., illogical.

Therefore, the framework proposed in this paper will not model (\ref{conflictingobligations1}.a) and (\ref{conflictingobligations1}.b) as contradictory, but rather as {\it conflicting} of one another, in which a conflict is defined as a situation in which two deontic statements hold in the context but complying with one of them entails violating the other one. This notion of conflict, originally investigated by Hans Kelsen \cite{Kelsen:91}, also encompasses cases where one of the two deontic statements is a permission, although  \cite{Vranes:06} suggests that in such cases the conflict is ``unilateral'', i.e., it exists in only one direction. A simple example is:

\enumsentence{\label{conflictingobligations1subcategory}
\begin{enumerate}
    \renewcommand{\labelenumi}{\alph{enumi}.}
    \item It is prohibited to leave the building.\hspace{5pt} [\hspace{1pt}logically equivalent to (\ref{conflictingobligations1}.b)\hspace{1pt}]
    \vspace{5pt}
    \item It is permitted to leave the building.
\end{enumerate}
}

\noindent If both (\ref{conflictingobligations1subcategory}.a-b) hold and we do not leave the building in order to comply with (\ref{conflictingobligations1subcategory}.a), we cannot say that we are actually ``violating'' (\ref{conflictingobligations1subcategory}.b): the latter does not state that we {\it must} leave the building, i.e., that this is an obligation. On the contrary, if we leave the building, in virtue of what (\ref{conflictingobligations1subcategory}.b) authorizes us to do, we do violate (\ref{conflictingobligations1subcategory}.a). 

A final category of conflicts discussed by Kelsen is exemplified in (\ref{conflictingobligationsPartial}).

\enumsentence{\label{conflictingobligationsPartial}
\begin{enumerate}
    \renewcommand{\labelenumi}{\alph{enumi}.}
    \item It is obligatory to pay in cash.
    \vspace{5pt}
    \item It is obligatory to pay by card.
\end{enumerate}
}

\noindent Kelsen classifies conflicts like the one between (\ref{conflictingobligationsPartial}.a) and (\ref{conflictingobligationsPartial}.b) as ``partial conflicts'' because what is in conflict here is only {\it a part} of the same obligatory action: 
(\ref{conflictingobligationsPartial}.a-b) prescribe two different instruments for paying and the conflict stems from the fact that these instruments are mutually exclusive: payments made in cash are not made by card and payment made by card are not made in cash.

In this paper, however, we are not interested in subcategorizing conflicts; therefore, all conflicts exemplified in (\ref{conflictingobligations1}), (\ref{conflictingobligations1subcategory}), and (\ref{conflictingobligationsPartial}) are here considered as conflicts of the same type.

In deontic logic literature, conflicts as such are also known as ``irresolvable conflicts'', in which ``irresolvable'' means that none of the two deontic statements is stronger than, i.e., can override, the other one, in which case the latter is not actually violated. Several {\it conflict-tolerant} deontic logics have been proposed to formalize conflicts, although they mostly focus on conflicts between {\it obligations}, i.e., they do not consider conflicts such as the one exemplified in (\ref{conflictingobligations1subcategory}), in which one of the two deontic statements is a permission.

Lou Goble is perhaps the author who mostly investigated the formalization of conflicts in deontic logic; his seminal work in \cite{Goble:13} is still considered nowadays as a reference survey\footnote{See \url{https://plato.stanford.edu/entries/logic-deontic/#DeonDileConfToleDeonLogiRejeNC}} on the topic. 

All conflict-tolerant deontic logics reviewed in \cite{Goble:13} represent conflicts as {\it consistent} formulae; in that sense these logics are said to be ``conflict-tolerant''. However, as it will be clarified below, in almost\footnote{\noindent The exceptions are Adaptive Deontic Logics, e.g., \cite{Goble:13b} \cite{vanDePutte-etal:19}, in which conflicts are seen as ``abnormalities'' and stored in a separate set during the derivation. See subsubsection \ref{otherRadicalStrategies} below.} all approaches reviewed in \cite{Goble:13} conflicts cannot be distinguished from other consistent formulae. In our view, this is not desirable either because, although conflicts are not contradictions, they still need to be {\it notified} as they must be {\it removed}.

It is not so infrequent for existing legislation to include conflicts among norms. These conflicts are rather difficult to be manually identified by humans, e.g., the legislators in charge of updating the law. Artificial Intelligence could be then of great help here, as it may lead to the creation of LegalTech applications able to {\it detect} them, for the legislators to update the law fit to remove them.

Similar considerations might be found in a recent interview to Leon van der Torre and Dov Gabbay published in \cite{Steen-Benzmuller:24}, in which the explicit representation of fallacies, violations, mistakes, etc., (henceforth referred under the general term ``abnormalities'') fit to {\it reason} about them, has been identified as a crucial gap of contemporary logical frameworks for AI. Indeed, even in the LegalTech application envisioned above, if conflicts would be explicitly represented, the application could not only notify them to the legislator, but also {\it reason} about them fit to suggest alternative  solutions for solving them, while assessing pros and cons of each solution, etc., for the legislator to better ponderate the decision on how to revise the law.

Furthermore, in our view conflicts are not the only type of ``abnormality'' that ought to be notified. The state of affairs may include {\it physical constraints} that either prevent compliance with obligations and prohibitions or that prevent to execute what is permitted. Situations as these must be notified as well because, of course, agents cannot be sanctioned if it was {\it impossible} for them to comply with their obligations. 

Contextual constraints might be also used by agents in order to infer {\it how} they can comply with their obligations. For instance, consider the obligation in (\ref{conflictingobligations2}.a) and the constraint in (\ref{conflictingobligations2}.b).

\enumsentence{\label{conflictingobligations2}
\begin{enumerate}
    \renewcommand{\labelenumi}{\alph{enumi}.}
    \item Whoever parks in a parking spot is obliged to pay £3 at the parking meter associated with that spot.
    \vspace{5pt}
    \item The parking meter in Sketty only accepts cash.
\end{enumerate}
}

\noindent (\ref{conflictingobligations2}.b) states that, in Sketty, it is necessary to pay by cash. Therefore, if John is parking in Sketty, not only he will infer that he is obliged to pay £3; more specifically, he will infer that he is obliged to pay the £3 {\it in cash}.

Finally, contextual constraints might also interact with conflicts. For instance, suppose that, in the future, the government will decide to abolish cash as a way to contrast corruption. In the new context, only payments by digital means, e.g., credit card, are allowed. The following prohibition is then added to the normative system:

\enumsentence{\label{conflictingobligations3}
    \mbox{}\hspace{10pt}It is prohibited to pay by cash.
}

\noindent Now, it became impossible for John to comply with both (\ref{conflictingobligations2}.a) and (\ref{conflictingobligations3}) when he parks in Sketty: the physical constraint in (\ref{conflictingobligations2}.b) prevents to comply with the prohibition in (\ref{conflictingobligations3}), so either John will violate this prohibition or he will violate his obligation to pay £3; in either case, John's violation is justified and so he should not be sanctioned for that.

The interplay between norms and contextual knowledge holding in the state of affairs has been scarcely considered in past conflict-tolerant deontic logics, at least at the level of granularity exemplified in (\ref{conflictingobligations2}) and (\ref{conflictingobligations3}). Most deontic logics proposed in past literature use {\it propositional} symbols as atomic formulae while, in order to model examples (\ref{conflictingobligations2}) and (\ref{conflictingobligations3}) above, we need a first-order framework capable of distinguishing the actions, e.g., paying, from their thematic roles, e.g., the {\it instrument} of paying (cash rather than card).

This paper presents a novel {\it computational ontology} to represent and reason with conflicts between deontic statements in all cases exemplified so far, most of which are not currently covered by state-of-the-art conflict-tolerant deontic logics. The proposed computational ontology aims at being the first step to define, in the future, LegalTech applications able to notify and reason with conflicts between norms from legislation.

RDF and the other W3C standards provide the required level of granularity. Most important of all, they provide {\it interoperability} with Big Data in RDF format publicly available in the World Wide Web, which are expected to grow in number in the forthcoming years.

In light of this, we envision a future in which, whenever new norms will enter into force, LegalTech applications such as the one advocated above will be able to automatically download publicly available data and to identify which norms either conflict of one another or they cannot be complied with and why. For instance, with respect to the example in  (\ref{conflictingobligations2}) and (\ref{conflictingobligations3}), we might assume that, in the future, the websites of the municipalities will publish online, in RDF format, information about the city services and infrastructures, including information about the parking meters located in the city. With these data at disposal, applications will be able to automatically infer and notify that it is impossible to comply with (\ref{conflictingobligations3}) in the specific case of the parking meter in Sketty.

In light of all considerations above, the research questions of this paper are summarized as follows:

\enumsentence{\label{RQs}
\begin{enumerate}
    \renewcommand{\labelenumi}{\alph{enumi}.}    
    
    \item How is it possible to {\it explicitly represent} conflicts between norms, while distinguishing them from contradictions, with an eye on future LegalTech applications that might notify them to the legislators?
    \vspace{5pt}
    \item What else should be also explicitly represented, in light of what is {\it impossible or necessary} in the context?
    \vspace{5pt}
    \item How is it possible to implement an automated reasoner to represent and reason with (a) and (b) {\it by using W3C standards only}, in order to enhance interoperability with publicly available RDF datasets?
\end{enumerate}
}

\noindent The next section will review recent literature on compliance checking on RDF data; the aim of the next section is to explain which W3C standards we have specifically chosen to develop the proposed computational ontology.

Section \ref{EncodingNLstatements} will then illustrate the second main building block from the literature that we imported in our solution. This is the framework for Natural Language Semantics described in \cite{Gordon-Hobbs:17}, which in turn summarizes the lifetime research work of Jerry R. Hobbs, developed throughout many of his previous papers, e.g., \cite{Hobbs:86}, \cite{Hobbs-etal:93}, \cite{Hobbs:85}, and \cite{Hobbs:08}. We have chosen Hobbs's framework because, in order to formalize natural language statements, e.g., those from existing legislation, we need a framework explicitly designed for Natural Language Semantics. Hobbs's is formally simple yet very expressive, as well as grounded on theories of commonsense psychology that we consider valid. In addition, Hobbs's formulae can be straightforwardly implemented in RDF and the other Semantic Web technologies, as it will shown below, making it particularly suitable for the objectives of this paper.

It must be also pointed out that the first author of this paper has already used Hobbs's framework to define a particular deontic logic tailored for Natural Language Semantics called ``reified Input/Output logic'' \cite{Robaldo-Sun:17}\cite{Robaldo-etal:20}. In reified Input/Output logic, Hobbs's framework is plugged into Input/Output logic \cite{Makinson-vanderTorre:00}, a well-known deontic logic more computationally efficient than other deontic logics grounded on possible-world semantics \cite{Sun-Robaldo:17}. Nevertheless, reified Input/Output logic is not suitable to address the research questions in (\ref{RQs}) because also its axiomatization represents conflicts as contradictions\footnote{This axiomatization is specifically the one defined in \cite{Sun-vanDerTorre:14}, which is turn imported from \cite{Parent-vanderTorre:14}: given a set of obligations $N$ and a set of input formulae $A$, the set of outputs $O_3$($N$, $A$) is defined as $x\in O_3$($N$, $A$) iff there is some finite $M\subseteq N$ such that: (1) $M$($Cn$($A$)) $\neq\emptyset$; (2) for all $B$ such that $A\subseteq B=Cn$($B$) $\supseteq M$($B$), $x\leftrightarrow\wedge$$M$($B$). In (1) and (2), $M$($X$) is the set of outputs given the pairs in $M$ and the inputs in $X$ while $Cn$ is the transitive closure under the derivation rules of the object logic. It is easy to see that this definition derives a contradiction from the conjunction of (\ref{conflictingobligations1}.a) and (\ref{conflictingobligations1}.b). This conjunction is represented by taking $N\equiv\{$($\top$, {\tt L}), ($\top$, $\neg${\tt L})$\}$ and $A\equiv\{\top\}$, where $\top$ refers to ``true''; the contradiction is immediately derived when $M$=$N$: $\top\in B$ because $A\subseteq B$, and so $\{{\tt L}, \neg{\tt L}\}\subseteq M$($B$) and $x\leftrightarrow({\tt L}\wedge\neg{\tt L}$) $\leftrightarrow\bot$.}. Thus, the proposed solution will keep Hobbs's framework but it will define a {\it new} axiomatization to properly address the research questions in (\ref{RQs}.a-b).

In order to understand how the new axiomatization is capable to achieve (\ref{RQs}.a-b), section 
\ref{DTSandConflictTolerantDLs} will review past literature on conflict-tolerant deontic logics, while highlighting insights and limitations of the main approaches.

Section \ref{DeonticModalitiesInRDFsAndSPARQL} will then present the core research of this paper: how to integrate the contributions from the three research strands illustrated in sections \ref{BackgroundLegalTechOnRDF},
\ref{EncodingNLstatements}, and 
\ref{DTSandConflictTolerantDLs} into a single unified RDF-based framework to model and reason with deontic modalities while achieving the research questions in (\ref{RQs}.a-c).
After that, sections \ref{ConjunctionDisjunctionImplicationOfDeonticStatements} and \ref{DeonticModalitiesAndContextualConstraints} will compare the proposed computational ontology with state-of-the-art conflict-tolerant deontic logics, along two different and orthogonal perspectives. Finally, section \ref{FutureWorks} will address some future works and section \ref{Conclusions} will conclude the paper.

\section{Background:  compliance checking on RDF data}\label{BackgroundLegalTechOnRDF}

\noindent As it is well-known, RDF is a graph-based data model that only {\it represents} knowledge, while it does not provide inference mechanisms to compute what logically follows from what is explicitly asserted. These mechanisms are instead needed for compliance checking, in order to infer whether a given state of affairs is compliant or not with respect to a given set of norms. In light of this, approaches from the past literature in modelling compliance checking with RDF data use a separate executable format, compatible with RDF, to encode the norms as inference rules; then, an automated reasoner able to process this format executes the inference rules.

Some of the first approaches belonging to this literature are \cite{Gordon:08} and \cite{Ceci:13}; these respectively formalize norms in Semantic Web Rule Language\footnote{\url{https://www.w3.org/Submission/SWRL}} (SWRL) and Legal Knowledge Interchange Format (LKIF) rules\footnote{\url{http://www.estrellaproject.org/doc/D1.1-LKIF-Specification.pdf}}. SWRL is a proposed language for the Semantic Web since 2004; however, it never became a W3C recommendation. LKIF rules are special if-then rules developed in the context of a past European project together with other resources widely used nowadays, the most important of which being the LKIF ontology \cite{Hoekstra-etal:07}. Automated reasoners to execute either SWRL or LKIF rules are publicly available on the Web. 

More recently, \cite{deVos-etal:19} and \cite{Palmirani-Governatori:18} propose to respectively model norms in the OASIS standards RuleML\footnote{\url{http://wiki.ruleml.org}} and LegalRuleML\footnote{\url{https://www.oasis-open.org/committees/legalruleml}}. LegalRuleML is in particular a specialization of RuleML for normative reasoning and it aims at being a standard for representing and interchanging the meaning of norms from legislation. However, since it does not prescribe any formal semantics with respect to which interpreting the rules, transformations must be defined from LegalRuleML into executable reasoning languages. For instance, in \cite{Palmirani-Governatori:18}, LegalRuleML representations are translated in the input format of a special reasoner called ``SPINdle'' \cite{Lam-Governatori:09}.

A similar approach is \cite{Gandon-etal:17}, which propose to extend the LegalRuleML meta model \cite{Athan-etal:13} 
and formalize it as an OWL ontology; norms are then represented as SPARQL rules in the form {\tt DELETE/INSERT-WHERE}. The use of SPARQL in combination with OWL is motivated by the fact that OWL does not support defeasibility, time management, and other operators that are instead required in normative reasoning. 

The fact that the expressivity of OWL only suffices for a limited set of normative inferences has been also discussed in \cite{Bonatti-etal:20}, which presents an OWL2-based framework for GDPR policy validation. OWL2 has been chosen in \cite{Bonatti-etal:20} to keep computational complexity under control; however, the authors themselves acknowledge (see \cite{Bonatti-etal:20}, \S3.3) that, due to the limited expressivity of OWL2, their framework is unable to represent and process defeasible norms, i.e., norms that might be overridden by other ``stronger'' norms.

Modelling norms in OWL2 to keep computational complexity under control was also the objective of \cite{Francesconi-Governatori:23}. Similarly to \cite{Bonatti-etal:20}, \cite{Francesconi-Governatori:23} uses OWL2 subsumption to infer the sets of individuals that comply with the norms. Contrary to 
\cite{Bonatti-etal:20}, however, \cite{Francesconi-Governatori:23} additionally propose to model defeasible norms by introducing OWL restrictions that refer to the complement subclasses. Thus, individuals that satisfy exceptions of the general norms, i.e., norms that override the latter in specific contexts, are inferred as members of special subclasses, which are then defined as {\tt owl:disjointWith} the sets of individuals that comply with the norms. 

An similar solution, proposed in \cite{Robaldo:21}, is to use SHACL Triple rules\footnote{\url{https://www.w3.org/TR/shacl-af}} for encoding the inference rules. These rules might be prioritized, thus providing a more straightforward and standard way to implement defeasibility. 

Nevertheless, both \cite{Francesconi-Governatori:23} and \cite{Robaldo:21} have been recently criticized in \cite{Anim-etal:24}; the latter shows that the semantics of certain norms requires aggregate functions and temporal management, which are supported by SPARQL but not by OWL or by SHACL Triple rules. Therefore, \cite{Anim-etal:24} propose to use SHACL-SPARQL rules. These are like SHACL Triple rules but allow for the embedding of SPARQL rules in the form {\tt CONSTRUCT-WHERE}, which provide the expressivity needed to deal with aggregate functions and temporal management. On the other hand, SHACL-SPARQL rules, like SWRL or SHACL Triple rules, are not (yet?) a W3C recommendation.

Finally, it is worth mentioning the ASP-based reasoners DLV2 \cite{Robaldo-etal:23b} and Vadalog \cite{Bellomarini-etal:22}, although they have been scarcely used for normative reasoning so far. On the one hand, both DLV2 and Vadalog are compatible with RDF and SPARQL; on the other hand, they are extremely efficient because they have been primarily designed to cope with NP-hard search problems, and so they are particularly suitable to process Big Data. Similarly, \cite{Governatori:23} recently proposed an implementation in ASP of Defeasible Deontic Logic \cite{Governatori-etal:13}, a well-known deontic logic used for normative reasoning. Modelling and reasoning with conflicts among deontic statements in Defeasible Deontic Logic is the object of ongoing research \cite{Governatori-Rotolo:23}. Nevertheless, contrary to DLV2 and Vadalog, the implementation in \cite{Governatori:23} does not support the RDF format, thus it is not suitable for the objective of this paper, specifically research question (\ref{RQs}.c).

In light of this very brief, and of course not exhaustive, literature review on automated compliance checking on RDF data, for this paper we decided to create a simple ad-hoc lightweight automated reasoner for SPARQL rules that iteratively re-executed the rules until no further new RDF triple is inferred. We chose to implement the inference rules as simple iterative SPARQL rules because computational efficiency is outside the scope of this paper. Nor we are interested in defeasibility because, as explained in the introduction, this paper only considers irresolvable conflicts, i.e., conflicts among norms in which none of them can override the others; therefore, standards constructs used to implement defeasibility, e.g., priorities as in \cite{Anim-etal:24} or partial orders as in Defeasible Deontic Logic, would just increase the verbosity of the rules without contributing towards the objectives of this paper.
An efficient and defeasible version of the lightweight automated reasoner that we created for this paper, perhaps implemented as a module of DLV2 or Vadalog, is seen as the object of future work.

\subsection{The lightweight automated reasoner used in this paper}\label{LightweightAutomatedReasoner}

\noindent The inference rules defined below in the paper are implemented as SPARQL rules, specifically as SPARQL rules in the form {\tt CONSTRUCT-WHERE}, similarly to what is done in \cite{Anim-etal:24}. As explained above, such rules are not expressive enough for normative reasoning in general because, for instance, they do not have priorities or other constructs to implement defeasibility. However, since this paper is not concerned with defeasibility, we opted for a simpler implementation to help the reader remain focused on the objectives of this paper.

The SPARQL rules are also part of the proposed computational ontology. In particular, the ontology includes a special class called {\tt InferenceRule} whose individuals refer to the SPARQL rules. A special RDF property {\tt has-sparql-code} associate these individuals with the strings encoding the rules in standard SPARQL v1.1. 

In all examples shown below or included in the GitHub repository the inference rules are anonymous RDF individuals, so they adhere to the following template\footnote{All RDFs representations in this paper and in the associated GitHub repository are encoded in Turtle syntax, \url{https://www.w3.org/TR/turtle}}:

\enumsentence{\label{SPARQLrulesTemplate}
\begin{minipage}[t]{400pt}\tt
\mbox{}\hspace{10pt}[a :InferenceRule;\\
\mbox{}\hspace{15pt}:has-sparql-code """CONSTRUCT\{$\ldots$\}WHERE\{$\ldots$\}"""].
\end{minipage}
}
\vspace{3pt}

\noindent The empty prefix ``{\tt :}'' is associated with the following namespace:

\begin{center}
{\tt https://w3id.org/ontology/conflict-tolerantdeontictraditionalscheme\#}
\end{center}

\noindent This is the namespace of the RDF resources included in the proposed computational ontology. We will use another prefix ``{\tt soa:}'', for the RDF resources that are part of the states of affairs, i.e., the ABoxes, encoding the examples: 

\begin{center}
{\tt https://w3id.org/ontology/conflict-tolerantdeontictraditionalscheme\#soa}
\end{center}

\noindent The automated reasoner processing the SPARQL rules, freely downloadable from the GitHub repository associated with this paper together with all examples shown below and clear instructions to re-execute them, has been implemented in Java by using the Apache Jena library\footnote{\url{https://jena.apache.org}} (v4.10), which is perhaps the most popular open source Java framework for building Semantic Web and Linked Data applications. As already mentioned earlier, the reasoner iteratively re-apply the rules to the inferred knowledge graph until no new RDF triples are inferred. 

Note that this is exactly what is also done in standard OWL reasoners, e.g., HermiT \cite{Glimm-etal:14}, which re-execute the OWL axioms until no further new RDF triple is inferred. Nevertheless, while the set of triples that might be iteratively inferred through OWL axioms are finite in number, and so the OWL reasoner will necessarily terminate after a finite number of iterations, the iterative re-execution of SPARQL rules may loop infinitely. The reason is that OWL axioms cannot extend the set of RDF resources; therefore, if {\tt R} is the set of RDF resources and {\tt P}$\subseteq${\tt R} the set of RDF properties, the cardinality of the maximal set of triples that can be created is |{\tt R}|$\times$|{\tt P}|$\times$|{\tt R}|. Conversely, SPARQL rules in the form {\tt CONSTRUCT-WHERE} can also add new (anonymous) individuals to the knowledge graph, thus extending its set of resources. This might easily lead to infinite loops. For instance, by iteratively re-executing the following rule, each man in the knowledge graph will be associated with an infinite number of wives, each corresponding to a new anonymous individual added by the rule to the knowledge graph in each iteration:

\enumsentence{\label{SPARQLsampleInfiniteLoopRule}
\begin{minipage}[t]{450pt}\tt
\mbox{}\hspace{0pt}[a :InferenceRule;\\
\mbox{}\hspace{15pt}:has-sparql-code """CONSTRUCT\{[soa:wife-of ?x]\}\\
\mbox{}\hspace{120pt}WHERE\{?x a soa:Man\}"""].
\end{minipage}
}
\vspace{2pt}

\noindent To avoid infinite loops, all SPARQL rules shown below and that create new anonymous individuals will include special clauses in the form \hspace{2pt}{\tt NOT\hspace{-2pt} EXISTS\{$\ldots$\}}\hspace{1pt} that block these loops. Specifically, the SPARQL rules shown below will create new anonymous individuals only if the knowledge graph {\it does not already contain individuals with the same characteristics}. For instance, the SPARQL rule in (\ref{SPARQLsampleInfiniteLoopRuleRevised}) is revised as follows:

\enumsentence{\label{SPARQLsampleInfiniteLoopRuleRevised}
\begin{minipage}[t]{450pt}\tt
\mbox{}\hspace{0pt}[a :InferenceRule;\\
\mbox{}\hspace{5pt}:has-sparql-code\hspace{-3pt} """CONSTRUCT\{[soa:wife-of ?x]\}\\
\mbox{}\hspace{80pt}WHERE\{\hspace{-1pt}?x\hspace{-1pt} a\hspace{-1pt} soa:Man.\hspace{-3pt} NOT\hspace{-1pt} EXISTS\{\hspace{-1pt}?w\hspace{-1pt} soa:wife-of\hspace{-1pt} ?x\hspace{-1pt}\}\hspace{-1pt}\}\hspace{-1pt}"\hspace{-1pt}"\hspace{-1pt}"\hspace{-1pt}]\hspace{-1pt}.
\end{minipage}
}
\vspace{2pt}

\noindent The revised rule creates a new wife and associate her with a man only if the man does not already have a wife. 

Although this solution is acceptable for all cases considered in this paper, it is not applicable in the general case in that it introduces restrictions that do not necessarily hold in all contexts; for instance, (\ref{SPARQLsampleInfiniteLoopRuleRevised}) is not applicable in contexts in which men are allowed to have more than one wife, for which more complex {\tt WHERE} clauses ought to be asserted. Nevertheless, the development of a proper reasoner for SPARQL rules is beyond the scope of this paper.

%This problem is not new and it concerns declarative languages in general, not only SHACL and SPARQL. For instance, it is well-known that Answer Set Programming (ASP)  with functional symbols in recursion can have a non-finite relevant ground program (see \cite{Calimeri-etal:09}, among others), leading to the infinite creation of functional terms; for example,  the ASP rule \hspace{2pt}{\tt p(f(x))}{\tt:-}{\tt p(x)} will generate the functional symbols {\tt f(x)}, {\tt f(f(x))}, {\tt f(f(f(x)))}, etc. in the answer set.

\section{Background: encoding natural language statements in (Gordon and Hobbs, 2017)}\label{EncodingNLstatements}

\noindent This section presents the framework illustrated in \cite{Gordon-Hobbs:17} as well as in previous publications by Jerry R. Hobbs; this framework sets the groundwork for the computational ontology proposed in this paper.

\cite{Gordon-Hobbs:17} presents a first-order logical framework for Natural Language Semantics massively grounded on the notion of {\it reification}. In philosophical terms, reification is the action of conceptualizing abstract entities as things of the world. In formal logic, this amounts at formalizing them as first-order {\it individuals}, i.e., constants or variables of the logic. The notion of reification was originally introduced by Donald Davidson in \cite{Davidson:67}; however, Donald Davidson was using reification with a limited set of abstract entities. Conversely, as stated above, \cite{Gordon-Hobbs:17} is {\it massively} grounded on the notion of reification, meaning that {\it any} abstract entity can be reified into a first-order individual; then, new assertions can be made on these individuals and the fact that one of these assertions hold for one of these individuals can be recursively reified again into a new first-order individual.

According to \cite{Gordon-Hobbs:17} and its philosophical and psycholinguistic foundations, reification parallels how people think about events, actions, and states in the world; for this reason, \cite{Gordon-Hobbs:17} is particularly suitable to handle the semantics of natural language statements and it has been chosen as the underlying logic of the proposed computational ontology.

Let's see how reification works on a simple example: the assertion ``John leaves''. This assertion is usually represented in standard first-order logic as {\tt leave}({\tt John}), where {\tt leave} is a first-order predicate and {\tt John} is a first-order individual. Conversely, in \cite{Gordon-Hobbs:17} this first-order assertion is associated with another first-order individual {\tt elj}, which is called an ``eventuality''. 
In \cite{Gordon-Hobbs:17}'s terminology, it is said that the assertion is ``reified into'' the eventuality {\tt elj}; in this paper we will also use the phrase ``the fact that'' to talk about the relation between the eventualities and the corresponding assertions; therefore, {\tt elj} denotes {\it the fact that} John leaves. Being first-order individuals, eventualities can be inserted as argument of first-order predicates; in particular, eventualities that reify actions or states such as ``to leave'' might be inserted as argument of predicates that denote particular {\it modalities} holding for the action or state. In light of this, \cite{Gordon-Hobbs:17} converts {\tt leave}({\tt John}) into the following formalization:

\begin{center}
{\tt Rexist}({\tt elj}) $\wedge$ {\tt leave}'({\tt elj}, {\tt John})
\end{center}

\noindent In which the predicate {\tt leave} has been converted in the primed predicate {\tt leave}', which includes an additional argument: the reification of the action. This reification is then inserted as argument of another predicate {\tt Rexist}, which represents the {\it modality} of John's leaving. In particular, {\tt Rexist}({\tt elj}) states that {\tt elj} {\it really exists} in the state of affairs. The use of the predicate {\tt Rexist} marks a neat difference between the way in which meaning is usually represented in standard first-order logic, e.g., {\tt leave}({\tt John}), and the way in which this is done in \cite{Gordon-Hobbs:17}'s framework. According to \cite{Gordon-Hobbs:17}, actions cannot be true or false; they can instead take place in the real world or not. 

However, {\tt Rexist} is not the single available modality.
As \cite{Gordon-Hobbs:17} explains, an eventuality could be part of someone's beliefs but not occur in the real world, it could exist in some fictional universe, it could be merely possible or likely but not real, etc. Each of these alternative modalities must be represented by a unary predicate different from {\tt Rexist}. This paper will  introduce predicates that implement deontic modalities, in section \ref{DeonticModalitiesInRDFsAndSPARQL} below.

Furthermore, in this paper, in order to facilitate the encoding of \cite{Gordon-Hobbs:17}'s formulae in RDFs, rather than defining primed predicate associated with their non-primed versions, we will adhere to the specific pattern proposed in \cite{Robaldo-etal:23a} and shown in (\ref{patternFormulae}). In (\ref{patternFormulae}), {\tt e} denotes an eventuality, {\tt M} denotes a modality, {\tt AoS} denotes an action or state while {\tt t$_{\tt *}$} denote thematic roles; the latter are represented as binary predicates having the eventuality {\tt e} as first argument and the value of the thematic role {\tt v$_{\tt *}$} as second one.

\enumsentence{\label{patternFormulae}
\hspace{5pt}{\tt M}\hspace{1pt}({\tt e})\hspace{2pt} $\wedge$\hspace{2pt} {\tt AoS}\hspace{1pt}({\tt e})\hspace{2pt} $\wedge$\hspace{2pt}
$\bigwedge\limits_{i=1}^n${\tt t$_{\tt i}$}\hspace{1pt}({\tt e}, {\tt v$_{\tt i}$}\hspace{1pt})
}

\noindent In this pattern, ``John leaves'' is represented as in (\ref{JohnLeavesAndPaysInHobbs}.a). Another example is shown in (\ref{JohnLeavesAndPaysInHobbs}.b), which represents the assertion ``John pays £3 in cash''.

\enumsentence{\label{JohnLeavesAndPaysInHobbs}
\begin{enumerate}
    \renewcommand{\labelenumi}{\alph{enumi}.}
    \item\hspace{-3pt} {\tt Rexist}({\tt elj}\hspace{1pt}) $\wedge$ {\tt Leave}({\tt elj}\hspace{1pt}) $\wedge$ {\tt has-agent}({\tt elj}, {\tt John}\hspace{0.5pt})
    
    \vspace{5pt}
    \item\hspace{-3pt} {\tt Rexist}({\tt epj}\hspace{1pt}) $\wedge$ {\tt Pay}({\tt epj}\hspace{1pt}) $\wedge$ {\tt has-agent}({\tt epj}, {\tt John}\hspace{0.5pt}) $\wedge$\\[2pt]
    \mbox{}\hspace{100pt}{\tt has-object}({\tt epj}, {\tt £3}\hspace{0.5pt})
    $\wedge$ {\tt has-instrument}({\tt epj}, {\tt cash})
\end{enumerate}
}

\noindent In (\ref{JohnLeavesAndPaysInHobbs}.a), {\tt Leave} is a predicate different from the previous predicate {\tt leave}, in that it applies to an eventuality rather than to a person. Therefore, while {\tt leave}({\tt John}) states that {\tt John} belongs to the set of leavers, {\tt Leave}({\tt elj}) states that {\tt elj} belongs to the set of leaving actions. On the other hand, {\tt Leave}({\tt elj}) does not state whether {\tt elj} also really exists in the state of affairs or only in someone's imagination, etc. To state that {\tt elj} really exists in the state of affairs, {\tt Rexist}({\tt elj}) must be asserted, so that {\tt elj} will also belong to the set of really existing eventualities.

The framework in \cite{Gordon-Hobbs:17} covers 
a fairly large theory of commonsense psychology while formalizing sets, abstract and instantiated eventualities, causal and temporal reasoning, composite entities, defeasibility, etc. via reification. The objectives of this paper do not need all these notions; therefore, this paper will import only the predicates and axioms from \cite{Gordon-Hobbs:17} strictly needed to address the research questions in (\ref{RQs}) above. Section \ref{FutureWorks}, devoted to future works, will discuss some extensions that we plan to incorporate within the proposed computational ontology in the future.

\subsection{Negation, conjunction, and disjunction of eventualities}\label{GordonHobbsNegationDisjunctionConjunction}

\noindent \cite{Gordon-Hobbs:17} defines the three predicates {\tt not}, {\tt and}, and {\tt or} to relate pairs or triples of eventualities. As it will explained below, although these predicates have the same name of the three well-known standard boolean connectives, they do {\it not} have the same meaning. The predicates {\tt not}, {\tt and}, and {\tt or} are intuitively defined as follows:

\enumsentence{\label{notandor}
\begin{enumerate}
    \renewcommand{\labelenumi}{\alph{enumi}.}
    \item ``{\tt not}\hspace{1pt}({\tt en}, {\tt e})'' states that {\tt en} and {\tt e} are {\it opposite} eventualities; e.g., if {\tt e} refers to the fact that ``John leaves'', {\tt en} may refer to the fact that ``John does not leave''. 
    \vspace{5pt}
    
    \item ``{\tt and}\hspace{1pt}({\tt ea}, {\tt e1}, {\tt e2})'' states that {\tt ea} is the {\it co-occurrence} of {\tt e1} and {\tt e2}; e.g., if {\tt e1} and {\tt e2} respectively refer to the facts that ``John eats'' and ``John drinks'', {\tt ea} may refer to the fact that ``John eats and drinks''.
    \vspace{5pt}
    
    \item ``{\tt or}\hspace{1pt}({\tt eo}, {\tt e1}, {\tt e2})'' states that {\tt eo} is the {\it disjunctive occurrence} of {\tt e1} and {\tt e2}; e.g., if {\tt e1} and {\tt e2} respectively refer to the facts that ``John eats'' and ``John drinks'', {\tt eo} may refer to the fact that ``John eats or drinks''.
\end{enumerate}
}

\noindent The modalities holding on {\tt en}, {\tt ea}, and {\tt eo} may be related with the modalities holding on {\tt e}, {\tt e1}, and {\tt e2}. For instance, with respect to the {\tt Rexist} modality, the  bi-implications in (\ref{GordonHobbsNotAndOrAxioms}) are stipulated as valid. Note that the bi-implications in (\ref{GordonHobbsNotAndOrAxioms}) make use of the standard boolean connectives $\neg$\hspace{1pt}, $\wedge$\hspace{1pt}, and $\vee$.

\enumsentence{\label{GordonHobbsNotAndOrAxioms}
\begin{enumerate}
    \renewcommand{\labelenumi}{\alph{enumi}.}
    \item $\forall_{{\tt e},\hspace{2pt} {\tt en}}$[ {\tt not}({\tt e}, {\tt en}) $\rightarrow$
    ({\tt Rexist}({\tt e}) $\leftrightarrow$
    $\neg$\hspace{1pt}{\tt Rexist}({\tt en})) ]
    \vspace{5pt}
    \item$\forall_{{\tt ea},\hspace{2pt}{\tt e1},\hspace{2pt} {\tt e2}}$[ {\tt and}\hspace{1pt}({\tt ea}, {\tt e1}, {\tt e2}) $\rightarrow$
    ({\tt Rexist}({\tt ea}) $\leftrightarrow$
    ({\tt Rexist}({\tt e1}) $\wedge$ {\tt Rexist}({\tt e2}))) ]
    \vspace{5pt}
    \item$\forall_{{\tt eo},\hspace{2pt}{\tt e1},\hspace{2pt} {\tt e2}}$[ {\tt or}\hspace{1pt}({\tt eo}, {\tt e1}, {\tt e2}) $\rightarrow$
    ({\tt Rexist}({\tt eo}) $\leftrightarrow$
    ({\tt Rexist}({\tt e1}) $\vee$ {\tt Rexist}({\tt e2}))) ]
\end{enumerate}
}
\vspace{2pt}

\noindent Therefore, if the fact that John leaves really exists, then the fact that John does not leave does not really exist (and vice versa), if the fact that John eats and drinks really exists, then also the fact that John eats and the fact that John drinks really exist (and vice versa), and if the fact that John eats or drinks really exists, then also the fact that John eats or the fact that John drinks really exist (and vice versa).

However, the bi-implications in (\ref{GordonHobbsNotAndOrAxioms}) only concern the {\tt Rexist} modality whereas corresponding bi-implications for other modalities could not be as valid as those in (\ref{GordonHobbsNotAndOrAxioms}). For instance, the fact that John is obliged to leave is {\it not} equivalent to the fact is not obliged to stay (i.e., to not leave): the meaning of the former is that John must leave otherwise he will violate the corresponding norm and perhaps he will be even sanctioned for that; conversely, the meaning of the latter is  that John does not have the obligation to stay and, therefore, unless he is subject to some other norms, that he is basically free to do whatever he wants: he can either leave or stay, as he wishes.

The bi-implications in (\ref{GordonHobbsNotAndOrAxioms}) make every theorem using the boolean connectives $\neg$\hspace{1pt}, $\wedge$\hspace{1pt}, and $\vee$ on the predicate {\tt Rexist} equivalent to a corresponding theorem using the predicates {\tt not}, {\tt and}\hspace{1pt}, and {\tt or}. For instance, the formula in (\ref{OrNotTheoremRexist}.a) bi-implies the formula in (\ref{OrNotTheoremRexist}.b), given the equivalences in (\ref{GordonHobbsNotAndOrAxioms}). Both (\ref{OrNotTheoremRexist}.a) and (\ref{OrNotTheoremRexist}.b) instantiate the so-called ``Disjunctive Syllogism'', i.e., the implication (({\tt A}$\vee${\tt B})\hspace{1pt}$\wedge$\hspace{0pt}$\neg${\tt B})\hspace{1pt}$\rightarrow$\hspace{1pt}{\tt A}, on the predicate {\tt Rexist}.

\enumsentence{\label{OrNotTheoremRexist}
\begin{enumerate}
    \renewcommand{\labelenumi}{\alph{enumi}.}
    \item$\forall_{{\tt eo},\hspace{2pt}{\tt e1},\hspace{2pt} {\tt e2},\hspace{2pt} {\tt en1}}$[\hspace{2pt}(\hspace{1pt}{\tt Rexist}({\tt eo}) $\wedge$ {\tt or}\hspace{1pt}({\tt eo}, {\tt e1}, {\tt e2}) $\wedge$
    {\tt Rexist}({\tt en1}) $\wedge$
    {\tt not}({\tt en1}, {\tt e1})\hspace{1pt}) $\rightarrow$
    {\tt Rexist}({\tt e2})\hspace{2pt}]
    \vspace{5pt}
    
    \item$\forall_{{\tt e1},\hspace{2pt} {\tt e2}}$[(\hspace{1pt}({\tt Rexist}({\tt e1}) $\vee$ {\tt Rexist}({\tt e2})) $\wedge$
    $\neg$\hspace{1pt}{\tt Rexist}({\tt e1})\hspace{1pt}) $\rightarrow$
    {\tt Rexist}({\tt e2})\hspace{2pt}]
\end{enumerate}
}
\vspace{2pt}

\noindent Other theorems on {\tt Rexist} such as 
the distributivity laws on conjunction and disjunction or De Morgan's laws might be ``translated'' \hspace{1pt}to\hspace{1pt}/\hspace{1pt}from the predicates {\tt not}, {\tt and}, and {\tt or}  from\hspace{1pt}/to the boolean connectives $\neg$, $\vee$, and $\wedge$\hspace{1pt}, as the reader can easily verify. However, the formalization below does not include them as they are not important for the objectives of this paper. On the other hand, the present paper will investigate how {\tt not}, {\tt and}, and {\tt or}  relate to the boolean connectives $\neg$, $\vee$, and $\wedge$\hspace{1pt} with respect to the deontic modalities. 

\subsection{Abstract and instantiated eventualities}\label{AbstractEventualitiesHobbs}

\noindent The second main ingredient that this paper imports from \cite{Gordon-Hobbs:17} is the distinction between abstract eventualities and their instantiations. This distinction is needed in particular to represent partial conflicts such as the one exemplified in above in 
(\ref{conflictingobligationsPartial}). Let us introduce the notion of abstract eventuality via the following example, much simpler than (\ref{conflictingobligationsPartial}):

\enumsentence{\label{conflictingFactsPartialOnJohn}
\begin{enumerate}
    \renewcommand{\labelenumi}{\alph{enumi}.}
    \item John pays in cash.
    \vspace{5pt}
    \item John pays by card.
\end{enumerate}
}

\noindent The two sentences in (\ref{conflictingFactsPartialOnJohn}) contradict of one another only if it is assumed that they refer to the {\it same} payment. Same considerations hold for the conflict in (\ref{conflictingobligationsPartial}): the two obligations are in conflict only under the assumption that they refer to the same payments. Under this assumption, the contradiction in (\ref{conflictingFactsPartialOnJohn}) and the conflict in (\ref{conflictingobligationsPartial}) stem from the fact that cash and card are two mutually exclusive instruments for paying\footnote{Unless the two sentences are interpreted under a {\it cumulative} reading, i.e., one in which John is paying part of the sum in cash and the rest by card. We do not consider these readings in this paper, although we will further comment on them below in section \ref{FutureWorks}, devoted to future works. Nevertheless, even under such an interpretation, it may be argued that {\it two} payment are indeed taking place, one in cash and the other one by card; it is only the sum of the amounts paid that, cumulatively, is equal to the overall amount of cash that John needs to pay.}. 

Nevertheless, the two sentences in (\ref{conflictingFactsPartialOnJohn}.a-b) do {\it not} explicitly states anywhere that they refer to the same payment: {\it what} John is paying is unknown in (\ref{conflictingFactsPartialOnJohn}.a) and (\ref{conflictingFactsPartialOnJohn}.b). Therefore, perhaps in (\ref{conflictingFactsPartialOnJohn}.a) he is obliged to pay {\it the rent} in cash while in (\ref{conflictingFactsPartialOnJohn}.b) he is obliged to pay {\it the taxes} by card. In such a case, (\ref{conflictingFactsPartialOnJohn}.a-b) are not in contradiction.

In other words, the fact that (\ref{conflictingFactsPartialOnJohn}.a-b) refer to the same payment is just a {\it pragmatic implicature}\footnote{See \url{https://plato.stanford.edu/entries/implicature}} that we make while interpreting the example in the context of this paper's scientific discussion. On the other hand, from a strict logical/semantic point of view, natural language sentences, as well as Hobbs's formulae and their RDF translations that will be shown below, adheres to the {\it Open World Assumption}, meaning that what is not specified is simply {\it unknown} and so it is not possible to reason about it unless we {\it spell it out}. We will therefore add inference rules that implement special pragmatic implicature, in order to impose eventualities such as those in (\ref{conflictingFactsPartialOnJohn}.a) and (\ref{conflictingFactsPartialOnJohn}.b) to be two different ``variants'' of the same payment, which differ of one another only for the instrument used.

The solution adopted in \cite{Gordon-Hobbs:17} to deal with issues like this is to introduce special eventualities called ``abstract eventualities'', which are then ``instantiated'' into more specific eventualities. For instance, (\ref{conflictingFactsPartialOnJohn}.a) and (\ref{conflictingFactsPartialOnJohn}.b) can be respectively denoted by two individuals {\tt epjcash} and {\tt epjcard}, which are two different instantiations of the same abstract eventuality that denotes the same contextually relevant {\it abstract} payment made by John. {\tt epjcash} and {\tt epjcard} are actions of the same type and they share the same thematic roles with the same value except the {\it instrument}: for {\tt epjcash} it is {\tt cash} while for {\tt epjcard} it is {\tt card}. This is encoded in Hobbs's as follows:

\vspace{2pt}
\enumsentence{\label{hasInstrumentCashCardHobbs}
\hspace{10pt}{\tt has-instrument}\hspace{1pt}({\tt epjcash}\hspace{1pt}, {\tt cash}\hspace{1pt}) $\wedge$ {\tt has-instrument}\hspace{1pt}({\tt epjcard}\hspace{1pt}, {\tt card}\hspace{1pt})
}
\vspace{2pt}

\noindent As explained above, these two values are mutually exclusive with respect to the action of paying and the thematic role {\tt has-instrument}. This may be encoded via the implications in (\ref{payCashNotCardAndCardNotCashHobbs}.a-b), which state that, for every payment, if the instrument is cash then it is not card and if the instrument is card then it is not cash.

\vspace{2pt}
\enumsentence{\label{payCashNotCardAndCardNotCashHobbs}
\begin{enumerate}
    \renewcommand{\labelenumi}{\alph{enumi}.}
    \item $\forall_{\tt e}$[({\tt Pay}\hspace{1pt}({\tt e}\hspace{1pt}) $\wedge$ {\tt has-instrument}\hspace{1pt}({\tt e}, {\tt cash})) $\rightarrow$ $\neg$\hspace{1pt}{\tt has-instrument}\hspace{1pt}({\tt e}, {\tt card})]
    \vspace{5pt}
    \item $\forall_{\tt e}$[({\tt Pay}\hspace{1pt}({\tt e}\hspace{1pt}) $\wedge$ {\tt has-instrument}\hspace{1pt}({\tt e}, {\tt card})) $\rightarrow$ $\neg$\hspace{1pt}{\tt has-instrument}\hspace{1pt}({\tt e}, {\tt cash})]
\end{enumerate}
}

\noindent From (\ref{hasInstrumentCashCardHobbs}) and (\ref{payCashNotCardAndCardNotCashHobbs}.a-b), the following is inferred, i.e., that the instrument of {\tt epjcash} is {\it not} card and that the instrument of {\tt epjcard} is {\it not} cash:

\enumsentence{\label{hasInstrumentNotCashCardHobbs}
\hspace{5pt}$\neg$\hspace{1pt}{\tt has-instrument}\hspace{1pt}({\tt epjcash}\hspace{1pt}, {\tt card}\hspace{1pt}) $\wedge$ $\neg$\hspace{1pt}{\tt has-instrument}\hspace{1pt}({\tt epjcard}\hspace{1pt}, {\tt cash}\hspace{1pt})
}

\noindent Finally, from (\ref{hasInstrumentCashCardHobbs}) and (\ref{hasInstrumentNotCashCardHobbs}) we want to deduct that if {\tt epjcash} really exists then {\tt epjcard} does not and vice versa, i.e., that the two eventualities are connected by the {\tt not} predicate. In this paper, this is achieved by adding an inference rule implementing the pragmatic implicature explained above. In particular, this paper stipulates that if two eventualities refer to the same action or state and they share the same thematic roles with the same value except at least one, for which it is instead asserted that one of the two eventualities has a value on that thematic role while the other eventuality does not, then it is inferred that the two eventualities hold for the {\tt not} predicate. This inference is enforced in Hobbs's framework by the following implication:

\enumsentence{\label{notFromNegThematicRolesHobbs}
\hspace{4pt}$\forall_{e1,\hspace{1pt} e2,\hspace{1pt} AoS,\hspace{1pt} TR,\hspace{1pt} v}$[\hspace{1pt}({\tt AoS}\hspace{1pt}({\tt e1}) $\wedge$ {\tt AoS}\hspace{1pt}({\tt e2}) $\wedge$ {\tt TR}\hspace{1pt}({\tt e1}, {\tt v}) $\wedge$ $\neg$\hspace{1pt}{\tt TR}\hspace{1pt}({\tt e2}, {\tt v}) $\wedge$ ({\tt e1}\hspace{-2pt}$\neq${\tt e2}) $\wedge$\\
\mbox{}\hspace{65pt}$\forall_{\tt TR'\hspace{-1pt},\hspace{1pt} v'}$[(({\tt TR'}\hspace{-2pt}$\neq${\tt TR}) $\wedge$ {\tt TR'}({\tt e1}, {\tt v'}\hspace{-1pt})) $\rightarrow$ {\tt TR'}({\tt e2}, {\tt v'}\hspace{-1pt})] $\wedge$\\
\mbox{}\hspace{65pt}$\forall_{\tt TR'\hspace{-1pt},\hspace{1pt} v'}$[(({\tt TR'}\hspace{-2pt}$\neq${\tt TR}) $\wedge$ {\tt TR'}({\tt e2}, {\tt v'}\hspace{-1pt})) $\rightarrow$ {\tt TR'}({\tt e1}, {\tt v'}\hspace{-1pt})]\hspace{2pt})\hspace{1pt}$\rightarrow$\hspace{1pt}{\tt not}\hspace{1pt}({\tt e1}, {\tt e2})\hspace{1pt}]
}

\noindent Assuming that 
{\tt epjcash} and {\tt epjcard} are both true for the predicate {\tt Pay} and that they share the same thematic roles with the same value but {\tt has-instrument}, for which instead (\ref{hasInstrumentCashCardHobbs}), (\ref{payCashNotCardAndCardNotCashHobbs}), and (\ref{hasInstrumentNotCashCardHobbs}) hold, the implication in (\ref{notFromNegThematicRolesHobbs}) infers that \hspace{1pt}{\tt not}\hspace{1pt}({\tt epjcash}, {\tt epjcard})\hspace{1pt} holds. The next subsection will show the SPARQL rules corresponding to the implications in 
(\ref{payCashNotCardAndCardNotCashHobbs}) and (\ref{notFromNegThematicRolesHobbs}); section \ref{DeonticModalitiesInRDFsAndSPARQL} below, on the other hand, will show how contradictions and conflicts are inferred through these SPARQL rules on RDF knowledge graphs representing the sentences in (\ref{conflictingobligationsPartial}) and  
(\ref{conflictingFactsPartialOnJohn}).

\vspace{5pt}
\subsection{Encoding (Gordon and Hobbs, 2017) in RDFs and SPARQL}\label{RepresentingHobbsInRDFsAndSPARQL}
\vspace{-5pt}

\noindent So far, this section illustrated the portion of  \cite{Gordon-Hobbs:17}'s  framework needed for our formalization. This subsection explains how this portion is encoded in RDFs and SPARQL rules within the proposed computational ontology. The SPARQL rules shown below are executable through the lightweight reasoner described above in subsection \ref{LightweightAutomatedReasoner}.

All predicates seen in the previous subsection can be directly encoded in RDFs. For instance, formula (\ref{JohnLeavesAndPaysInHobbs}) is directly implementable in RDFs by mapping the predicates {\tt Leave} and {\tt Rexist} into homonym RDFs classes, {\tt has-agent} into an homonym RDF property, and {\tt elj} and {\tt John} into homonym RDF individuals. Furthermore, in order to state that all leaving actions are eventualities, {\tt Rexist} is a modality, and {\tt has-agent} is a thematic role, the class {\tt Leave} is asserted as individual of an additional class {\tt Eventuality}, the class {\tt Rexist} as individual of an additional class {\tt Modality}, and the property {\tt has-agent} as individual of an additional class {\tt ThematicRole}. Formula (\ref{JohnLeavesAndPaysInHobbs}.a) is then encoded in RDFs as in (\ref{JohnLeavesInRDFs}):

\enumsentence{\label{JohnLeavesInRDFs}
\mbox{}\hspace{5pt}{\tt :Eventuality}\hspace{3pt}{\tt a}\hspace{3pt} {\tt rdfs:Class.} {\tt :ThematicRole}\hspace{3pt}{\tt a}\hspace{3pt} {\tt rdfs:Class.}\\
\mbox{}\hspace{5pt}{\tt :Modality}\hspace{3pt}{\tt a}\hspace{3pt} {\tt rdfs:Class.} {\tt :Rexist}\hspace{3pt}{\tt a}\hspace{3pt} {\tt rdfs:Class,:Modality.}\\[2pt]
\mbox{}\hspace{7pt}{\tt soa:Leave}\hspace{3pt}{\tt a}\hspace{3pt} {\tt rdfs:Class,:Eventuality.}\\
\mbox{}\hspace{7pt}{\tt soa:has-agent} \hspace{3pt}{\tt a}\hspace{3pt} {\tt rdf:Property,:ThematicRole.}\\[2pt]
\mbox{}\hspace{7pt}{\tt soa:elj} \hspace{3pt}{\tt a}\hspace{3pt} {\tt :Rexist,soa:Leave; soa:has-agent}\hspace{3pt} {\tt soa:John.}
}
\vspace{2pt}

\noindent (\ref{JohnLeavesAndPaysInHobbs}.b) may be similarly represented by introducing further RDFs class and properties to encode the predicates {\tt Pay}, {\tt has-object}, and {\tt has-instrument}. The reader may find the RDFs representation of (\ref{JohnLeavesAndPaysInHobbs}.b) in the GitHub repository associated with this paper.

\vspace{2pt}
\subsubsection{\hspace{-1pt}Representing {\tt not}, {\tt and}\hspace{1pt}, {\tt or} and $\neg$\hspace{1pt}, $\wedge$\hspace{1pt}, \hspace{-1pt}$\vee$ in RDF: the level of the eventualities and the level of the statements}
\vspace{2pt}

\noindent The content of this subsubsection is crucial to understand the whole architecture of the proposed computational ontology; therefore, we invite the reader to pay particular attention.

As stated at the beginning of subsection \ref{GordonHobbsNegationDisjunctionConjunction} above, {\tt not}, {\tt and}, and {\tt or} do {\it not} correspond to the standard boolean connectives $\neg$\hspace{1pt}, $\wedge$\hspace{1pt}, and $\vee$. The former are first-order predicates that can be applied to first-order individuals and that denote relations among eventualities; the latter are operators that can transform statements into new statements and that denote functions from truth values to other truth values. 

Therefore, the two triples of constructs respectively belong to two different levels, which we will call the {\it level of the eventualities} and the {\it  level of the statements}. The bi-implications in (\ref{GordonHobbsNotAndOrAxioms}) stipulate a 1:1 correspondence between the two levels for what concern the {\tt Rexist} modality.

Representing the level of the eventualities in RDFs is rather straightforward as already exemplified in (\ref{JohnLeavesInRDFs}). The binary predicate {\tt not} may be also represented via an homonym RDF property connecting two eventualities:

\enumsentence{\label{notdeclaration}\tt
\mbox{}:not a rdf:Property; rdfs:domain :Eventuality; rdfs:range :Eventuality.
}

\noindent Conversely, it is not so immediate to encode the bi-implication in (\ref{GordonHobbsNotAndOrAxioms}.a) because RDFs does not include standard negation, i.e., the operator $\neg$, used in (\ref{GordonHobbsNotAndOrAxioms}.a) to relate the truth values of the {\tt Rexist} predicate applied to the two eventualities connected by {\tt not}. 

RDFs's semantics adheres to the Open World Assumption: if an RDF triple is not included in the knowledge graph, it does not mean that the triple is false, it simply means that it is {\it unknown} whether it is true or false.

Therefore, in order to encode (\ref{GordonHobbsNotAndOrAxioms}.a) in RDFs and SPARQL we decided to again use reification. However, not the same reification use in \cite{Gordon-Hobbs:17}'s framework but rather the reification defined in 
RDF v1.1\footnote{\url{https://www.w3.org/TR/rdf11-mt/\#reification}}. RDF v1.1 vocabulary includes the class  {\tt rdf:Statement} and the three RDF properties {\tt rdf:subject}, {\tt rdf:predicate}, and {\tt rdf:object} by means of which it is possible to reify the triples of the knowledge graph. For example, the RDF triple ``{\tt soa:elj\hspace{-2pt} a\hspace{-2pt} :Rexist}'' from
(\ref{JohnLeavesInRDFs}) might be reified into the following anonymous individual:

\enumsentence{\label{RDFreificationExample}\tt
[a rdf:Statement;\\
\mbox{}\hspace{5pt}rdf:subject soa:elj; rdf:predicate rdf:type; rdf:object :Rexist]
}

\noindent This anonymous individual explicitly refers to the RDF triple and it can be used to assert meta-properties such as the triple' author or the date in which it has been created.

One of the main research insights of this paper is to use RDF reification to represent the fact that some triples are true or false; in other words, {\it the truth value of an RDF triple is seen as a meta-property of that triple}. To implement so in RDFs, the following classes are added to the proposed computational ontology; these classes encode ``the level of the statements'' mentioned above.

\enumsentence{\label{truefalseholdRDFsClasses}\tt
\hspace{5pt}:statement a rdfs:Class; rdfs:subClassOf rdf:Statement.\\
\mbox{}\hspace{5pt}:true a rdfs:Class; rdfs:subClassOf :statement.\\
\mbox{}\hspace{5pt}:false a rdfs:Class; rdfs:subClassOf :statement.\\
\mbox{}\hspace{5pt}:hold a rdfs:Class; rdfs:subClassOf :statement.
}

\noindent Now, in order to represent that the triple ``{\tt soa:elj\hspace{-2pt} a\hspace{-2pt} :Rexist}'' holds false in the state of affairs, the anonymous individual in (\ref{ExampleFalse}) is added to the knowledge graph. (\ref{ExampleFalse}) reads: ``the fact that the triple {\tt soa:elj\hspace{-2pt} a\hspace{-2pt} :Rexist} is false holds in the state of affairs''. In Hobbs' framework, (\ref{ExampleFalse}) corresponds to $\neg$\hspace{1pt}{\tt Rexist}({\tt elj}\hspace{1pt})\hspace{1pt}.

\enumsentence{\label{ExampleFalse}\tt
[a :false,:hold;\\\mbox{}\hspace{5pt}rdf:subject soa:elj; rdf:predicate rdf:type; rdf:object :Rexist]
}

\noindent On the other hand, the fact that the triple ``{\tt soa:elj\hspace{-2pt} a\hspace{-2pt} :Rexist}'' holds true is represented as follows:

\enumsentence{\label{ExampleTrue}\tt
[a :true,:hold;\\
\mbox{}\hspace{5pt}rdf:subject soa:elj; rdf:predicate rdf:type; rdf:object :Rexist]
}

\noindent (\ref{ExampleTrue}) corresponds to {\tt Rexist}({\tt elj}\hspace{1pt}) in Hobbs' framework; given this correspondence, for practical reasons we will never assert when the reification of an RDF triple holds true in the state of affairs (although this could be derived through the SPARQL rules, as it will be exemplified below). Instead, we will simply assert the triple itself.

Conversely, a triple equivalent to (\ref{ExampleFalse}) cannot be encoded in RDF, because, as explained earlier, the RDF vocabulary does not include operators to represent standard negation. Therefore, whenever we will need to state that an RDF triple is {\it not} true in the state of affairs, we will state that its reification holds false.

Having introduced the RDFs classes in (\ref{truefalseholdRDFsClasses}), it is now possible to implement the bi-implication in (\ref{GordonHobbsNotAndOrAxioms}.a). This is encoded into the two SPARQL rules in (\ref{SPARQLnot1}) and (\ref{SPARQLnot2}). In (\ref{SPARQLnot1}), if the triples ``{\tt ?e\hspace{-2pt} :not\hspace{-2pt} ?ne}'' and ``{\tt ?e\hspace{-2pt} a\hspace{-2pt} :Rexist}'' belong to the knowledge graph, the rule creates a new anonymous individual therein stating that the fact that {\tt ?ne} really exists holds false; note that the {\tt WHERE} clause of this rule includes a {\tt NOT\hspace{-2pt} EXISTS} clause to prevent infinite loops, as explained above in (\ref{SPARQLsampleInfiniteLoopRule}). On the other hand, in (\ref{SPARQLnot2}), if the fact that {\tt ?e} really exists holds false, the rule asserts that {\tt ?ne} really exists, i.e., it adds the triple ``{\tt ?ne a\hspace{4pt}:Rexist}'' to the knowledge graph. In both rules, the {\tt UNION} operator is needed for the rules to trigger both when the subject of {\tt not} really exists or when its object does.

\enumsentence{\label{SPARQLnot1}
\begin{minipage}[t]{450pt}\tt
\mbox{}\hspace{0pt}[a :InferenceRule; :has-sparql-code\hspace{-3pt} """\\
\mbox{}\hspace{16pt}CONSTRUCT\{\hspace{-1pt}[a :false,:hold;\\
\mbox{}\hspace{40pt}rdf:subject\hspace{-1pt} ?ne;\hspace{-3pt} rdf:predicate\hspace{-1pt} rdf:type;\hspace{-3pt} rdf:object\hspace{-3pt} :Rexist]\hspace{-1pt}\}\\
\mbox{}\hspace{16pt}WHERE\{\hspace{-1pt}\{?e :not ?ne\}UNION\{?ne :not ?e\}
?e rdf:type :Rexist.\\
\mbox{}\hspace{54pt}NOT EXISTS\{\hspace{-1pt}?f a :false,:hold; rdf:subject\hspace{-1pt} ?ne;\\
\mbox{}\hspace{100pt}rdf:predicate\hspace{-1pt} rdf:type;\hspace{-3pt} rdf:object\hspace{-3pt} :Rexist\hspace{-1pt}\}\}"""]\hspace{-1pt}.
\end{minipage}
}

\vspace{0pt}
\enumsentence{\label{SPARQLnot2}
\begin{minipage}[t]{450pt}\tt
\mbox{}\hspace{0pt}[a :InferenceRule; :has-sparql-code\hspace{-3pt} """\\
\mbox{}\hspace{12pt}CONSTRUCT\{\hspace{-1pt}?ne a :Rexist\}\\
\mbox{}\hspace{12pt}WHERE\{\hspace{-1pt}\{?e :not ?ne\}UNION\{?ne :not ?e\} ?r a :false,:hold;\\
\mbox{}\hspace{28pt}rdf:subject\hspace{-1pt} ?e;\hspace{-1pt} rdf:predicate\hspace{-1pt} rdf:type;\hspace{-1pt} rdf:object\hspace{-1pt} :Rexist\}"""]\hspace{-1pt}.
\end{minipage}
}
\vspace{5pt}

\noindent Finally, as already pointed out above, it is unpractical to create anonymous individuals that reifies the triples in the knowledge graph and to assert that they hold true (also because, by iteratively re-doing so, as our lightweight reasoner does, the computation will loop infinitely). Conversely, as it will be exemplified below, in cases where the knowledge graph contains such an anonymous individual but {\it not} the corresponding RDF triple, it is convenient to add the latter to the knowledge graph in order to enable further inferences at the level of the eventualities\footnote{Indeed, besides adding the RDF triple we could also remove the anonymous individual that reifies it and holds true. In other words, we could {\it replace} the latter with the former in the knowledge graph. In order to do so, the rule must use {\tt INSERT} and {\tt DELETE} in place of {\tt CONSTRUCT}.}. The following SPARQL rule implements the described implication: from an anonymous individual that holds true, i.e., that belongs to the classes {\tt true} and {\tt hold}, to the triple that is reified by that individual:

\vspace{2pt}
\enumsentence{\label{SPARQLholdtrue}
\begin{minipage}[t]{450pt}\tt
\mbox{}\hspace{-4pt}[a :InferenceRule; :has-sparql-code\hspace{-3pt} """CONSTRUCT\{\hspace{-1pt}?s ?p ?o\}\\
\mbox{}\hspace{30pt}WHERE\{\hspace{-1pt}?r\hspace{-2pt} a\hspace{-2pt} :true,\hspace{-1pt}:hold;\hspace{-3pt} rdf:subject\hspace{-3pt} ?s;\\
\mbox{}\hspace{62pt}rdf:predicate\hspace{-2pt} ?p;\hspace{-3pt} rdf:object\hspace{-3pt} ?o\hspace{-1pt}\}\hspace{-1pt}"\hspace{-1pt}"\hspace{-1pt}"\hspace{-1pt}].
\end{minipage}
}
\vspace{5pt}

\subsubsection{Representing conjunction and disjunction of eventualities in RDFs and SPARQL}
\vspace{2pt}

\noindent This subsubsection completes the previous one by illustrating how the predicates {\tt and} and {\tt or} are represented in the proposed computational ontology. Representing these predicates in RDFs is more complex than representing the predicate {\tt not} because they are {\it ternary} predicates. In line with what it is standardly done in the Semantic Web to encode n-ary relations\footnote{See \url{https://www.w3.org/TR/swbp-n-aryRelations}}, {\tt and}\hspace{1pt}({\tt ea}, {\tt e1}, {\tt e2}) is encoded by taking {\tt ea} as subject of two different RDF properties {\tt and1} and {\tt and2}, respectively connected with {\tt e1} and {\tt e2}; {\tt or}\hspace{1pt}({\tt eo}, {\tt e1}, {\tt e2}) is similarly encoded.

\begin{center}\tt
soa:ea :and1 soa:e1.\hspace{20pt}
soa:ea :and2 soa:e2.
\end{center}

\noindent Let's now see how the two bi-implications in (\ref{GordonHobbsNotAndOrAxioms}.b) and (\ref{GordonHobbsNotAndOrAxioms}.c) are implemented as SPARQL rules.

Representing the bi-implications (\ref{GordonHobbsNotAndOrAxioms}.b) is straightforward, as RDF supports the boolean connective ``$\wedge$'': all triples in a RDF knowledge graph are intended to hold together; in other words, the knowledge graph itself is the conjunction of all its triples. Therefore, (\ref{GordonHobbsNotAndOrAxioms}.b) is simply encoded via the two SPARQL rules shown in (\ref{SPARQLand1}) and (\ref{SPARQLand2}):

\enumsentence{\label{SPARQLand1}
\begin{minipage}[t]{450pt}\tt
\mbox{}\hspace{0pt}[a :InferenceRule; :has-sparql-code\hspace{-3pt} """\\
\mbox{}\hspace{20pt}CONSTRUCT\{?e1 a :Rexist. ?e2 a :Rexist.\}\\
\mbox{}\hspace{20pt}WHERE\{?ea :and1 ?e1. ?ea :and2 ?e2. ?ea a :Rexist\}"""]\hspace{-1pt}.
\end{minipage}
}

\enumsentence{\label{SPARQLand2}
\begin{minipage}[t]{450pt}\tt
\mbox{}\hspace{0pt}[a :InferenceRule; :has-sparql-code\hspace{-3pt} """\\
\mbox{}\hspace{20pt}CONSTRUCT\{?ea a :Rexist.\}\\
\mbox{}\hspace{20pt}WHERE\{?ea\hspace{-1pt} :and1\hspace{-1pt} ?e1.\hspace{-3pt} ?ea\hspace{-1pt} :and2 ?e2.\\
\mbox{}\hspace{52pt}?e1\hspace{-1pt} a\hspace{-1pt} :Rexist\hspace{-3pt}.\hspace{-1pt} ?e2\hspace{-1pt} a\hspace{-1pt} :Rexist\hspace{-1pt}\}"""]\hspace{-1pt}.
\end{minipage}
}
\vspace{2pt}

\noindent On the other hand, the encoding in RDF and SPARQL of the bi-implications in (\ref{GordonHobbsNotAndOrAxioms}.c) is as problematic as the encoding of the bi-implications in  (\ref{GordonHobbsNotAndOrAxioms}.a), because RDF vocabulary does not include operators that implement standard boolean disjunction ($\vee$) either. Therefore, similarly to what has been done in the previous subsubsection to encode the boolean operator $\neg$\hspace{1pt}, a new RDF property {\tt disjunction} is introduced to encode the boolean operator $\vee$\hspace{1pt}:

\enumsentence{\label{disjunctionDeclaration}\tt
\mbox{}\hspace{5pt}:disjunction rdf:type rdf:Property;\\
\mbox{}\hspace{25pt}rdfs:domain :statement; rdfs:range :statement.
}

\noindent ({\tt Rexist}({\tt e1}) $\vee$ {\tt Rexist}({\tt e2})) is now represented as:

\enumsentence{\label{ExampleDisjunction1}\tt
\mbox{}\hspace{2pt}[a :true; rdf:subject :e1; rdf:predicate rdf:type; rdf:object :Rexist]\\
\mbox{}\hspace{3pt}:disjunction;\\
\mbox{}\hspace{2pt}[a :true; rdf:subject :e2; rdf:predicate rdf:type; rdf:object :Rexist].
}

\noindent Note that (\ref{ExampleDisjunction1}) embeds two anonymous individuals in its {\tt rdf:subject} and {\tt rdf:object}. However, contrary to the anonymous individual in (\ref{ExampleTrue}), the ones in (\ref{ExampleDisjunction1}) belong to the class {\tt true} but not to the class {\tt hold}. In fact, if 
({\tt Rexist}({\tt e1}) $\vee$ {\tt Rexist}({\tt e2})) holds true, it is unknown whether {\tt Rexist}({\tt e1}) or {\tt Rexist}({\tt e2}) also hold true; it is only known that at least one of the two holds true but it is unknown which one(s). 

It should be now clear that ({\tt Rexist}({\tt e1}) $\vee$ $\neg$\hspace{1pt}{\tt Rexist}({\tt e2})) is instead represented as follows:

\enumsentence{\label{ExampleDisjunction2}\tt
\mbox{}\hspace{2pt}[a :true; rdf:subject :e1; rdf:predicate rdf:type; rdf:object :Rexist]\\
\mbox{}\hspace{3pt}:disjunction;\\
\mbox{}\hspace{2pt}[a :false; rdf:subject :e2; rdf:predicate rdf:type; rdf:object :Rexist].
}

\noindent The single difference between (\ref{ExampleDisjunction1}) and (\ref{ExampleDisjunction2}) is that, in the former, the anonymous individual embedded in the {\tt rdf:object}  belongs to the class {\tt false} rather than to the class {\tt true}.

Now the bi-implications in (\ref{GordonHobbsNotAndOrAxioms}.c) can be encoded through the SPARQL rules shown in (\ref{SPARQLor1}) and (\ref{SPARQLor2}):

\enumsentence{\label{SPARQLor1}
\begin{minipage}[t]{450pt}\tt
\mbox{}\hspace{0pt}[a :InferenceRule; :has-sparql-code\hspace{-3pt} """\\
\mbox{}\hspace{12pt}CONSTRUCT\{[a :true; rdf:subject ?e1; rdf:predicate rdf:type;\\
\mbox{}\hspace{68pt}rdf:object :Rexist] :disjunction [a :true; rdf:subject ?e2;\\
\mbox{}\hspace{68pt}rdf:predicate rdf:type; rdf:object :Rexist]\}\\
\mbox{}\hspace{12.5pt}WHERE\{?eo :or1 ?e1. ?eo :or2 ?e2. ?eo a :Rexist.\\
\mbox{}\hspace{28pt}NOT EXISTS\{\hspace{-1pt}?r1 a :true; rdf:predicate rdf:type; rdf:object :Rexist.\\
\mbox{}\hspace{85pt}?r2 a :true; rdf:predicate rdf:type; rdf:object :Rexist.\\
\mbox{}\hspace{84pt}\{?r1 rdf:subject ?e1. ?r2 rdf:subject ?e2.\}UNION\\
\mbox{}\hspace{84pt}\{?r1 rdf:subject ?e2. ?r2 rdf:subject ?e1.\}\}\}"""]\hspace{-1pt}.
\end{minipage}
}
\vspace{2pt}

\enumsentence{\label{SPARQLor2}
\begin{minipage}[t]{450pt}\tt
\mbox{}\hspace{0pt}[a :InferenceRule; :has-sparql-code\hspace{-3pt} """\\
\mbox{}\hspace{10pt}CONSTRUCT\{?eo a :Rexist\}\\
\mbox{}\hspace{10pt}WHERE\{\{?eo \hspace{-1pt}:or1 ?e1. ?eo \hspace{-1pt}:or2 ?e2\}UNION\{?eo \hspace{-1pt}:or1 ?e2. ?eo \hspace{-1pt}:or2 ?e1\}\\ 
\mbox{}\hspace{46pt}\{?e1 rdf:type :Rexist\}UNION\{?e2 rdf:type :Rexist\}\}"""].
\end{minipage}
}
\vspace{2pt}

\noindent In (\ref{SPARQLor1}) and (\ref{SPARQLor2}), the {\tt UNION} operator is again needed: the rules trigger both when the objects of {\tt or1} and {\tt or2} are respectively embedded in the subject and the object of {\tt disjunction} {\it and} the other way round: when they are respectively embedded in the object and the subject of {\tt disjunction}.

\subsubsection{Abstract and instantiated eventualities}\label{AbstractEventualitiesRDFsAndSPARQL}
\vspace{2pt}

\noindent In subsection \ref{AbstractEventualitiesHobbs} it has been explained that this paper needs to import from \cite{Gordon-Hobbs:17} the distinction between abstract eventualities and their instantiations in order to properly represent and reason with partial conflicts. On the other hand, dealing with abstract eventualities and their instantiations {\it in general} is rather complex and it would deserve a paper on its own, also because abstract eventualities are intimately related to the representation of sets and natural
language quantification, as it will be discussed in section \ref{FutureWorks} below, devoted to future works. Therefore, the proposed computational ontology will not encode all predicates and axioms defined in \cite{Gordon-Hobbs:17} to deal with abstract eventualities, but only those strictly needed to enable the right inferences on the partial conflicts exemplified in this paper, namely the SPARQL rules corresponding to (\ref{payCashNotCardAndCardNotCashHobbs}) and (\ref{notFromNegThematicRolesHobbs}) above. These are shown in (\ref{payCashNotCardAndCardNotCashSPARQL}) and (\ref{notFromNegThematicRolesRDFsAndSPARQL}) respectively.

\enumsentence{\label{payCashNotCardAndCardNotCashSPARQL}
\begin{minipage}[t]{450pt}\tt
\mbox{}\hspace{0pt}[a :InferenceRule; :has-sparql-code\hspace{-3pt} """\\
\mbox{}\hspace{20pt}CONSTRUCT\{[a :false,:hold; rdf:subject ?e;\\
\mbox{}\hspace{75pt}rdf:predicate soa:has-instrument; rdf:object soa:card]\}\\
\mbox{}\hspace{20pt}WHERE\{?e a soa:Pay. ?e soa:has-instrument soa:cash\\
\mbox{}\hspace{30pt}NOT EXISTS\{?r a :false,:hold; rdf:subject ?e;\\
\mbox{}\hspace{40pt}rdf:predicate soa:has-instrument; rdf:object soa:card\}\}"""]\hspace{-1pt}.
\end{minipage}

\vspace{8pt}
\begin{minipage}[t]{450pt}\tt
\mbox{}\hspace{0pt}[a :InferenceRule; :has-sparql-code\hspace{-3pt} """\\
\mbox{}\hspace{20pt}CONSTRUCT\{[a :false,:hold; rdf:subject ?e;\\
\mbox{}\hspace{75pt}rdf:predicate soa:has-instrument; rdf:object soa:cash]\}\\
\mbox{}\hspace{20pt}WHERE\{?e a soa:Pay. ?e soa:has-instrument soa:card\\
\mbox{}\hspace{30pt}NOT EXISTS\{?r a :false,:hold; rdf:subject ?e;\\
\mbox{}\hspace{40pt}rdf:predicate soa:has-instrument; rdf:object soa:cash\}\}"""]\hspace{-1pt}.
\end{minipage}
}
\vspace{2pt}

\enumsentence{\label{notFromNegThematicRolesRDFsAndSPARQL}
\begin{minipage}[t]{450pt}\tt
\mbox{}\hspace{0pt}[a\hspace{-2pt} :InferenceRule; :has-sparql-code\hspace{-3pt} """\\
\mbox{}\hspace{18pt}CONSTRUCT\{?e1 :not ?e2\}\\
\mbox{}\hspace{18pt}WHERE\{?e1 a ?c. ?e2 a ?c. ?c a :Eventuality. FILTER(?e1!=?e2)\\
\mbox{}\hspace{34pt}?trn a\hspace{-1pt} :ThematicRole. ?e1 ?trn ?tv. ?r a :false,:hold;\\
\mbox{}\hspace{35pt}rdf:subject ?e2; rdf:predicate ?trn; rdf:object ?vn.\\
\mbox{}\hspace{35pt}NOT EXISTS\{?tr a :ThematicRole. FILTER(?tr!=?trn) ?e1 ?tr ?tv1.\\
\mbox{}\hspace{35pt}NOT EXISTS\{?e2 ?tr ?tv2\}\} NOT EXISTS\{?tr a :ThematicRole.\\
\mbox{}\hspace{35pt}FILTER(?tr!=?trn) ?e2 ?tr ?tv2. NOT EXISTS\{?e1 ?tr ?tv1\}\}\\
\mbox{}\hspace{35pt}NOT EXISTS\{?tr a :ThematicRole. FILTER(?tr!=?trn)\\
\mbox{}\hspace{35pt}?e1 ?tr ?tv1. ?e2 ?tr ?tv2. FILTER(?tv1!=?tv2)\}\}"""]\hspace{-1pt}.
\end{minipage}
}
\vspace{2pt}

\subsubsection{Representing and inferring contradictions}\label{representingAndInferringContradictions}
\vspace{2pt}

\noindent The final ingredient to complete the encoding in RDFs and SPARQL of the portion of \cite{Gordon-Hobbs:17}'s framework used in this paper is an explicit representation of {\it contradictions}, usually associated in classical logic with the symbol ``$\bot$''.

The proposed computational ontology includes a new RDF property  to explicitly represent contradictions: ``{\tt is-in-contradiction-with}'', defined as follows:

\enumsentence{\label{is-in-contradiction-withDefinition}\tt
\mbox{}\hspace{5pt}:is-in-contradiction-with a rdf:Property;\\
\mbox{}\hspace{25pt}rdfs:domain :statement; rdfs:range :statement.
}

\noindent We can now define SPARQL rules to connect, through this property, reifications of triples that hold both true and false in the same state of affairs. For instance, the rule in (\ref{SPARQLcontradictionRexistNotRexist}) triggers when an eventuality {\tt e} both really exists and it does not, i.e., in \cite{Gordon-Hobbs:17}'s terms, when \hspace{1pt}{\tt Rexist}({\tt e}) $\wedge$ $\neg$\hspace{1pt}{\tt Rexist}({\tt e})\hspace{1pt} holds true in the state of affairs; if so, the {\tt CONSTRUCT} clause of the rule creates a new RDF triple stating that the fact that the triple ``{\tt e\hspace{-2pt} a\hspace{-2pt} :Rexist}'' holds true in the state of affairs is inconsistent with the fact that it holds false.

\enumsentence{\label{SPARQLcontradictionRexistNotRexist}
\begin{minipage}[t]{550pt}\tt
[a :InferenceRule; :has-sparql-code """\\
\mbox{}\hspace{13pt}CONSTRUCT\{\hspace{-1pt}[a :true,:hold;\\
\mbox{}\hspace{68pt}rdf:subject ?e; rdf\hspace{-1pt}:\hspace{-1pt}predicate\hspace{-2pt} rdf\hspace{-1pt}:\hspace{-1pt}type;\hspace{-2pt} rdf\hspace{-1pt}:\hspace{-1pt}object\hspace{-2pt} :Rexist\hspace{-1pt}]\\
\mbox{}\hspace{63pt}:is-in-contradiction-with ?r\}\\
\mbox{}\hspace{14pt}WHERE\{?e\hspace{-1pt} a\hspace{-1pt} :\hspace{-1pt}Rexist\hspace{-1pt}.\hspace{-2pt} ?r\hspace{-1pt} a\hspace{-1pt} :false,\hspace{-1pt}:hold;\\ 
\mbox{}\hspace{30pt}rdf:subject\hspace{-1pt} ?e; rdf:predicate rdf:type; rdf:object :Rexist.\\
\mbox{}\hspace{31.5pt}NOT\hspace{-3pt} EXISTS\hspace{-1pt}\{\{?t :is-in-contradiction-with ?r\} UNION\\
\mbox{}\hspace{38pt}\{?r :is-in-contradiction-with ?t\} ?t a :true,:hold;\\
\mbox{}\hspace{30pt}rdf:subject\hspace{-2pt} ?e;\hspace{-2pt} rdf:predicate\hspace{-2pt} rdf:type;\hspace{-2pt} rdf:object\hspace{-2pt} :Rexist\hspace{-1pt}\}\hspace{-1pt}\}\hspace{-1pt}"\hspace{-1pt}"\hspace{-1pt}"\hspace{-1pt}]\hspace{-1pt}.
\end{minipage}
}
\vspace{2pt}

\noindent Let's now show a simple example, i.e., the RDF triples in 
(\ref{ExamplesContradictions1}.a) and (\ref{ExamplesContradictions1}.b), which respectively formalize the sentences ``John leaves'' and ``John does not leave''.

\enumsentence{\label{ExamplesContradictions1}
\begin{enumerate}
    \renewcommand{\labelenumi}
    {\alph{enumi}.}

    \item\noindent {\tt soa:elj a :Rexist,\hspace{-4pt} soa:Leave; soa:has-agent soa:John.}
    
    \item\noindent {\tt soa:enlj a :Rexist; :not soa:elj.}
\end{enumerate}
}

\noindent From (\ref{ExamplesContradictions1}.b), the SPARQL rule in (\ref{SPARQLnot1}) infers the anonymous individual in (\ref{ExamplesContradictions1inferences}.a); from this individual and the triple ``{\tt soa:elj a :Rexist}'' in (\ref{ExamplesContradictions1}.a), the SPARQL rule in (\ref{SPARQLcontradictionRexistNotRexist}) infers (\ref{ExamplesContradictions1inferences}.b), which states that the fact that the triple ``{\tt soa:elj a :Rexist}'' holds true is in contradiction with the fact that it holds false.

\enumsentence{\label{ExamplesContradictions1inferences}
\begin{enumerate}
    \renewcommand{\labelenumi}
    {\alph{enumi}.}

    \item\noindent {\tt [a :false,:hold;\\
    \mbox{}\hspace{5pt}rdf:subject soa:elj; rdf:predicate rdf:type; rdf:object :Rexist]}
    \vspace{2pt}
    
    \item\noindent {\tt [a :true,:hold;\\
    \mbox{}\hspace{5pt}rdf:subject soa:elj; rdf:predicate rdf:type; rdf:object :Rexist]\\[2pt]
    \mbox{}\hspace{0pt}:is-in-contradiction-with\\[2pt]
    \mbox{}\hspace{0pt}[a :false,:hold;\\
    \mbox{}\hspace{5pt}rdf:subject soa:elj; rdf:predicate rdf:type; rdf:object :Rexist]}
\end{enumerate}
}

\noindent Consider, on the other hand, the two sentences in (\ref{conflictingFactsPartialOnJohn}) above, repeated again in 
(\ref{contradictionsPartialOnJohn}) for reader's convenience:

\enumsentence{\label{contradictionsPartialOnJohn}
\begin{enumerate}
    \renewcommand{\labelenumi}{\alph{enumi}.}
    \item John pays in cash.
    \vspace{5pt}
    \item John pays by card.
\end{enumerate}
}

\noindent If both eventualities associated with (\ref{contradictionsPartialOnJohn}.a-b) really exist, as shown in (\ref{JohnPayInCashAndCard}), a contradiction is again inferred.

\enumsentence{\label{JohnPayInCashAndCard}
{\tt soa:epjcash a soa:Pay; soa:has-agent soa:John;}\\[2pt]
\mbox{}\hspace{160pt}{\tt soa:has-instrument soa:cash.}\\[5pt]
{\tt soa:epjcard a soa:Pay; soa:has-agent soa:John;}\\[2pt]
\mbox{}\hspace{160pt}{\tt soa:has-instrument soa:card.}\\[5pt]
\mbox{}{\tt soa:epjcash a :Rexist.\hspace{3pt} soa:epjcard a :Rexist.}
}

\noindent Through the SPARQL rules in (\ref{payCashNotCardAndCardNotCashSPARQL}), the triples in (\ref{JohnPayInCashAndCardInference1}) are inferred from (\ref{JohnPayInCashAndCard}): the instrument of {\tt epjcash} is not card and, symmetrically, the instrument of {\tt epjcard} is not cash:

\enumsentence{\label{JohnPayInCashAndCardInference1}\tt
[a :false,:hold; rdf:subject epjcash;\\
\mbox{}\hspace{60pt}rdf:predicate soa:has-instrument; rdf:object soa:card]\\[5pt]
\mbox{}\hspace{0pt}[a :false,:hold; rdf:subject epjcard;\\
\mbox{}\hspace{60pt}rdf:predicate soa:has-instrument; rdf:object soa:cash]
}

\noindent Then, the SPARQL rule in (\ref{notFromNegThematicRolesRDFsAndSPARQL}) infer the triples ``{\tt epjcash :not epjcard}'' and ``{\tt epjcard :not epjcash}'' and, finally, the SPARQL rule in (\ref{SPARQLcontradictionRexistNotRexist}) infers {\it two} contradictions: neither {\tt epjcash} nor {\tt epjcard} can both really exist and not really exist. The first contradiction is encoded in RDF as in (\ref{JohnPayInCashAndCardInference2}); the second one is symmetric.

\enumsentence{\label{JohnPayInCashAndCardInference2}\tt [a :true,:hold;\\
\mbox{}\hspace{5pt}rdf:subject soa:epjcash; rdf:predicate rdf:type; rdf:object :Rexist]\\[2pt]
\mbox{}\hspace{0pt}:is-in-contradiction-with\\[2pt]
\mbox{}\hspace{0pt}[a :false,:hold;\\
\mbox{}\hspace{5pt}rdf:subject soa:epjcash; rdf:predicate rdf:type; rdf:object :Rexist]
}

\noindent The two inferences described step-by-step in this subsubsection are both available on the GitHub repository.

\subsubsection{Representing further inferences}\label{representingFurtherInferences}
\vspace{2pt}

\noindent This subsubsection discusses how and to what extent it is possible to add  SPARQL rules that enable further inferences at the level of the eventualities ({\tt not}, {\tt and}, and {\tt or}), at the level of the statements ($\neg$\hspace{1pt}, $\wedge$\hspace{1pt}, and $\vee$), or across these two levels. Although this discussion only concerns the {\tt Rexist} modality, it will be crucial to understand how and to what extent it is possible to model inferences on eventualities holding for the deontic modalities.

For example, (\ref{OrNotTheoremRexist}.a) above, which implements Disjunctive Syllogism at the level of the eventualities, may be encoded into the following SPARQL rule:

\enumsentence{\label{DisjunctiveSyllogismOnRexist}
\begin{minipage}[t]{450pt}\tt
\mbox{}\hspace{0pt}[a :InferenceRule; :has-sparql-code\hspace{-3pt} """\\
\mbox{}\hspace{16pt}CONSTRUCT\{?e2 a :Rexist\}\\
\mbox{}\hspace{16pt}WHERE\{?eo a :Rexist. ?en1 a :Rexist.\\
\mbox{}\hspace{51pt}\{?eo :or1 ?e1; :or2 ?e2\}UNION\{?eo :or1 ?e2; :or2 ?e1\}\\
\mbox{}\hspace{51pt}\{?e1 :not ?en1\}UNION\{?en1 :not ?e1\}\}"""]\hspace{-1pt}.
\end{minipage}
}

\noindent (\ref{DisjunctiveSyllogismOnRexist}) uses the operator {\tt UNION} for the rule to trigger on all four combinations: when the negated eventuality is either the one as object of {\tt or1} rather than {\tt or2} and when it is either the subject or the object of {\tt not}.

On the other hand, Disjunctive Syllogism at the level of the statements (implemented in \cite{Gordon-Hobbs:17}'s via (\ref{OrNotTheoremRexist}.b) above, may be encoded into the following SPARQL rule:

\enumsentence{\label{DisjunctiveSyllogismAtStatementLevel}
\begin{minipage}[t]{450pt}\tt
\mbox{}\hspace{-2pt}[a :InferenceRule; :has-sparql-code\hspace{-3pt} """\\
\mbox{}\hspace{14pt}CONSTRUCT\{?r2 a :hold\}\\
\mbox{}\hspace{14pt}WHERE\{\{\hspace{-1pt}?r1 :disjunction ?r2\}UNION\{?r2 :disjunction ?r1\}\\
\mbox{}\hspace{35pt}?r1\hspace{-1pt} a\hspace{-1pt} ?tvr1;\hspace{-1pt} rdf:subject\hspace{-1pt} ?s;\hspace{-1pt} rdf:predicate\hspace{-1pt} ?p;\hspace{-1pt} rdf:object\hspace{-1pt} ?o.\\
\mbox{}\hspace{35pt}?rn1\hspace{-1pt} a :\hspace{-1pt}hold,\hspace{-2pt}?tvrn1\hspace{-1pt};\hspace{-4pt} rdf\hspace{-1pt}:\hspace{-1pt}subject\hspace{-3pt} ?s\hspace{-1pt};\hspace{-4pt} rdf\hspace{-1pt}:\hspace{-1pt}predicate\hspace{-2pt} ?p\hspace{-1pt};\hspace{-4pt} rdf\hspace{-1pt}:\hspace{-1pt}object\hspace{-3pt} ?o\hspace{-1pt}.\\
\mbox{}\hspace{35pt}FILTER(((?tvr1=:true)\&\&(?tvrn1=:false))||\\
\mbox{}\hspace{77pt}((?tvr1=:false)\&\&(?tvrn1=:true)))\}"""]\hspace{-1pt}.
\end{minipage}
}
\vspace{2pt}

\noindent Similarly to (\ref{DisjunctiveSyllogismOnRexist}), the SPARQL rule in (\ref{DisjunctiveSyllogismAtStatementLevel}) uses the operators {\tt FILTER} and {\tt UNION} to trigger on all four possible combinations: when the first conjunct is either true or false while its negation holds and when the second conjunct is either true or false while its negation holds.

Let's now see a quick example involving (\ref{DisjunctiveSyllogismOnRexist}) and (\ref{DisjunctiveSyllogismAtStatementLevel}). Consider sentence (\ref{ExampleJohnLeavesOrJohnEatsAndDrink}.a), encoded in \cite{Gordon-Hobbs:17} as in (\ref{ExampleJohnLeavesOrJohnEatsAndDrink}.b) and then in RDFs as in (\ref{ExampleJohnLeavesOrJohnEatsAndDrink}.c):

\vspace{2pt}
\enumsentence{\label{ExampleJohnLeavesOrJohnEatsAndDrink}
\begin{enumerate}
    \renewcommand{\labelenumi}
    {\alph{enumi}.}
    \item\noindent John leaves or John eats and drinks.
    \vspace{5pt}

    \item\noindent {\tt Rexist}({\tt eo}) $\wedge$
    {\tt or}\hspace{1pt}({\tt eo}, {\tt elj}, {\tt ea}) $\wedge$ {\tt Leave}({\tt elj}) $\wedge$ {\tt and}\hspace{1pt}({\tt ea}, {\tt eej}, {\tt edj}) $\wedge$\\ 
    {\tt Eat}({\tt eej}) $\wedge$
    {\tt Drink}({\tt edj}) $\wedge$ {\tt has-agent}({\tt elj}, {\tt John})
    $\wedge$\\ 
    {\tt has-agent}({\tt eej}, {\tt John}) $\wedge$ {\tt has-agent}({\tt edj}, {\tt John})
    \vspace{5pt}

    \item\noindent {\tt soa:eo a :Rexist; :or1 soa:elj; :or2 soa:ea.\\
    soa:elj a soa:Leave; soa:has-agent soa:John.\\
    soa:ea :and1 soa:eej; :and2 soa:edj.\\
    soa:eej a soa:Eat; soa:has-agent soa:John.\\
    soa:edj a soa:Drink; soa:has-agent soa:John.}
\end{enumerate}
}
\vspace{2pt}

\noindent To the knowledge graph in (\ref{ExampleJohnLeavesOrJohnEatsAndDrink}.c), we add the triples in (\ref{ExampleJohnDoesNotLeave}.b), which encode sentence (\ref{ExampleJohnDoesNotLeave}.a).

\enumsentence{\label{ExampleJohnDoesNotLeave}
\begin{enumerate}
    \renewcommand{\labelenumi}
    {\alph{enumi}.}
    \item\noindent John does not leave.
    \vspace{3pt}
    \item\noindent {\tt soa:enlj a :Rexist; :not :elj.}
\end{enumerate}
}

\noindent Intuitively, from (\ref{ExampleJohnLeavesOrJohnEatsAndDrink}.a) and (\ref{ExampleJohnDoesNotLeave}.a), it can be inferred that John eats and John drinks, i.e., that the two corresponding eventualities, i.e., {\tt eej} and {\tt edj}, really exist. This is indeed achieved through the SPARQL rules seen above, from two different inference paths, each at one of the two levels.

At the level of the eventualities,
the rule in 
(\ref{DisjunctiveSyllogismOnRexist}) infers, from (\ref{ExampleJohnLeavesOrJohnEatsAndDrink}.c) and (\ref{ExampleJohnDoesNotLeave}.b), the triple ``{\tt soa:ea a :Rexist}''. Then, the rule in (\ref{SPARQLand1}) infers, from this triple and the triples ``{\tt soa:ea :and1 soa:eej; :and2 soa:edj}'' in (\ref{ExampleJohnLeavesOrJohnEatsAndDrink}.c), that also {\tt eej} and {\tt edj}, i.e., the fact that John eats and the fact that John drinks, really exist:

\enumsentence{\label{ExampleJohnLeavesOrJohnEatsAndDrinkFinalInferenceStep}
\hspace{3pt}{\tt soa:eej a :Rexist.}\hspace{8pt} {\tt soa:edj a :Rexist.}
}
\vspace{2pt}

\noindent At the level of the statements, the rules in (\ref{SPARQLnot1}) and (\ref{SPARQLor1}) transform (\ref{ExampleJohnLeavesOrJohnEatsAndDrink}.c) and (\ref{ExampleJohnDoesNotLeave}.b) into the equivalent RDF representation in (\ref{ExampleJohnLeavesOrJohnEatsAndDrinkFirstInferenceStepStatementsLevel}); the latter encodes that the fact that {\tt elj} really exists holds false and that the property {\tt disjunction} holds between the fact that {\tt elj} really exists and the fact that {\tt ea} really exists.

\enumsentence{\label{ExampleJohnLeavesOrJohnEatsAndDrinkFirstInferenceStepStatementsLevel}
{\tt [a :false, :hold;\\
\mbox{}\hspace{30pt}rdf:subject elj; rdf:predicate rdf:type; rdf:object :Rexist].}\\[5pt]
{\tt [a :true; rdf:subject elj;
rdf:predicate rdf:type; rdf:object :Rexist]\\
\mbox{}:disjunction\\
\mbox{}[a :true; rdf:subject ea;
rdf:predicate rdf:type; rdf:object :Rexist].}
}

\noindent Then, from the triples in (\ref{ExampleJohnLeavesOrJohnEatsAndDrinkFirstInferenceStepStatementsLevel}), the rule in (\ref{DisjunctiveSyllogismAtStatementLevel}) infers that the anonymous individual occurring as the object of {\tt disjunction} holds true in the state of affairs:

\enumsentence{\label{ExampleJohnLeavesOrJohnEatsAndDrinkSecondInferenceStepStatementsLevel}
{\tt [a :true, hold;\\
\mbox{}\hspace{30pt}rdf:subject ea; rdf:predicate rdf:type; rdf:object :Rexist].}
}

\noindent Then, the SPARQL rule in (\ref{SPARQLholdtrue}) adds the triple in (\ref{ExampleJohnLeavesOrJohnEatsAndDrinkThirdInferenceStep}), equivalent to (\ref{ExampleJohnLeavesOrJohnEatsAndDrinkSecondInferenceStepStatementsLevel}), to the knowledge graph, thus moving back to the level of the eventualities. Finally, from this triple, the SPARQL rule in (\ref{SPARQLand1}) infers again the two triples in (\ref{ExampleJohnLeavesOrJohnEatsAndDrinkFinalInferenceStep}).

\enumsentence{\label{ExampleJohnLeavesOrJohnEatsAndDrinkThirdInferenceStep}
\hspace{3pt}{\tt soa:ea a :Rexist.}
}

\noindent The example just shown is also available on the GitHub repository associated with this paper; the reader is invited to re-execute it locally and double-check the inferences. Furthermore, as already explained above, additional SPARQL rules could be added to implement, at both levels, other implications from classical logic, e.g., the distributivity laws on $\wedge$ and $\vee$ or De Morgan's laws. The reader is likewise invited to encode these additional rules as an exercise while executing them via the lightweight reasoner downloadable from the GitHub repository.

Nevertheless, it is important to understand that some implications that are valid in classical logic should {\it not} be encoded as SPARQL rules and added to the proposed computational ontology. The reason is that these rules would lead to knowledge graphs that cannot exist in reality or that would nonetheless be highly undesirable.
Knowledge graphs are {\it concrete} computational artefacts; therefore, certain implications, although logically valid, are incompatible or highly undesirable with respect to the physical limits of a computer.

An example is {\tt A}$\rightarrow$\hspace{1pt}({\tt A}\hspace{0.5pt}$\vee$\hspace{1pt}{\tt B}), known as ``Disjunction Introduction'' and stating that if a formula {\tt A} holds true, then also the disjunction of {\tt A} with any other formula {\tt B} holds true, regardless of {\tt B}'s truth value. It is clear that a SPARQL rule encoding this entailment would populate a non-empty knowledge graph with an {\it infinite} number of triples: the rule would (recursively) state that every triple of the knowledge graph is disjoint with any other possible triple. Obviously, the lightweight reasoner used in our implementation would then raise a ``Stack overflow'' exception. 

Therefore, Disjunction Introduction should not be added to the proposed computational ontology, or, at most, not in its general version but only in some controlled/restricted versions.

On the other hand, as it is well-known in the literature, the combined used of Disjunction Introduction and Disjunctive Syllogism leads to the so-called ``Ex falso quodlibet'', typically formalized as ({\tt A}\hspace{2pt}$\wedge$\hspace{0.5pt}$\neg$\hspace{1pt}{\tt A})\hspace{1pt}$\rightarrow$\hspace{1pt}{\tt B} and stating that a contradiction entails any formula {\tt B}. A SPARQL rule implementing the Ex falso quodlibet would again populate the knowledge graph with an infinite number of triples whenever the RDF property {\tt is-in-contradiction-with} is derived. This paper will further comment on the Ex falso quodlibet in subsubsection \ref{ParaconsistentDeonticLogic} and section \ref{ConjunctionDisjunctionImplicationOfDeonticStatements} below.

\vspace{-2pt}
\section{Background: the Deontic Traditional Scheme and the state-of-the-art conflict-tolerant deontic logics}\label{DTSandConflictTolerantDLs}
\vspace{-1pt}

\noindent First studies in deontic logic can be traced back to the Middle Age \cite{Knuutila:81}. The Stanford Encyclopedia of Philosophy\footnote{\url{https://plato.stanford.edu/entries/logic-deontic}} offers a rather exhaustive literature review of proposed systems of deontic logic, drawn from \cite{McNamara:96a} and \cite{McNamara:96b}. 

Six normative statuses have been historically identified and largely accepted by today's scientific community: obligatory ({\tt OB}), permitted ({\tt PE}), prohibited\footnote{The Stanford Encyclopedia of Philosophy uses the term ``impermissible'' in place of ``prohibited''; on the contrary, we prefer the latter as it is mostly used in contemporary everyday language. The two terms are here assumed to have the same meaning.} ({\tt PR}), omissible ({\tt OM}), optional ({\tt OP}), and non-optional ({\tt NO}). In formal deontic logic, these statuses are usually encoded into corresponding {\it deontic modalities} applied to a well-formed formula of the underlying logic, e.g., a proposition $p$, in cases where propositional logic is taken as underlying logic. Thus, {\tt OB}({\tt p}), {\tt PE}({\tt p}), {\tt PR}({\tt p}), {\tt OM}({\tt p}), {\tt OP}({\tt p}), and {\tt NO}({\tt p}) are basic deontic (propositional) statements representing that the proposition $p$ is respectively obligatory, permitted, prohibited, omissible, optional, and non-optional. 

It has been also largely accepted that the six deontic modalities are logically related to each other. In particular, {\tt OB} has been traditionally considered as the ``main'' deontic modality from which the other five are formally defined as in (\ref{DeonticModalitiesDefinitions}). The Stanford Encyclopedia of Philosophy calls (\ref{DeonticModalitiesDefinitions}) as ``The Traditional Definitional Scheme''.

\enumsentence{\label{DeonticModalitiesDefinitions}
\begin{enumerate}
    \renewcommand{\labelenumi}{(\alph{enumi})}
    
    \item\noindent {\tt PE(p)} $\leftrightarrow$\hspace{2pt} {\tt $\neg$OB($\neg$p)}
    \vspace{2pt}
    
    \item\noindent {\tt PR(p)} $\leftrightarrow$\hspace{2pt} {\tt OB($\neg$p)}
    \vspace{2pt}
    
    \item\noindent {\tt OM(p)} $\leftrightarrow$\hspace{2pt} {\tt $\neg$OB(p)}
    \vspace{2pt}
    
    \item\noindent {\tt OP(p)} $\leftrightarrow$\hspace{2pt} ({\tt $\neg$OB(p)} $\wedge$ {\tt $\neg$OB($\neg$p)})
    \vspace{2pt}
    
    \item\noindent {\tt NO(p)} $\leftrightarrow$\hspace{2.6pt} ({\tt OB(p)} $\vee$ {\tt OB($\neg$p)})
    
\end{enumerate}
}

\begin{figure}[ht]
  \centering
  \includegraphics[width=0.54\linewidth]{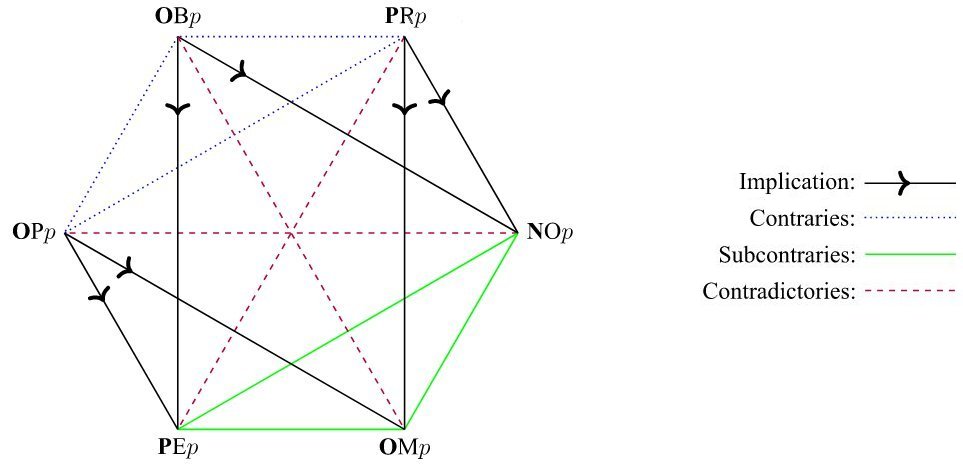}
  \vspace{-2pt}
  \caption{The Deontic Hexagon. In the hexagon, when two vertexes are connected via the ``Implication'' arrow, it means that the antecedent logically entails the consequent; when they are marked as ``Contraries'', it means that they cannot both be true; when they are marked as ``Subcontraries'', that they cannot both be false, and when they are marked as ``Contradictories'', that they always have opposing truth values, i.e., that they cannot neither be both true nor be both false)
  }\label{DeonticHexagon}
\end{figure}

\noindent Moreover, the deontic modalities are logically related of one another as depicted in the Deontic Hexagon, shown\footnote{Figure \ref{DeonticHexagon} has been taken from 
\url{https://plato.stanford.edu/entries/logic-deontic/\#TradScheModaAnal}; consistently with the previous footnote, we replaced the label ``{\tt IM}\hspace{0.5pt}$p$'' (referring to ``impermissible'') with ``{\tt PR}\hspace{0.5pt}$p$'' (referring to ``prohibited'').} in Figure \ref{DeonticHexagon} and also called, in the Stanford Encyclopedia of Philosophy, as the ``Traditional Threefold Classification''.

This paper will consider the logical entailments in the Traditional Definitional Scheme, i.e., the ones in (\ref{DeonticModalitiesDefinitions}), and in the Traditional Threefold Classification, i.e., the ones from the Deontic Hexagon, altogether; the set of all these entailments is henceforth called the ``Deontic Traditional Scheme''.

\vspace{20pt}

\noindent In symbols, as explained in the caption of Figure \ref{DeonticHexagon}, the Deontic Hexagon specifies that:

\vspace{2pt}
\enumsentence{\label{DeonticHexagonEntailments}
\begin{enumerate}
    \renewcommand{\labelenumi}{\roman{enumi}.}
    
    \item\noindent Implication:
    \vspace{-3pt}
    \begin{enumerate}
    \item\noindent {\tt OB(p)} $\rightarrow$ {\tt PE(p)}
    \vspace{1pt}
    \item\noindent {\tt OB(p)} $\rightarrow$ {\tt NO(p)}
    \vspace{1pt}
    \item\noindent {\tt PR(p)} $\rightarrow$ {\tt OM(p)}
    \vspace{1pt}
    \item\noindent {\tt PR(p)} $\rightarrow$ {\tt NO(p)}
    \vspace{1pt}
    \item\noindent {\tt OP(p)} $\rightarrow$ {\tt PE(p)}
    \vspace{1pt}
    \item\noindent {\tt OP(p)} $\rightarrow$ {\tt OM(p)}
    \end{enumerate}
    
    \item\noindent Contraries:
    \vspace{-3pt}
    \begin{enumerate}
    \item\noindent $\neg${\tt (OB(p)} $\wedge$ {\tt PR(p))} equivalent to\hspace{2pt}
    {\tt OB(p)} $\rightarrow$ {\tt $\neg$PR(p)} $\wedge$ 
    {\tt PR(p)} $\rightarrow$ {\tt $\neg$OB(p)}

    \item\noindent $\neg${\tt (OB(p)} $\wedge$ {\tt OP(p))} equivalent to\hspace{2pt}
    {\tt OB(p)} $\rightarrow$ {\tt $\neg$OP(p)} $\wedge$ 
    {\tt OP(p)} $\rightarrow$ {\tt $\neg$OB(p)}

    \item\noindent $\neg${\tt (PR(p)} $\wedge$ {\tt OP(p))} equivalent to\hspace{2pt}
    {\tt PR(p)} $\rightarrow$ {\tt $\neg$OP(p)} $\wedge$ 
    {\tt OP(p)} $\rightarrow$ {\tt $\neg$PR(p)}
    \end{enumerate}
    \vspace{1pt}

    \item\noindent Subcontraries:
    \vspace{-3pt}
    
    \begin{enumerate}
    \renewcommand{\labelenumi}{-}
    \item\noindent $\neg${\tt ($\neg$NO(p)} $\wedge$ {\tt $\neg$PE(p))} equivalent to\hspace{2pt}
    {\tt $\neg$NO(p)} $\rightarrow$ {\tt PE(p)} $\wedge$ 
    {\tt $\neg$PE(p)} $\rightarrow$ {\tt NO(p)}

    \item\noindent $\neg${\tt ($\neg$NO(p)} $\wedge$ {\tt $\neg$OM(p))} equivalent to\hspace{2pt}
    {\tt $\neg$NO(p)} $\rightarrow$ {\tt OM(p)} $\wedge$ 
    {\tt $\neg$OM(p)} $\rightarrow$ {\tt NO(p)}

    \item\noindent $\neg${\tt ($\neg$PE(p)} $\wedge$ {\tt $\neg$OM(p))} equivalent to\hspace{2pt}
    {\tt $\neg$PE(p)} $\rightarrow$ {\tt OM(p)} $\wedge$ 
    {\tt $\neg$OM(p)} $\rightarrow$ {\tt PE(p)}
    
    \end{enumerate}
    
    \item\noindent Contradictories:
    \vspace{-3pt}
    \begin{enumerate}
    \renewcommand{\labelenumi}{-}

    \item\noindent $\neg${\tt (OB(p)} $\wedge$ {\tt OM(p))} $\wedge$ $\neg${\tt ($\neg$OB(p)} $\wedge$ {\tt $\neg$OM(p))} equivalent 
    to\hspace{2pt} {\tt OB(p)}$\leftrightarrow$ {\tt $\neg$OM(p)}
    
    \item\noindent $\neg${\tt (PR(p)} $\wedge$ {\tt PE(p))} $\wedge$ $\neg${\tt ($\neg$PE(p)} $\wedge$ {\tt $\neg$PE(p))} equivalent to\hspace{2pt} {\tt PE(p)}$\leftrightarrow$ {\tt $\neg$PR(p)}

    \item\noindent $\neg${\tt (OP(p)} $\wedge$ {\tt NO(p))} $\wedge$ $\neg${\tt ($\neg$OP(p)} $\wedge$ {\tt $\neg$NO(p))} equivalent to\hspace{2pt} {\tt OP(p)} $\leftrightarrow$ {\tt $\neg$NO(p)}
    
    \end{enumerate}
\end{enumerate}
}
\vspace{2pt}

\noindent This paper, in order to minimize the set of symbols and to avoid redundancies between
(\ref{DeonticModalitiesDefinitions}) and (\ref{DeonticHexagonEntailments}), will only use the deontic modalities {\tt OB}, {\tt PE}, and {\tt OP} (obligatory, permitted, and optional); on the other hand, by virtue of the bi-implications in (\ref{DeonticHexagonEntailments}.iv.(a)-(c)), it will represent {\tt OM}, {\tt PR}, and {\tt NO} (omissible, prohibited, and not-optional) as $\neg$\hspace{1pt}{\tt OB}, $\neg$\hspace{1pt}{\tt PE}, and $\neg$\hspace{1pt}{\tt OP}. Still, the narrative below will often use the terms ``prohibitions'' and ``prohibited'', although these will be formally represented as non-permissions, because they occur rather frequently in existing legislation.

By using the bi-implications in (\ref{DeonticHexagonEntailments}.iv.(a)-(c)), the entailments in the Deontic Traditional Scheme can be all represented in terms of
{\tt OB}, {\tt PE}, and {\tt OP}, plus the boolean connectives $\neg$\hspace{1pt}, $\wedge$\hspace{1pt}, $\rightarrow$, and $\leftrightarrow$. The set of entailments 
is then reduced to the set of only the ones shown in (\ref{DeonticTraditionalSchemeEntailments}), as all other entailments are implied by one of these three.

\enumsentence{\label{DeonticTraditionalSchemeEntailments}
\begin{enumerate}
    \renewcommand{\labelenumi}{\alph{enumi}.}
    
    \item\noindent {\tt PE(p)} $\leftrightarrow$\hspace{2pt} {\tt $\neg$OB($\neg$p)}
    \vspace{2pt}
    
    \item\noindent {\tt OP(p)} $\leftrightarrow$\hspace{2pt} ({\tt $\neg$OB(p)} $\wedge$ {\tt $\neg$OB($\neg$p)})
    \vspace{2pt}
    \item\noindent {\tt OB(p)} $\rightarrow$ {\tt PE(p)}
\end{enumerate}
}

\noindent Although, as explained earlier, it is widely accepted that the entailments of the Deontic Traditional Scheme properly reflect our intuitions about the six deontic modalities, most deontic logics proposed in the literature only formalize (what is considered as) the ``main'' deontic modality, i.e., obligatoriness ({\tt OB}).

In particular, according to \cite{Goble:13}, the benchmark deontic logic is the propositional modal logic that axiomatizes {\tt OB} as in (\ref{axiomatizationSDLfromGoble}). In (\ref{axiomatizationSDLfromGoble}), {\tt D}, {\tt C}, {\tt NM}, {\tt P}, and {\tt N} are the names of the axioms while $\Box$ and $\Diamond$ are the classical operators of necessity and possibility. The axioms in (\ref{axiomatizationSDLfromGoble}) form the basis of normal modal
logics {\tt KD}\footnote{\url{https://plato.stanford.edu/entries/logic-modal}} for the operator {\tt OB}.

\enumsentence{\label{axiomatizationSDLfromGoble}
\begin{enumerate}
    \renewcommand{\labelenumi}{}
    
    \item\noindent ({\tt D})\hspace{20pt}
    {\tt OB}(\hspace{1pt}{\tt p})    $\rightarrow$ $\neg$\hspace{1pt}{\tt OB}($\neg$\hspace{1pt}{\tt p})
    
    \vspace{2pt}
    \item\noindent ({\tt C})\hspace{20pt}
    ({\tt OB}(\hspace{1pt}{\tt p}\hspace{1pt}) $\wedge$ {\tt OB}({\tt q}\hspace{1pt})) $\rightarrow$ {\tt OB}(\hspace{1pt}{\tt p}\hspace{1pt}$\wedge$\hspace{1pt}{\tt q}\hspace{1pt})
    
    \vspace{2pt}
    \item\noindent ({\tt NM})\hspace{17pt}$\Box$\hspace{1pt}(\hspace{1pt}{\tt p}\hspace{1pt}$\rightarrow$\hspace{1pt}{\tt q}\hspace{1pt}) $\rightarrow$ ({\tt OB}(\hspace{1pt}{\tt p}) $\rightarrow$ {\tt OB}({\tt q}\hspace{1pt}))

    \vspace{2pt}
    \item\noindent ({\tt P})\hspace{20pt} {\tt OB}({\tt p})    $\rightarrow$ $\Diamond$\hspace{1pt}(\hspace{1pt}{\tt p})
    
    \vspace{2pt}
    \item\noindent ({\tt N})\hspace{20pt}
    $\Box$\hspace{1pt}(\hspace{1pt}{\tt p}\hspace{1pt}) $\rightarrow$ {\tt OB}(\hspace{1pt}{\tt p})
\end{enumerate}
}

\noindent It is immediate to realize that the axiomatization in (\ref{axiomatizationSDLfromGoble}) is not suitable for the objectives of this paper, in particular because of the axiom {\tt D}. This axiom states that the obligation of a proposition {\tt p} is in contradiction with the obligation of its opposite proposition, i.e., $\neg$\hspace{1pt}{\tt p}: the two deontic statements cannot hold together. Therefore, with respect to first example seen above in the Introduction, by taking {\tt OB}({\tt l}) as ``It is obligatory to leave the building'' and {\tt OB}($\neg$\hspace{1pt}{\tt l}) as ``It is obligatory to not leave the building'', ({\tt D}) derives that {\tt OB}({\tt l}) $\wedge$ {\tt OB}($\neg$\hspace{1pt}{\tt l})\hspace{2pt}$\rightarrow$\hspace{1pt}$\bot$. 

On the contrary, this paper aims at representing  
{\tt OB}({\tt l}) and {\tt {\tt OB}($\neg$\hspace{1pt}{\tt l})} as {\it conflicting}, rather than {\it contradictory}, deontic statements. Therefore, to achieve our objectives, a different axiomatization for the deontic operators must be postulated. To do so, the first step is of course to analyze conflict-tolerant deontic logics proposed in past literature, of which \cite{Goble:13} represents, in our view, the most complete survey.

\subsection{Conflict-tolerant deontic logics proposed in the literature}\label{conflictTolerantDeonticLogicGoble}

\noindent Since the axiomatization in (\ref{axiomatizationSDLfromGoble}) represents conflicts as contradictions, \cite{Goble:13} states that the first Desideratum for a conflict-tolerant deontic logic must be:

\enumsentence{\label{Desideratum1}
\textbf{Desideratum \#1}: A conflict-tolerant deontic logic accepts conflicts between pairs of deontic statements as {\it consistent}. In \cite{Goble:13}, a conflict is intuitively defined as a situation in which ``an agent ought to do a number of things, each of which is
possible for the agent, but it is impossible for the agent to do them all''.
}

\noindent As exemplified above, what specifically prevents to achieve Desideratum \#1 with the axiomatization in (\ref{axiomatizationSDLfromGoble})
is the axiom {\tt D}. However, the mere removal of {\tt D} is likewise unsatisfactory, because the remaining axioms would lead to the so-called ``deontic explosion'': as soon as a conflict between two or more deontic statements is inferred, it would be also inferred that everything is obligatory, which is of course meaningless. In light of this, \cite{Goble:13} imposes a second Desideratum for a conflict-tolerant deontic logic:

\enumsentence{\label{Desideratum2}
\textbf{Desideratum \#2}: A conflict-tolerant deontic logic must not generate deontic explosion from a conflict of deontic statements. In other words, if two deontic statements conflict of one another, the conflict-tolerant deontic logic must {\it not} infer that everything is obligatory.
}

\noindent Finally, \cite{Goble:13} adds a third Desideratum generally stating that conflict-tolerant deontic logics must be able to capture our intuitions about obligatoriness and related notions. The Desideratum is postulated in \cite{Goble:13} as follows:

\enumsentence{\label{Desideratum3}
\textbf{Desideratum \#3}: A conflict-tolerant deontic logic should explain in a plausible way the apparent validity of several paradigm arguments.
}

\noindent However, although  (\ref{Desideratum3}) refers to paradigm arguments {\it in general}, \cite{Goble:13} focuses on a particular one called ``The Smith argument'', together with some variants called the ``Jones'', the ``Roberts'', and the ``Thomas'' arguments. \cite{Goble:13} specify that the paradigm arguments mentioned in (\ref{Desideratum3}) {\it include} the Smith argument and its variants but actually no further paradigmatic argument is discussed in \cite{Goble:13}. This paper, on the other hand, examines a wider sets of situations featuring conflicts among deontic statements, for which the conflict-tolerant deontic logics reviewed in \cite{Goble:13} do not seem to provide a satisfactory account, as it will be argued in the next section.

The Smith argument was originally introduced in \cite{Horty:94} and it states that:

\enumsentence{\label{SmithArgument}
\noindent\hspace{0pt}{\it From}:\hspace{2pt} Smith ought to fight in the army or perform alternative
national service.\\[4pt]
\noindent\mbox{}\hspace{0pt}{\it And}:\hspace{2pt} Smith ought not to fight in the army.\\[4pt]
\noindent\mbox{}\hspace{0pt}{\it It is intuitive to conclude that:}\hspace{2pt} Smith ought to perform alternative national service.
}

\noindent The axioms in (\ref{axiomatizationSDLfromGoble}) are unable to derive the Smith argument because none of them involves the disjunctive boolean connective ``$\vee$''. Other axioms must be then introduced to account for the Smith argument. The Jones, Roberts, and Thomas argument are more elaborated variants of the Smith argument in which disjunction further interplays with the conjunctive boolean connective ``$\wedge$''.

The next three subsubsections briefly summarize the main approaches that have been proposed in the literature to deal with irresolvable conflicts. These have been categorized in \cite{Goble:13} as: (1) Revisionist strategies, (2) Paraconsistent deontic logics, and 
(3) Other radical strategies.

\subsubsection{Revisionist strategies}\label{revisionistStrategies}
\vspace{2pt}

\noindent Revisionists strategies are approaches that represent and reason with conflicts in deontic logic via axiomatizations alternative to the one in (\ref{axiomatizationSDLfromGoble}). Although \cite{Goble:13} presents twelve different deontic logics as such, only the axiomatization of one of them, called {\tt BDL}, as well as a slight variant of it called {\tt BDLcc}, achieve all three Desiderata.
 
{\tt BDL} only imports the axiom ({\tt C}) from (\ref{axiomatizationSDLfromGoble}) above, whereas ({\tt D}), ({\tt P}), and ({\tt N}) are rejected. ({\tt NM}) is partially rejected in the sense that it is replaced by another axiom called ({\tt RBE}), whose inferential power is much more restricted than ({\tt NM})'s. Finally other two axioms called ({\tt DDS}) and ({\tt M}) are added to the axiomatization in order to enable the derivations on the boolean connective ``$\vee$'' and ``$\wedge$'' needed to explain the Smith argument and its variants. {\tt BDL}'s axiomatization is shown in (\ref{axiomatizationBDL}). In (\ref{axiomatizationBDL}), ``$\leftrightarrow_{\tt\hspace{-0.5pt} A}$'' is a restricted version of the standard bi-implication operator ``$\leftrightarrow$''; the restricted bi-implication ``$\leftrightarrow_{\tt\hspace{-0.5pt} A}$'' is defined in a way that avoids deontic explosion (see \cite{Goble:13} for formal details).

\vspace{10pt}
\enumsentence{\label{axiomatizationBDL}
\begin{enumerate}
    \renewcommand{\labelenumi}{}
    
    \item\noindent ({\tt M})\hspace{23pt}
    {\tt OB}(\hspace{1pt}{\tt p}\hspace{1pt}$\wedge$\hspace{1pt}{\tt q}\hspace{1pt}) $\rightarrow$ {\tt OB}(\hspace{1pt}{\tt p})
    
    \vspace{2pt}
    \item\noindent ({\tt C})\hspace{23pt}
    ({\tt OB}(\hspace{1pt}{\tt p}\hspace{1pt}) $\wedge$ {\tt OB}({\tt q}\hspace{1pt})) $\rightarrow$ {\tt OB}(\hspace{1pt}{\tt p}\hspace{1pt}$\wedge$\hspace{1pt}{\tt q}\hspace{1pt})
    
    \vspace{2pt}
    \item\noindent ({\tt DDS})\hspace{11pt}
    ({\tt OB}(\hspace{1pt}{\tt p}\hspace{1pt}$\vee$\hspace{1pt}{\tt q}\hspace{1pt}) $\wedge$ {\tt OB}(\hspace{1pt}$\neg$\hspace{1pt}{\tt q}\hspace{1pt})) $\rightarrow$ {\tt OB}(\hspace{1pt}{\tt p})

    \vspace{2pt}
    \item\noindent ({\tt RBE})\hspace{11pt}
    (\hspace{1pt}{\tt p}\hspace{2pt}$\leftrightarrow_{\tt\hspace{-0.5pt} A}$\hspace{2pt}{\tt q}\hspace{1pt}) $\rightarrow$ ({\tt OB}(\hspace{1pt}{\tt p}\hspace{1pt})\hspace{2pt}$\leftrightarrow$\hspace{2pt}{\tt OB}(\hspace{1pt}{\tt q}\hspace{1pt}))
\end{enumerate}
}

\noindent As it will be discussed in the next sections, {\tt BDL}'s  is the axiomatization closest to the one implemented in the proposed computational ontology, which, not surprisingly, is also capable to achieve all three Desiderata.

In particular, the authors of this paper fully share the decision of rejecting axioms ({\tt D}), ({\tt P}), and ({\tt N}) from the axiomatization of a conflict-tolerant deontic logic. 

The axiom ({\tt D})  must be rejected because, as explained earlier, it infers that conflicts are contradictions in cases where both a proposition {\tt p} and its negation are obligatory. 

Axiom ({\tt P}), on the other hand, seems to be rather counter-intuitive. ({\tt P}) stipulates that if something is obligatory, then it is possible. However, it is easy to imagine situations in which this does not seem to be the case. For instance, (\ref{conflictingobligations2}) above exemplified a situation where it is prohibited to pay in cash in a state of affairs in which it is {\it impossible} to avoid paying by cash, i.e., when someone must pay at the parking meter in Sketty, which only accepts cash. 

Similar considerations hold for axiom ({\tt N}), which stipulates that if something is necessary then it is obligatory. This entailment does not seem to match our intuitions either. For instance, in the same state of affairs just considered, it is necessary to pay in cash at the parking meter in Sketty while it is {\it not} obligatory to do so; in fact, this is prohibited.

\subsubsection{Paraconsistent deontic logics}\label{ParaconsistentDeonticLogic}
\vspace{2pt}

\noindent Revisionist strategies propose axiomatizations, in classical modal logic enriched with the operator {\tt OB}\hspace{1pt}, that represent conflicts as consistent formulae while avoiding deontic explosion. In particular, to avoid deontic explosion, the axiom ({\tt C}) or the axiom ({\tt NM}) are restricted fit to inhibit the inferences that generate it; for instance, the logic {\tt BDL} prevents deontic explosion by replacing the axiom ({\tt NM}) with its restricted version ({\tt RBE}).

Paraconsistent deontic logics, on the other hand, avoid deontic explosion by blocking a specific inference rule holding in the underlying classical modal logic: the Ex falso quodlibet, which, together with ({\tt C}) and ({\tt NM}), is responsible for the deontic explosion. By blocking the Ex falso quodlibet, the axiomatization on the operator {\tt OB} can retain the full inferential power offered by ({\tt C}) and ({\tt NM}).

The Ex falso quodlibet was already mentioned above in subsubsection \ref{representingFurtherInferences}. This inference rule  stipulates that {\it any} proposition is derivable from a contradiction.
The Ex falso quodlibet, however, is in turn derived from other two well-known inference rules characterizing the boolean connective ``$\vee$'': Disjunction Introduction, stating that {\tt A}$\rightarrow$\hspace{1pt}({\tt A}\hspace{0.5pt}$\vee$\hspace{1pt}{\tt B}), and Disjunctive Syllogism, stating that (({\tt A}$\vee${\tt B})\hspace{1pt}$\wedge$$\neg${\tt B}) \hspace{-1pt}$\rightarrow$\hspace{1pt}{\tt A}. Therefore, to block the Ex falso quodlibet the paraconsistent logic must block at least one among Disjunction Introduction and Disjunctive Syllogism.

According to \cite{Goble:13}, the tendency in the literature in paraconsistent logics, as well as relevant logics, is to block Disjunctive Syllogism. Nevertheless, without Disjunctive Syllogism the axiomatization the Smith argument and in turn Desiderata \#3 cannot be easily achieved. Some solutions to accommodate the Smith argument in paraconsistent deontic logics that lack Disjunctive Syllogism are discussed in \cite{Goble:13}.

Concerning relevant logics, on the other hand, it must be pointed out that ``the rejection of Disjunctive Syllogism, however, has become one of the most controversial aspects of relevance logic''\footnote{Cit. from \url{https://plato.stanford.edu/entries/logic-relevance}}.
Indeed, even if mainstream relevance logicians opt to reject Disjunctive Syllogism, it is worth observing that such a principle is at least admissible in some relevant logics, as the corresponding rule is theorem-preserving: if {\tt A}$\vee${\tt B} and $\neg$\hspace{1pt}{\tt A} are theorems, so is {\tt B} \cite{Jago:20}. Furthermore, it turns out that weaker relevant logics  consider what happens when Disjunctive Syllogism is added as a primitive rule \cite{Robles-Mendez:10} \cite{Robles-Mendez:11}, whose semantic counterpart is a Routley-Meyer ternary semantics \cite{Routley-Mayer:73} \cite{Routley-Mayer:72a} \cite{Routley-Mayer:72b}, with respect to whom the logic is sound and complete. Another crucial aspect is the deep relationship between Disjunctive Syllogism and the cut rule, a well-established and known property needed to provide proof methods and reasoning mechanisms for a given logic \cite{Gentzen:64}: removing Disjunctive Syllogism should have a negative impact on the opportunity of having sound and complete provers for relevant logics. As it is well-known,  cut  is at the basis of the Prolog programming language, perhaps the most popular logic programming language, which suggests to prefer Disjunctive Syllogism over Disjunctive Introduction also from the point of view of practical applications. Similar ``practical'' considerations also hold for SPARQL, as already discussed at the end of subsubsection \ref{representingFurtherInferences} above: including a SPARQL rule corresponding to the (general form) of Disjunctive Introduction would populate the knowledge graph with an infinite number of triples.

Given the above arguments, the proposed computational ontology avoids Disjunctive Introduction and, consequently, the Ex falso quodlibet. We will further elaborate on these issues again in section \ref{ConjunctionDisjunctionImplicationOfDeonticStatements} below.

\subsubsection{Other radical strategies}\label{otherRadicalStrategies}
\vspace{2pt}

\noindent In \cite{Goble:13}, paraconsistent deontic logics are classified as a type of ``radical strategies'' to deal with conflicts among obligations. However, in this paper we prefer to keep them separated from other radical strategies because they share with the revisionist strategies the fact that they are also grounded on classical modal logic on top of which the operator {\tt OB} is defined and axiomatized.

Conversely, the other radical strategies presented in \cite{Goble:13} differ from the revisionist strategies and the paraconsistent logics because they also include {\it an additional upper level} to control and constrain the inferences underneath.

Three examples, discussed in \cite{Goble:13}, of such radical strategies are Two-phase deontic logic \cite{vanderTorre-Tan:00}, Imperatival approaches, e.g., \cite{Hansen:08} and \cite{Horty:12}, and Adaptive Deontic Logics, e.g., \cite{Goble:13b} and \cite{vanDePutte-etal:19}.

In Two-phase deontic logic, the upper level controls and constraints the order of application of the inference rules on {\tt OB} and on the classical modal logic operators. In particular, Two-phase deontic logic stipulates that rules of aggregation, e.g., the axiom ({\tt C}), must be always executed {\it before} rules of distribution, e.g., the axiom ({\tt NM}). \cite{vanderTorre-Tan:00} show that, by imposing this order on the execution of the rules, the logic is able to represent conflicts as consistent formulae, it properly infers the Smith argument, and it avoids generating deontic explosion. Nevertheless, although \cite{vanderTorre-Tan:00}'s solution works from a technical point of view, at least in all considered examples, \cite{Goble:13} questions the rationale behind it. Why should rules be executed in that particular order? Which intuitions about the notion/interpretation of conflicts does that order reflect? \cite{vanderTorre-Tan:00} do not seem to provide an answer to these questions.

Imperatival approaches maintain that obligations are always associated with commands or directives. The intuition is that, whenever some agents are obliged to do something, it is because they were {\it ordered} to do so, e.g., by the law or by other entities having the power to give them orders. In Imperatival approaches, therefore, the upper level consists of a set of commands associated with the set of obligations in the lower level. In this setting, whenever two  obligations conflict of one another, it is because the associated commands conflict of one another. The agents to whom the commands were given will at most comply with a maximal set of non-conflicting obligations and, unless otherwise specified, i.e., unless some obligations are given priority over the others, the agents will be free to decide which obligations include in this set. Once this set has been identified, since the obligations therein are non-conflicting, there is no need to restrict the underlying classical modal logic nor {\tt OB}'s axiomatization. Note that Imperatival approaches are non-monotonic, because the agent in charge of executing conflicting orders will have multiple choices to identify the maximal set of non-conflicting obligations.  

Finally, Adaptive Deontic Logics 
are dynamic and non-monotonic logical frameworks in which the constraints on the application of the rules may {\it change over time}. Specifically, Adaptive Logics are defined, in the most general way, as triples $\langle${\tt LLL}, $\Omega$, {\tt Strategy}$\rangle$, in which {\tt LLL} is a (basic) logic featuring certain properties (e.g., reflexivity, monotonicity, etc. see \cite{Goble:13b} for details), $\Omega$ is a set of {\it abnormalities}, formalized as a set of formulae that are considered problematic and, therefore, for which the logic must ``adapt'', and {\tt Strategy} is a set of rules specifying {\it how} the inference rules of the logic adapt to deal with abnormalities once these have been detected. Given this definition, an Adaptive Deontic Logic might be built by setting {\tt LLL} as classical modal logic enriched with the operator {\tt OB}, $\Omega$ as the set of all possible formulae that denote a conflict, i.e., $\Omega\equiv\{A: \exists C\hspace{1pt}[A = {\tt OB}(C\hspace{1pt})\wedge {\tt OB}(\neg\hspace{1pt}C\hspace{1pt})]\}$, and {\tt Strategy} as a set of rules that block undesirable consequences, in primis deontic explosion, as soon as a conflict/abnormality is derived. Several configurations of inference rules in {\tt LLL} and {\tt Strategy} can be defined in order to explain the Smith arguments and, more generally, Desideratum \#3, while blocking the undesirable consequences.

\subsection{An (introductive) comparison between the conflict-tolerant deontic logics proposed in the literature and the computational ontology proposed in this paper}

\noindent This final subsection recaps the main insights of the conflict-tolerant presented in \cite{Goble:13} and compares them from a technical point of view with the proposed computational ontology.

\begin{figure}[ht]
  \centering
  \includegraphics[width=0.9\linewidth]{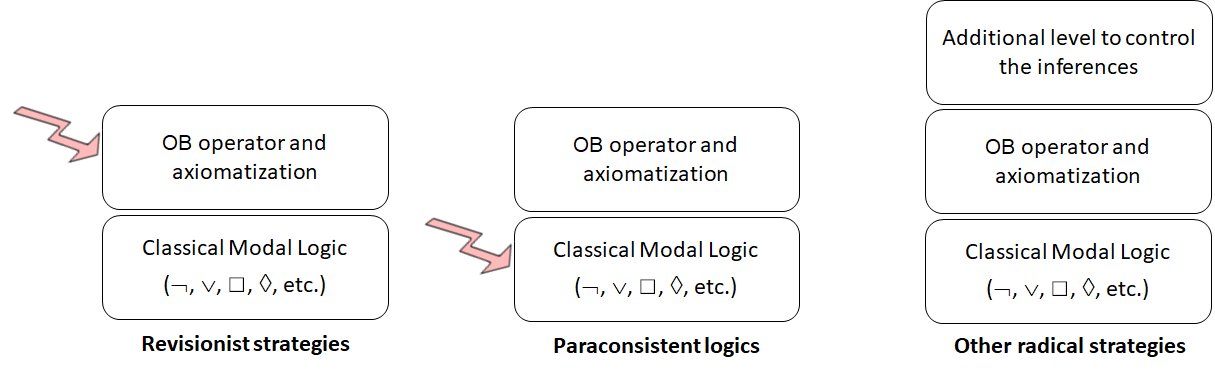}
  \vspace{-6pt}
  \caption{A graphical comparison among the conflict-tolerant deontic logics proposed in the literature}\label{CtDlfromLiterature}
\end{figure}

Figure \ref{CtDlfromLiterature} shows a graphical representation that illustrates and compares the insights of the conflict-tolerant deontic logics mentioned in the previous subsection. Revisionist strategies, as the red arrow indicates, revise the benchmark axiomatization in
(\ref{axiomatizationSDLfromGoble}). Paraconsistent deontic logics, on the other hand, act on the axioms of the underlying classical modal logic, specifically on those responsible for the Ex falso quodlibet, which is in turn responsible for deontic explosion. Finally, other radical strategies feature an additional level whose constructs control the inferences on the first two levels; the constructs in the additional level could impose a particular execution order of the rules, as in Two-phase deontic logic, represent a set of commands associated with the obligations as in the Imperatival approaches, or a set of abnormalities together with a strategy to deal with them as in the Adaptive Deontic Logics.

\noindent On the other hand, what all the three approaches have in common is the fact that they are all grounded on an extension of classical modal logic. Specifically, the latter is extended by adding a new operator, {\tt OB}, whose inferences are then enabled/constrained by a specific axiomatization. 

\begin{figure}[ht]
  \centering
  \includegraphics[width=0.23\linewidth]{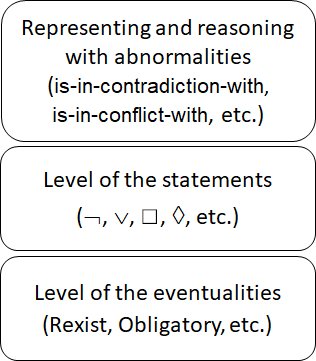}
  \vspace{-5pt}
  \caption{The three levels of the computational ontology proposed in this paper}\label{CtDlComputationalOntology}
\end{figure}

The main technical difference between the conflict-tolerant deontic logics proposed in the literature and the computational ontology proposed in this paper is that, as it should be already clear from section \ref{EncodingNLstatements} above and as it is depicted in Figure \ref{CtDlComputationalOntology}, in the proposed computational ontology the first two levels are {\it swapped}: modalities, such as {\tt Rexist} or the deontic modalities, e.g., {\tt Obligatory}, which will be defined in the next section, are represented in the {\it lowest} level, i.e., the level of the eventualities; on top of that, the level of the statements implements the operators from classical modal logic: boolean connectives such as $\neg$ and $\vee$ and, as it will be shown below in section \ref{DeonticModalitiesAndContextualConstraints}, the modal operators $\Box$ and $\Diamond$, for necessity and possibility respectively; finally, the third level represents abnormalities such as contradictions and conflicts, which are respectively encoded by the properties {\tt is-in-contradiction-with}, introduced above in subsection \ref{representingAndInferringContradictions}, and {\tt is-in-conflict-with}, which will be introduced in the next section.

\noindent On the other hand, the proposed computational ontology also shares insights with each of the three categories of conflict-tolerant deontic logics depicted in Figure \ref{CtDlfromLiterature}. The SPARQL rules that will be defined at the level of the eventualities for handling the meaning of the deontic modalities mostly parallel the axiomatization of the {\tt BDL} logic seen in subsubsection \ref{revisionistStrategies} above. The Ex falso quodlibet is prevented by avoiding  SPARQL rules at the level of the statements that would add infinite sets of triples to the knowledge graph. Finally, similarly to the Adaptive Deontic Logics, also the proposed computational ontology explicitly represents abnormalities.

However, it must be pointed out that while the Adaptive Deontic Logics proposed in the literature only consider conflicts of obligations in the set of abnormalities, the proposed computational ontology  considers a wider set of abnormalities, including contradictions. In addition, the proposed computational ontology does not explicitly represent abnormalities with the sole aim of blocking undesired inferences, as it is done in Adaptive (Deontic) Logics. More generally, as explained in the Introduction, the solution proposed here aims at {\it reasoning with} the abnormalities. For instance, in cases where conflicts among norms are derived, we could reason about possible revisions of the law that solve them; in cases where contradictions are derived, we could reason about the causes that led to those contradictions, e.g., whether there are errors in the database and where these could be located. 

Finally, a second significant technical difference between the conflict-tolerant deontic logics presented in \cite{Goble:13} and the computational ontology proposed in this paper is that in the latter everything is implemented in RDFs and SPARQL, thus obtaining all advantages offered by these technologies in terms of interoperability and deployment within existing applications: all main contemporary programming languages offer libraries to work with these two standards and more and more RDF datasets are becoming available online. By contrast, to the best of our knowledge, no implementation of the logics presented in \cite{Goble:13} is available: the results presented therein are exclusively theoretical. In addition, the operators and constructs used therein appear to us rather difficult to implement, especially for the radical strategies illustrated in subsubsection \ref{otherRadicalStrategies}, which employ truth-functional constructs together with constructs featuring operational semantics.

On the other hand, there are also {\it more fundamental and non-technical} differences between the proposed computational ontology and the conflict-tolerant deontic logics discussed above, namely different {\it intuitions} about the notion of irresolvable conflict. These different intuitions will be explained in the next section.

\section{Modelling and reasoning with deontic modalities in RDFs and SPARQL}\label{DeonticModalitiesInRDFsAndSPARQL}

\noindent This section contains the core of the research presented in this paper. The three previous sections illustrated the three research strands from past literature that this paper will funnel contributions from in order to develop the proposed computational ontology: (1) past literature on implemented LegalTech solutions for compliance checking on RDF data, illustrated in section \ref{BackgroundLegalTechOnRDF}, (2) past literature in Natural Language Semantics via reification, specifically the framework in \cite{Gordon-Hobbs:17} illustrated in section \ref{EncodingNLstatements}, and (3) past literature in conflict-tolerant deontic logics, specifically the seminal survey in \cite{Goble:13} illustrated in section \ref{DTSandConflictTolerantDLs}.

While section \ref{EncodingNLstatements} has already shown how the contributions from (1) and (2) might be merged into the proposed computational ontology, namely how the framework in \cite{Gordon-Hobbs:17} might be implemented in RDFs and SPARQL, this section will show how it is possible to extend that implementation fit to incorporate deontic modalities.

At the end of the previous section, it has been pointed out that the formalization of deontic modalities in the proposed computational ontology is grounded on fundamental intuitions, about the notion of conflicts among deontic statements, different from the ones assumed in \cite{Goble:13}. These intuitions are summarized in (\ref{intuitions}.a-c).

\enumsentence{\label{intuitions}
\begin{enumerate}
    \renewcommand{\labelenumi}{\alph{enumi}.}
    \item The deontic modalities and the inferences of the Deontic Traditional Scheme are assumed to be valid. The  approaches reviewed in \cite{Goble:13} only focus on obligatoriness while this paper considers all deontic modalities defined in the Deontic Traditional Scheme, without focusing on any of them in particular. Therefore, this paper will also consider conflicts among prohibitions and permissions, e.g., the one exemplified in (\ref{conflictingobligations1subcategory}) above; these are not addressed in \cite{Goble:13} because, according to the Deontic Traditional Scheme, permissions do not entail obligations, they only entail {\it not-}obligations. Furthermore, this paper will also consider partial conflicts among obligations, e.g., the one exemplified in (\ref{conflictingobligationsPartial}) above. These are not addressed in \cite{Goble:13} either because the approaches discussed therein only involve propositional symbols and propositional symbols are not expressive enough to model actions/states and the thematic roles that the individuals in the state of affairs play in these actions/states. To do so, first-order formats, such as RDF and SPARQL, are needed.
    Reasoning with all deontic modalities, and not only with obligations, is also what led us to the next intuition, which we consider as the central one.
    
    \item The conflict-tolerant deontic logics reviewed in \cite{Goble:13} are grounded on the intuition/definition reported in (\ref{Desideratum1}) above: a conflict (of obligations) is defined as a situation in which ``an agent ought to do a number of things, each of which is possible for the agent, but it is impossible for the agent to do them all'', formalized in \cite{Goble:13} as ``{\tt OB}({\tt A}) $\wedge$ {\tt OB}({\tt B}) $\wedge\hspace{2pt}\neg\hspace{0.5pt}\Diamond$({\tt A}$\wedge${\tt B})''. On the contrary, as explained in the Introduction, this paper is grounded on an alternative definition inspired by the work of Hans Kelsen: a conflict (of deontic statements) is a situation in which ``two or more deontic statements hold in the context but complying with one of them entails violating (or not permitting) another one''.  Still, we do not intend to claim that \cite{Goble:13}'s definition is ``wrong'' and that our formalization does not encompass it. In fact, it does, as it will be shown below in section \ref{DeonticModalitiesAndContextualConstraints}. This paper only claims that \cite{Goble:13}'s is not the right definition of conflicts among obligations but it is rather a specific case of the interplay between obligatoriness and contextual knowledge.
    \vspace{5pt}
    
    \item All deontic statements have an agent. This claim is not new; there is actually a vast literature about the interplay between deontic modalities and agency\footnote{See \url{https://plato.stanford.edu/entries/logic-deontic/\#DeonLogiAgen}}, a recent approach being \cite{Frijters-etal:21}. This literature led to the definition of the so-called ``normative positions'', i.e., the formalization of the well-known Hohfeld's legal relations \cite{Sergot:13}, as well as to the view that a proper truth-functional logic of norms is impossible in that norms do not carry truth values. This view is assumed in the literature as the most suitable explanation of the well-known J{\o}rgensen's dilemma \cite{Jorgensen:37} and it is at the basis of Input/Output logic \cite{Makinson-vanderTorre:00}, mentioned at the end of the Introduction. In line with this literature, this paper assumes that deontic modalities, similarly to the {\tt Rexist} modality, only applies to  eventualities, and, therefore, that they must be formalized in the lowest level of Figure \ref{CtDlComputationalOntology}. Furthermore, eventualities holding for deontic modalities always specify the {\tt has-agent} thematic role, called the ``bearer'' of the corresponding deontic statements.
    \vspace{3pt}
\end{enumerate}
}

\noindent The next three subsections further elaborate each (\ref{intuitions}.a-c) respectively.

\vspace{5pt}
\subsection{Modelling the Deontic Traditional Scheme}

\noindent A new RDFs class {\tt DeonticModality} is added to the computational ontology as subclass of {\tt Modality}. Other three RDFs classes {\tt Obligatory}, {\tt Permitted}, and {\tt Optional} are added as instances of {\tt DeonticModality}. These respectively correspond to the operators {\tt OB}, {\tt PE}, and {\tt OP} of the Deontic Hexagon. The operators {\tt OM}, {\tt PR}, and {\tt NO} are instead modelled as $\neg$\hspace{1pt}{\tt OB}, $\neg$\hspace{1pt}{\tt PE}, and $\neg$\hspace{1pt}{\tt OP} respectively, as explained in the previous section.

\enumsentence{\label{DeonticModalitiesClasses}
\mbox{}\hspace{5pt}{\tt :DeonticModality}\hspace{5pt}{\tt a}\hspace{3pt} {\tt rdfs:Class;} {\tt rdfs:subClassOf}\hspace{3pt}{\tt :Modality.}\\[2pt]
\mbox{}\hspace{5pt}{\tt :Obligatory}\hspace{3pt}{\tt a}\hspace{3pt} {\tt rdfs:Class,:DeonticModality.}\\[2pt]
\mbox{}\hspace{5pt}{\tt :Permitted}\hspace{3pt}{\tt a}\hspace{3pt} {\tt rdfs:Class,:DeonticModality.}\\[2pt]
\mbox{}\hspace{5pt}{\tt :Optional}\hspace{3pt}{\tt a}\hspace{3pt} {\tt rdfs:Class,:DeonticModality.}
}

\noindent Furthermore, the computational ontology includes SPARQL rules that implement the entailments of the Deontic Traditional Scheme. It has been explained at the beginning of the previous section that, in terms of the three deontic modalities {\tt OB}, {\tt PE}, and {\tt OP}, the entailments in the Deontic Traditional Scheme are all equivalent to one of the entailments in (\ref{DeonticTraditionalSchemeEntailments}), repeated again in (\ref{DeonticTraditionalSchemeEntailmentsRepeated}) for reader's convenience.

\enumsentence{\label{DeonticTraditionalSchemeEntailmentsRepeated}
\begin{enumerate}
    \renewcommand{\labelenumi}{\alph{enumi}.}
    
    \item\noindent {\tt PE(p)} $\leftrightarrow$\hspace{2pt} {\tt $\neg$OB($\neg$p)}
    \vspace{2pt}
    
    \item\noindent {\tt OP(p)} $\leftrightarrow$\hspace{2pt} ({\tt $\neg$OB(p)} $\wedge$ {\tt $\neg$OB($\neg$p)})
    \vspace{2pt}
    
    \item\noindent {\tt OB(p)} $\rightarrow$ {\tt PE(p)}
\end{enumerate}
}

\noindent Each of the entailments in (\ref{DeonticTraditionalSchemeEntailmentsRepeated}) enables {\it two} derivations, via Modus Ponens and via Modus Tollens respectively. For example, the two entailments in (\ref{DeonticTraditionalSchemeEntailmentsRepeated}.a) correspond to following four derivations: 
\vspace{10pt}

\enumsentence{\label{DerivationsBetweenOBandPE}
\begin{enumerate}
    \renewcommand{\labelenumi}{\alph{enumi}.}    
    \item\noindent 
    {\tt PE(p)}\hspace{1pt}$\vdash$\hspace{3pt}{\tt $\neg$OB($\neg$p)}, via Modus Ponens.
    \vspace{2pt}
    \item {\tt $\neg$OB($\neg$p)}\hspace{1pt}$\vdash$\hspace{3pt}{\tt PE(p)}, via Modus Ponens.
    \vspace{2pt}
    \item {\tt OB($\neg$p)}\hspace{1pt}$\vdash$\hspace{3pt}{\tt $\neg$PE(p)}, via Modus Tollens.
    \vspace{2pt}
    \item {\tt $\neg$PE(p)}\hspace{1pt}$\vdash$\hspace{3pt}{\tt OB($\neg$p)}, via Modus Tollens.
\end{enumerate}
}
\vspace{2pt}

\noindent 
(\ref{DerivationsBetweenOBandPE}.a) and (\ref{DerivationsBetweenOBandPE}.c) can be implemented via the single SPARQL rule shown in (\ref{SPARQLrelatingObligatoryAndPermitted1}), because the two {\tt UNION} operators used therein respectively refer to the opposite eventuality {\tt ?ne} of a given eventuality {\tt ?e} and bind the variable {\tt ?ddm} to the dual deontic modality of the one of which {\tt ?e} is an instance: if the latter is an instance of {\tt Obligatory}, {\tt ?ddm} is bound to {\tt Permitted} and vice versa.

\enumsentence{\label{SPARQLrelatingObligatoryAndPermitted1}
\begin{minipage}[t]{450pt}\tt
\mbox{}\hspace{0pt}[a :InferenceRule; :has-sparql-code\hspace{-3pt} """\\
\mbox{}\hspace{16pt}CONSTRUCT\{\hspace{-1pt}[a :false,:hold;\\
\mbox{}\hspace{70pt}rdf:subject\hspace{-1pt} ?ne;\hspace{-3pt} rdf:predicate\hspace{-1pt} rdf:type;\hspace{-3pt} rdf:object\hspace{-3pt} ?ddm]\hspace{-1pt}\}\\
\mbox{}\hspace{16pt}WHERE\{\hspace{-1pt}\{?e :not ?ne\}UNION\{?ne :not ?e\}\\
\mbox{}\hspace{51pt}\{?e a :Obligatory. BIND(:Permitted AS ?ddm)\}UNION\\
\mbox{}\hspace{51pt}\{?e a :Permitted. BIND(:Obligatory AS ?ddm)\}\\
\mbox{}\hspace{54pt}NOT EXISTS\{\hspace{-1pt}?f a :false,:hold; rdf:subject\hspace{-1pt} ?ne;\\
\mbox{}\hspace{110pt}rdf:predicate\hspace{-1pt} rdf:type;\hspace{-3pt} rdf:object\hspace{-3pt} ?ddm\hspace{-1pt}\}\}"""]\hspace{-1pt}.
\end{minipage}
}
\vspace{4pt}

\noindent Similarly, the derivations in (\ref{DerivationsBetweenOBandPE}.b) and (\ref{DerivationsBetweenOBandPE}.d) can be implemented via the following SPARQL rule:

\enumsentence{\label{SPARQLrelatingObligatoryAndPermitted2}
\begin{minipage}[t]{450pt}\tt
\mbox{}\hspace{0pt}[a :InferenceRule; :has-sparql-code\hspace{-3pt} """\\
\mbox{}\hspace{16pt}CONSTRUCT\{\hspace{-1pt}?ne a ?ddm\}\\
\mbox{}\hspace{16pt}WHERE\{\hspace{-1pt}\{?e :not ?ne\}UNION\{?ne :not ?e\}\\
\mbox{}\hspace{52pt}?r a :false,:hold; rdf:subject\hspace{-1pt} ?e;\hspace{-1pt} rdf:predicate\hspace{-1pt} rdf:type.\\
\mbox{}\hspace{51pt}\{?r rdf:object ?dm. BIND(:Permitted AS ?ddm)\}UNION\\
\mbox{}\hspace{51pt}\{?r rdf:object ?dm. BIND(:Obligatory AS ?ddm)\}\}"""]\hspace{-1pt}.
\end{minipage}
}
\vspace{5pt}

\noindent The two entailments in (\ref{DeonticTraditionalSchemeEntailmentsRepeated}.b) correspond to the following  four derivations:
 
\enumsentence{\label{DerivationsBetweenOPandOB}
\begin{enumerate}
    \renewcommand{\labelenumi}{\alph{enumi}.}    
    \item\noindent {\tt OP(p)}\hspace{1pt}$\vdash$\hspace{1pt}{\tt ($\neg$OB(p)}\hspace{1pt}$\wedge$ {\tt $\neg$OB($\neg$p)\hspace{-2pt})}, via Modus Ponens.
    \vspace{2pt}
    \item {\tt ($\neg$OB(p)}\hspace{1pt}$\wedge$ {\tt $\neg$OB($\neg$p)\hspace{-2pt})}\hspace{1pt}$\vdash$\hspace{1pt}{\tt OP(p)}, via Modus Ponens.
    \vspace{2pt}
    \item {\tt (OB(p)}\hspace{1pt}$\vee$ {\tt OB($\neg$p)\hspace{-2pt})}\hspace{1pt}$\vdash$\hspace{3pt}{\tt $\neg$\hspace{1pt}OP(p)}, via Modus Tollens.
    \vspace{2pt}
    \item {\tt $\neg$\hspace{1pt}OP(p)}\hspace{1pt}$\vdash$\hspace{1pt}{\tt (OB(p)}\hspace{1pt}$\vee$ {\tt OB($\neg$p)\hspace{-2pt})}, via Modus Tollens.
\end{enumerate}
}
\vspace{2pt}

\noindent Nevertheless, contrary to the entailments in (\ref{DeonticTraditionalSchemeEntailmentsRepeated}.a), here it is not possible to write a single SPARQL rule to implement a pair of these entailments, because the two entailments in the pair involve two different connectives ($\wedge$ and $\vee$).

The four entailments in (\ref{DerivationsBetweenOPandOB}.a-d) are respectively implemented by the SPARQL rules shown in (\ref{SPARQLrelatingOptionalAndObligatory1}), (\ref{SPARQLrelatingOptionalAndObligatory2}), (\ref{SPARQLrelatingOptionalAndObligatory3}), and (\ref{SPARQLrelatingOptionalAndObligatory4}); for practical reasons, we chose to implement (\ref{DerivationsBetweenOPandOB}.a) and (\ref{DerivationsBetweenOPandOB}.c) via {\it two} SPARQL rules each, i.e., we decided to separate the two conjuncts in the consequent of (\ref{DerivationsBetweenOPandOB}.a) and the two disjuncts in the antecedent of (\ref{DerivationsBetweenOPandOB}.c).

\enumsentence{\label{SPARQLrelatingOptionalAndObligatory1}

\begin{minipage}[t]{450pt}\tt
\mbox{}\hspace{0pt}[a :InferenceRule; :has-sparql-code\hspace{-3pt} """\\
\mbox{}\hspace{16pt}CONSTRUCT\{\hspace{-1pt}[a :false,:hold; rdf:subject\hspace{-1pt} ?e;\\
\mbox{}\hspace{81pt}rdf:predicate\hspace{-1pt} rdf:type;\hspace{-3pt} rdf:object\hspace{-3pt} :Obligatory]\hspace{-1pt}\}\\
\mbox{}\hspace{16pt}WHERE\{\hspace{-1pt}?e a :Optional. \{?e :not ?ne\}UNION\{?ne :not ?e\}\\
\mbox{}\hspace{52pt}NOT EXISTS\{\hspace{-1pt}?f a :false,:hold; rdf:subject ?e;\\
\mbox{}\hspace{117pt}rdf:predicate\hspace{-1pt} rdf:type; rdf:object\hspace{-3pt} :Obligatory\hspace{-1pt}\}\}"""]\hspace{-1pt}.
\end{minipage}\\[5pt]

\begin{minipage}[t]{450pt}\tt
\mbox{}\hspace{0pt}[a :InferenceRule; :has-sparql-code\hspace{-3pt} """\\
\mbox{}\hspace{16pt}CONSTRUCT\{\hspace{-1pt}[a :false,:hold; rdf:subject\hspace{-1pt} ?ne;\\
\mbox{}\hspace{81pt}rdf:predicate\hspace{-1pt} rdf:type;\hspace{-3pt} rdf:object\hspace{-3pt} :Obligatory]\hspace{-1pt}\}\\
\mbox{}\hspace{16pt}WHERE\{\hspace{-1pt}?e a :Optional. \{?e :not ?ne\}UNION\{?ne :not ?e\}\\
\mbox{}\hspace{52pt}NOT EXISTS\{\hspace{-1pt}?f a :false,:hold; rdf:subject ?ne;\\
\mbox{}\hspace{117pt}rdf:predicate\hspace{-1pt} rdf:type; rdf:object\hspace{-3pt} :Obligatory\hspace{-1pt}\}\}"""]\hspace{-1pt}.
\end{minipage}

}

\enumsentence{\label{SPARQLrelatingOptionalAndObligatory2}
\begin{minipage}[t]{450pt}\tt
\mbox{}\hspace{0pt}[a :InferenceRule; :has-sparql-code\hspace{-3pt} """\\
\mbox{}\hspace{16pt}CONSTRUCT\{\hspace{-1pt}?e a :Optional. ?ne a :Optional\}\\
\mbox{}\hspace{16pt}WHERE\{\hspace{-1pt}\{?e :not ?ne\}UNION\{?ne :not ?e\}\\
\mbox{}\hspace{52pt}?r1 a :false,:hold; rdf:subject\hspace{-1pt} ?e;\\
\mbox{}\hspace{100pt} rdf:predicate\hspace{-1pt} rdf:type; rdf:object  :Obligatory.\\
\mbox{}\hspace{52pt}?r2 a :false,:hold; rdf:subject\hspace{-1pt} ?ne;\\
\mbox{}\hspace{100pt} rdf:predicate\hspace{-1pt} rdf:type; rdf:object  :Obligatory\}"""]\hspace{-1pt}.
\end{minipage}
}

\enumsentence{\label{SPARQLrelatingOptionalAndObligatory3}
\begin{minipage}[t]{450pt}\tt
\mbox{}\hspace{0pt}[a :InferenceRule; :has-sparql-code\hspace{-3pt} """\\
\mbox{}\hspace{16pt}CONSTRUCT\{\hspace{-1pt}[a :false,:hold; rdf:subject\hspace{-1pt} ?e;\\
\mbox{}\hspace{81pt}rdf:predicate\hspace{-1pt} rdf:type;\hspace{-3pt} rdf:object\hspace{-3pt} :Optional]\hspace{-1pt}\}\\
\mbox{}\hspace{16pt}WHERE\{?e a :Obligatory. \{?e :not ?ne\}UNION\{?ne :not ?e\}\\
\mbox{}\hspace{54pt}NOT EXISTS\{?f a :false,:hold; rdf:subject ?e;\\
\mbox{}\hspace{120pt}rdf:predicate\hspace{-1pt} rdf:type; rdf:object\hspace{-3pt} :Optional\}\}"""]\hspace{-1pt}.
\end{minipage}\\[5pt]

\begin{minipage}[t]{450pt}\tt
\mbox{}\hspace{0pt}[a :InferenceRule; :has-sparql-code\hspace{-3pt} """\\
\mbox{}\hspace{16pt}CONSTRUCT\{\hspace{-1pt}[a :false,:hold; rdf:subject\hspace{-1pt} ?ne;\\
\mbox{}\hspace{81pt}rdf:predicate\hspace{-1pt} rdf:type;\hspace{-3pt} rdf:object\hspace{-3pt} :Optional]\hspace{-1pt}\}\\
\mbox{}\hspace{16pt}WHERE\{?e a :Obligatory. \{?e :not ?ne\}UNION\{?ne :not ?e\}\\
\mbox{}\hspace{54pt}NOT EXISTS\{?f a :false,:hold; rdf:subject ?ne;\\
\mbox{}\hspace{120pt}rdf:predicate\hspace{-1pt} rdf:type; rdf:object\hspace{-3pt} :Optional\}\}"""]\hspace{-1pt}.
\end{minipage}
}

\enumsentence{\label{SPARQLrelatingOptionalAndObligatory4}
\begin{minipage}[t]{450pt}\tt
\mbox{}\hspace{0pt}[a :InferenceRule; :has-sparql-code\hspace{-3pt} """\\
\mbox{}\hspace{16pt}CONSTRUCT\{\hspace{-1pt}[a :true;  rdf:subject ?e; rdf:predicate\hspace{-1pt} rdf:type;\\
\mbox{}\hspace{76pt}rdf:object\hspace{-4pt} :Obligatory]\hspace{-6pt} :\hspace{-1pt}disjunction\hspace{-4pt} [a\hspace{-2pt} :true;\hspace{-2pt}  rdf:subject\hspace{-2pt} ?ne;\\
\mbox{}\hspace{76pt}rdf:predicate\hspace{-1pt} rdf:type;\hspace{-2pt} rdf:object\hspace{-2pt} :Obligatory\hspace{-1pt}]\hspace{-1pt}\}\\
\mbox{}\hspace{16pt}WHERE\{\{?e :not ?ne\}UNION\{?ne :not ?e\} ?r a :false,:hold;\\
\mbox{}\hspace{53pt}rdf:subject\hspace{-1pt} ?e;\hspace{-1pt} rdf:predicate\hspace{-1pt} rdf:type;\hspace{-1pt} rdf:object\hspace{-1pt} :Optional\hspace{-1pt}.\\
\mbox{}\hspace{53pt}NOT EXISTS\{?e a :Obligatory\} NOT EXISTS\{?ne a :Obligatory\}\\
\mbox{}\hspace{53pt}NOT EXISTS\{\hspace{-1pt}\{?r1 :disjunction ?r2\}UNION\{?r2 :disjunction ?r1\}\\
\mbox{}\hspace{70pt}?r1 a :true; rdf:subject ?e; rdf:predicate rdf:type;\\
\mbox{}\hspace{70pt}rdf:object :Obligatory. ?r2 a :true; rdf:subject ?ne;\\
\mbox{}\hspace{70pt}rdf:predicate rdf:type; rdf:object :Obligatory.\}\}"""]\hspace{-1pt}.
\end{minipage}
}
\vspace{5pt}

\noindent Finally, the single entailment in (\ref{DeonticTraditionalSchemeEntailmentsRepeated}.c) corresponds to the following two derivations:

\enumsentence{\label{DerivationsFromOBtoPE}
\begin{enumerate}
    \renewcommand{\labelenumi}{\alph{enumi}.}    
    \item\noindent {\tt OB(p)}\hspace{1pt}$\vdash$\hspace{3pt}{\tt PE(p)}, via Modus Ponens.
    \vspace{2pt}
    \item $\neg$\hspace{1pt}{\tt PE(p)}\hspace{1pt}$\vdash$\hspace{3pt}$\neg$\hspace{1pt}{\tt OB(p)}, via Modus Tollens.
\end{enumerate}
}
\vspace{2pt}

\noindent And, in turn, (\ref{DerivationsFromOBtoPE}.a) ad (\ref{DerivationsFromOBtoPE}.b) respectively correspond to the two following SPARQL rules:

\enumsentence{\label{SPARQLruleFromOBtoPE1}
\begin{minipage}[t]{450pt}\tt
\mbox{}\hspace{0pt}[a :InferenceRule; :has-sparql-code\hspace{-3pt} """\\
\mbox{}\hspace{20pt}CONSTRUCT\{\hspace{-1pt}?e a :Permitted\} WHERE\{\hspace{-1pt}?e a :Obligatory\}"""]\hspace{-1pt}.

\end{minipage}
}

\enumsentence{\label{SPARQLruleFromOBtoPE2}
\begin{minipage}[t]{450pt}\tt
\mbox{}\hspace{0pt}[a :InferenceRule; :has-sparql-code\hspace{-3pt} """\\
\mbox{}\hspace{16pt}CONSTRUCT\{\hspace{-1pt}[a :false,:hold; rdf:subject ?e;\\
\mbox{}\hspace{71pt}rdf:predicate rdf:type; rdf:object :Obligatory]\hspace{-1pt}\}\\
\mbox{}\hspace{16pt}WHERE\{\hspace{-1pt}?r a :false, :hold;\\ 
\mbox{}\hspace{32pt}rdf:subject ?e; rdf:predicate rdf:type; rdf:object :Permitted.\\
\mbox{}\hspace{32pt}NOT EXISTS\{?f a :false,:hold; rdf:subject ?e;\\
\mbox{}\hspace{82pt}
rdf:predicate rdf:type; rdf:object :Obligatory\hspace{-1pt}\}\hspace{-1pt}"\hspace{-1pt}"\hspace{-1pt}"]\hspace{-1pt}.

\end{minipage}
}

\noindent The GitHub repository contains some examples involving the SPARQL rules shown in this subsection. In addition, these rules will be also used in several derivations discussed below.

\subsection{Modelling conflicts between deontic modalities}
\vspace{-7pt}

\noindent In order to model conflicts between deontic modalities, we studied in depth the seminal work in \cite{Goble:13}, which is perhaps the main survey about the state of the art on this topic. However, as explained in (\ref{intuitions}.b) above, \cite{Goble:13} only consider conflicts between pairs of obligations while it does not consider conflicts between  obligations/prohibitions and permissions. Nor it is possible to reconcile this second category of conflicts with the first one because permissions  do no entail obligations, they only entail not-obligations, as stipulated by (\ref{DeonticTraditionalSchemeEntailmentsRepeated}.a) above. 

Furthermore, even by considering only conflicts between obligations, we do not agree with the definition given in \cite{Goble:13}, i.e., that a conflict of obligations is a situation in which ``an agent ought to do a number of things, each of which is possible for the agent, but it is impossible for the agent to do them all''.
This paper instead follows Kelsen's work (\cite{Kelsen:91}, \cite{Vranes:06}) and so it defines a conflicts of deontic statements as a situation in which two or more deontic statements hold in the context but complying with one of them entails violating (or not permitting) another one.

This alternative intuition/definition is grounded on the observation that obligations and prohibitions stipulate what an agent {\it must or must not} do, but this is unrelated to what an agent {\it can or cannot} do. It appears to be  instead related to the notion of {\it violation}, i.e., when agents {\it do not or do} what they respectively  {\it must or must not} do. 

Nevertheless, despite its intimate relation with obligations and prohibitions, the notion of violation, as well as its corresponding formalization, has been by and large neglected in past literature in deontic logic, while this paper argues that it should be taken as the starting point. For this reason, the next subsubsection will identify and formalize the notion of violation in the proposed computational ontology. After that, we will reason on this notion in order to identify and formalize the notion of conflict between deontic statements.

\subsubsection{Understanding and formalizing violations}\label{violations}
\vspace{2pt}

\noindent Under the above assumption, the starting point of our reasoning is to understand which eventualities either comply with or violate obligations or prohibitions, and how this might be implemented in the proposed computational ontology. Consider the sentences in (\ref{johnpaid}.a) and (\ref{johnpaid}.c), respectively formalized in Hobbs's as in (\ref{johnpaid}.b) and (\ref{johnpaid}.d):

\enumsentence{\label{johnpaid}
\begin{enumerate}
    \renewcommand{\labelenumi}{\alph{enumi}.}
    \item John pays £3.
    \vspace{2pt}
    \item {\tt Rexist}({\tt e£3}) $\wedge$ {\tt Pay}({\tt e£3}) $\wedge$ {\tt has-agent}({\tt e£3}, {\tt John}) $\wedge$ {\tt has-object}({\tt e£3}, {\tt £3})
    \vspace{5pt}
    \item John pays in cash.
    \vspace{2pt}
    \item {\tt Rexist}({\tt ec}) $\wedge$ {\tt Pay}({\tt ec}) $\wedge$ {\tt has-agent}({\tt ec}, {\tt John}) $\wedge$ {\tt has-instrument}({\tt ec}, {\tt cash})
\end{enumerate}
}

\noindent (\ref{johnpaid}.b) and (\ref{johnpaid}.d) include two individual eventualities, {\tt e£3} and {\tt ec}, that really exist in the state of affairs. Although the instrument of (\ref{johnpaid}.a) as well as the object of (\ref{johnpaid}.c) are unknown, it is clear that they are also {\it specific} individual entities of the domain: in (\ref{johnpaid}.a), John pays £3 {\it in a specific (unknown) way} while, in  (\ref{johnpaid}.c), John pays {\it a specific (unknown) amount of money} in cash. This is in line with the Open World Assumption.

Consider now the two sentences in (\ref{johnisobligedorprohibitedtopaid}.a) and (\ref{johnisobligedorprohibitedtopaid}.c), formalized as in (\ref{johnisobligedorprohibitedtopaid}.b) and (\ref{johnisobligedorprohibitedtopaid}.d)

\enumsentence{\label{johnisobligedorprohibitedtopaid}
\begin{enumerate}
    \renewcommand{\labelenumi}{\alph{enumi}.}
    \item John is obliged to pay £3.
    \vspace{2pt}
    \item {\tt Obligatory}({\tt eo}) $\wedge$ {\tt Pay}({\tt eo}) $\wedge$ {\tt has-agent}({\tt eo}, {\tt John}) $\wedge$ {\tt has-object}({\tt eo}, {\tt £3})
    \vspace{5pt}
    \item John is prohibited to pay in cash.
    \vspace{2pt}
    \item $\neg$\hspace{1pt}{\tt Permitted}\hspace{1pt}({\tt enp}) $\wedge$ {\tt Pay}({\tt enp}) $\wedge$ {\tt has-agent}({\tt enp}, {\tt John}) $\wedge$\\ 
    {\tt has-instrument}({\tt enp}, {\tt cash})
\end{enumerate}
}

\noindent Although in (\ref{johnisobligedorprohibitedtopaid}.b) and (\ref{johnisobligedorprohibitedtopaid}.d) the instrument is  also unknown, in this case it is {\it not} a specific individual entity: John is obliged to pay £3 {\it in any way}, e.g., in cash rather than by card. Similarly, in (\ref{johnisobligedorprohibitedtopaid}.b), John is prohibited to pay {\it any amount of money} in cash, i.e., the prohibition is violated if John pay in cash £3, or £4, or £5, etc.

In light of this, it appears evident that obligations and prohibitions correspond to (what \cite{Gordon-Hobbs:17} term as) ``abstract eventualities'', discussed above in subsection \ref{AbstractEventualitiesHobbs} and subsubsection
\ref{AbstractEventualitiesRDFsAndSPARQL}: what respectively complies with or violates {\tt eo} and {\tt enp} in (\ref{johnisobligedorprohibitedtopaid}) are individual eventualities that really exist and that instantiate {\tt eo} and {\tt enp}; however, since multiple instantiations of an abstract eventuality are possible, {\tt eo} and {\tt enp} can be (respectively) complied with or violated by any eventuality in the set of {\tt eo}'s and {\tt enp}'s instantiations.

Therefore, this paper  stipulates that an obligatory eventuality is complied with by an eventuality that really exists if the latter  instantiates the former. To check so in RDF and SPARQL, with respect to the pattern chosen to encode the statements (shown above in
(\ref{patternFormulae})), it must be checked that:

\enumsentence{\label{InferringSpecificEventualitiesInSPARQL}
\begin{enumerate}
    \renewcommand{\labelenumi}
    {\alph{enumi}.}
    
    \item\noindent The two eventualities are instances of the same instance of {\tt Eventuality}.
    
    \vspace{5pt}
    \item\noindent Every thematic role specified for the obligatory eventuality is also specified for the  eventuality that really exists and it has the same value. This check cannot be directly implemented in SPARQL; however, it is equivalent to the conjunction of the following two checks, each of which can be implemented in SPARQL:

    \begin{enumerate}
    \renewcommand{\labelenumii}
    {\roman{enumii}.}
    \item No thematic role is specified for the obligatory eventuality but not for the eventuality that really exists.
    \vspace{3pt}
    \item No thematic role is specified for both eventualities but it has a different value for each of them.
    \end{enumerate}
\end{enumerate}
}
\vspace{3pt}

\noindent If (\ref{InferringSpecificEventualitiesInSPARQL}.a-b) hold, it is inferred that the obligatory eventuality is complied with by the eventuality that really exists; this is formalized by introducing a new RDF property {\tt is-complied-with-by} between the reifications of the two eventualities. The SPARQL rule implementing the described inference is:

\enumsentence{\label{SPARQLiscompliedwithbyObligatoryRexist}
\begin{minipage}[t]{550pt}\tt
[a :InferenceRule; :has-sparql-code """\\
\mbox{}\hspace{-10pt}CONSTRUCT\hspace{-1pt}\{[a :true,:hold; rdf:subject\hspace{-2pt} ?eo; rdf\hspace{-1pt}:\hspace{-1pt}predicate\hspace{-1pt} rdf\hspace{-1pt}:\hspace{-1pt}type;\\
\mbox{}\hspace{30pt}rdf\hspace{-1pt}:\hspace{-1pt}object\hspace{-3pt} :Obligatory\hspace{-1pt}] :is-complied-with-by [a :true,:hold;\\
\mbox{}\hspace{30pt}rdf:subject\hspace{-2pt} ?e;\hspace{-4pt} rdf\hspace{-1pt}:\hspace{-1pt}predicate\hspace{-1pt} rdf\hspace{-1pt}:\hspace{-1pt}type;\hspace{-4pt} rdf\hspace{-1pt}:\hspace{-1pt}object\hspace{-3pt} :Rexist\hspace{-1pt}]\}\\
\mbox{}\hspace{-10pt}WHERE\{?eo\hspace{-1pt} a\hspace{-1pt} :Obligatory, ?c. ?e\hspace{-1pt} a\hspace{-1pt} :Rexist, ?c. ?c a :Eventuality.\\
\mbox{}\hspace{-6pt}NOT\hspace{-2pt} EXISTS\hspace{-1pt}\{\hspace{-1pt}?tr\hspace{-2pt} a\hspace{-2pt} :\hspace{-1pt}ThematicRole.\hspace{-4pt} ?eo\hspace{-1pt} ?tr\hspace{-2pt} ?vo.\hspace{-3pt} NOT\hspace{-1pt} EXISTS\hspace{-1pt}\{?e\hspace{-1pt} ?tr\hspace{-1pt} ?ve\}\}\\
\mbox{}\hspace{-6pt}NOT\hspace{-2pt} EXISTS\hspace{-1pt}\{\hspace{-1pt}?tr\hspace{-2pt} a\hspace{-2pt} :\hspace{-1pt}ThematicRole.\hspace{-4pt} ?eo\hspace{-1pt} ?tr\hspace{-2pt} ?vo.\hspace{-3pt} ?e\hspace{-2pt} ?tr\hspace{-2pt} ?ve.\hspace{-3pt} FILTER\hspace{-1pt}(?vo!=?ve)\hspace{-2pt}\}\\
\mbox{}\hspace{-6pt}NOT\hspace{-3pt} EXISTS\hspace{-1pt}\{?eor :is-complied-with-by ?er. ?eor a :true,:hold;\\
\mbox{}\hspace{20pt}rdf:subject\hspace{-2pt} ?eo;\hspace{-2pt} rdf:predicate\hspace{-2pt} rdf:type;\hspace{-2pt} rdf:object\hspace{-2pt} :Obligatory.\\
\mbox{}\hspace{20pt}?er a :true,:hold; rdf:subject\hspace{-2pt} ?e;\\
\mbox{}\hspace{20pt}rdf:predicate\hspace{-2pt} rdf:type;\hspace{-2pt} rdf:object\hspace{-2pt} :Rexist\}\}"""]\hspace{-1pt}.
\end{minipage}
}
\vspace{6pt}

\noindent The SPARQL rule in (\ref{SPARQLiscompliedwithbyObligatoryRexist}) is indeed able to infer that the RDF triples encoding 
(\ref{johnpaid}.a-b) comply with the RDF triples encoding 
(\ref{johnisobligedorprohibitedtopaid}.a-b); the rule does not instead infer so for the RDF triples encoding (\ref{johnpaid}.c-d), in which the amount of money paid by John is unknown.

To deal with prohibitions, i.e., non-permissions, we introduce a dual SPARQL rule, shown in (\ref{SPARQLisviolatedbyNotPermittedRexist}): if an eventuality is prohibited and another eventuality that instantiates the former really exists, then it is inferred that the former {\tt is-violated-by} the latter.

\enumsentence{\label{SPARQLisviolatedbyNotPermittedRexist}
\begin{minipage}[t]{550pt}\tt
[a :InferenceRule; :has-sparql-code """\\
\mbox{}\hspace{-10pt}CONSTRUCT\{\hspace{1pt}?epr :is-violated-by [a :true,:hold;\\
\mbox{}\hspace{31pt}rdf:subject\hspace{-2pt} ?e;\hspace{-4pt} rdf\hspace{-1pt}:\hspace{-1pt}predicate\hspace{-1pt} rdf\hspace{-1pt}:\hspace{-1pt}type;\hspace{-4pt} rdf\hspace{-1pt}:\hspace{-1pt}object\hspace{-3pt} :Rexist\hspace{-1pt}]\}\\
\mbox{}\hspace{-10pt}WHERE\{?epr\hspace{-1pt} a\hspace{-1pt} :false,:hold;\\
\mbox{}\hspace{-6pt}rdf:subject ?ep; rdf:predicate rdf:type; rdf:object :Permitted.\\
\mbox{}\hspace{-6pt}?ep a ?c. ?e\hspace{-1pt} a\hspace{-1pt} :Rexist,\hspace{-4pt} ?c. ?c a :Eventuality.\\
\mbox{}\hspace{-6pt}NOT\hspace{-2pt} EXISTS\hspace{-1pt}\{\hspace{-1pt}?tr\hspace{-2pt} a\hspace{-2pt} :\hspace{-1pt}ThematicRole.\hspace{-4pt} ?ep\hspace{-1pt} ?tr\hspace{-2pt} ?vp.\hspace{-3pt} NOT\hspace{-1pt} EXISTS\hspace{-1pt}\{?e\hspace{-1pt} ?tr\hspace{-1pt} ?ve\}\}\\
\mbox{}\hspace{-6pt}NOT\hspace{-2pt} EXISTS\hspace{-1pt}\{\hspace{-1pt}?tr\hspace{-2pt} a\hspace{-2pt} :\hspace{-1pt}ThematicRole.\hspace{-4pt} ?ep\hspace{-1pt} ?tr\hspace{-2pt} ?vp.\hspace{-3pt} ?e\hspace{-2pt} ?tr\hspace{-2pt} ?ve.\hspace{-3pt} FILTER\hspace{-1pt}(?vp!=?ve)\hspace{-2pt}\}\\
\mbox{}\hspace{-6pt}NOT\hspace{-3pt} EXISTS\hspace{-1pt}\{?epr :is-violated-by ?te. ?te rdf:type :true,:hold;\\
\mbox{}\hspace{20pt}rdf:subject\hspace{-2pt} ?e;\hspace{-2pt} rdf:predicate\hspace{-2pt} rdf:type;\hspace{-2pt} rdf:object\hspace{-2pt} :Rexist\}\}"""]\hspace{-1pt}.
\end{minipage}
}
\vspace{6pt}

\noindent The SPARQL rule in (\ref{SPARQLisviolatedbyNotPermittedRexist}) is indeed able to infer that the RDF triples encoding (\ref{johnpaid}.c-d) violate the RDF triples encoding 
(\ref{johnisobligedorprohibitedtopaid}.c-d); conversely, the rule does not infer so for the RDF triples encoding (\ref{johnpaid}.a-b), in which the instrument used by John for paying £3 is unknown.

This paper omits the RDF triples encoding (\ref{johnpaid}.a-d) and (\ref{johnisobligedorprohibitedtopaid}.a-d) as well as the RDF triples inferred from them through the rules in (\ref{SPARQLiscompliedwithbyObligatoryRexist}) and (\ref{SPARQLisviolatedbyNotPermittedRexist}); these, together with several other examples, are however available on GitHub. 

On the other hand, we discuss now a full example that involves the SPARQL rules presented in this subsubsection and the ones seen in the previous subsection, which implement the Deontic Traditional Scheme. The example is shown in (\ref{johnisprohibitedtonotpay}). Sentence (\ref{johnisprohibitedtonotpay}.a), encoded in Hobbs's as in (\ref{johnisprohibitedtonotpay}.b) and in RDF as in (\ref{johnisprohibitedtonotpay}.c), is complied with by sentence (\ref{johnisprohibitedtonotpay}.d), encoded in Hobbs's as in (\ref{johnisprohibitedtonotpay}.e) and in RDF as in (\ref{johnisprohibitedtonotpay}.f). In fact, the meaning of sentence (\ref{johnisprohibitedtonotpay}.a) is that John is not permitted to not pay, therefore that he is obliged to pay. This obligation is complied by (\ref{johnisprohibitedtonotpay}.d).

\enumsentence{\label{johnisprohibitedtonotpay}
\begin{enumerate}
    \renewcommand{\labelenumi}{\alph{enumi}.}
    \item John is prohibited to not pay.
    \vspace{2pt}
    \item $\neg$\hspace{1pt}{\tt Permitted}\hspace{1pt}({\tt enpj}) $\wedge$ {\tt not}({\tt enpj}, {\tt epj}) $\wedge$ {\tt Pay}({\tt epj}) $\wedge$
    {\tt has-agent}({\tt epj}\hspace{1pt}, {\tt John})
    
    \vspace{3pt}
    \item{\tt \hspace{-1pt}[a :false,:hold; rdf:subject soa:enpj; rdf:predicate rdf:type;\\ 
    \mbox{}\hspace{5pt}rdf:object :Permitted]\hspace{-1pt}. soa:enpj :not soa:epj.}\\
    \mbox{}\hspace{2pt}{\tt  soa:epj a soa:Pay; soa:has-agent soa:John.}
    
    \vspace{6pt}
    \item John pays £3.
    \vspace{2pt}
    \item {\tt Rexist}({\tt epj3}) $\wedge$ {\tt Pay}({\tt epj3}) $\wedge$ {\tt has-agent}({\tt epj3}, {\tt John})
    \vspace{3pt}
    \item{\tt soa:epj3 a :Rexist,soa:Pay; soa:has-agent soa:John.}
\end{enumerate}
}

\noindent Inferring that (\ref{johnisprohibitedtonotpay}.a-c) is complied with by (\ref{johnisprohibitedtonotpay}.d-f) requires the SPARQL rules that implement the Deontic Traditional Scheme, specifically the rule shown above in 
(\ref{SPARQLrelatingObligatoryAndPermitted2}). This rule derives {\tt OB($\neg$p)} from {\tt $\neg$PE(p)} via Modus Tollens on {\tt $\neg$OB($\neg$p)}\hspace{-1pt}$\rightarrow$\hspace{1pt}{\tt PE(p)}, thus it adds the following RDF triple to the inferred knowledge graph:

\enumsentence{\label{johnisprohibitedtonotpay1stinference}\tt
\hspace{5pt}soa:epj a :Obligatory.
}

\noindent From this triple and the ones in (\ref{johnisprohibitedtonotpay}.f), the SPARQL rule in (\ref{SPARQLiscompliedwithbyObligatoryRexist}) infers that the fact that the triple in (\ref{johnisprohibitedtonotpay1stinference}) holds true is complied with by the fact that also the triple ``{\tt soa:epj3 a :Rexist}'' holds true, encoded in RDF as follows:

\enumsentence{\label{johnisprohibitedtonotpay2ndinference}\tt
\hspace{4pt}{\tt [a :true,:hold; rdf:subject soa:epj; rdf:predicate rdf:type;\\ 
\mbox{}\hspace{10pt}rdf:object :Obligatory] is-complied-with-by [a :true,:hold;}\\ 
\mbox{}\hspace{10pt}{\tt rdf:subject soa:epj3; rdf:predicate rdf:type; rdf:object :Rexist]\hspace{-1pt}.}
}

\noindent Before proceeding to the next subsubsection, we wish to clarify when and how it is possible to infer that an obligation has been violated and, symmetrically, that a prohibition has been complied with.

It is possible to infer that an obligation has been violated only when it has not been complied with by {\it any} of its instantiations. However, under the Open World Assumption, e.g., in RDF but also, indeed, in the real world, it could be extremely difficult to prove so. Suppose for instance that John is obliged to pay £3 and that the police must demonstrate whether
John violated this obligations or not. Suppose also that the knowledge graph does not contain any triple asserting that John paid £3; according to the Open World Assumption, this does not mean that John truly did not, it simply means that it is {\it unknown} whether he did it or not. The police detectives could then demonstrate the violation by investigating whether John made a payment at all. In other words, they could demonstrate that John did not make any payment in cash, that John did not make any payment by card, and so forth for all other possible instruments with which John could have made that payment. If the police detectives manage to demonstrate {\it all} of these, then they can conclude that, no, John did not pay £3 and so that he violated his obligation.

Precisely because investigating whether obligations have been violated or not could be very time- and resource-consuming, many existing norms from legislation requires burden of proof\hspace{1pt}\footnote{An example is 
Art. 136 of the UK Equality Act 2010, available at \url{https://www.legislation.gov.uk/ukpga/2010/15/section/136}. An approach that tries to formalize burden of proof in legal reasoning is \cite{Satoh:12}.}: the agent does not only have to comply with the obligations, he must also provide {\it evidence of compliance}. In cases where they do not provide such evidence, even if they did in fact comply with their obligations, it is (abductively) established that they did not. This might be easily implemented through a SPARQL rule that, by using the {\tt NOT\hspace{4pt}EXISTS} clause, flags as violated all obligatory eventualities that do not occur as subject of {\tt is-complied-with-by}. Alternatively, the authority checking for compliance may contact the agents in charge of the obligations and enquiry them about their {\it alleged} violations.

Similar considerations hold for prohibitions: we might introduce a SPARQL rule that flags a prohibition as ``complied with'' if the knowledge graph does not include any eventuality that violates it. In case of prohibitions, however, another legal principle would justify such an (abductive) inference: {\it presumption of innocence}.

The two advocated SPARQL rules, and the legal principles supporting them, are outside the scope of the present paper, thus they were not implemented in the proposed computational ontology.

\subsubsection{Understanding and formalizing conflicts among deontic statements}
\vspace{2pt}

\noindent The previous subsubsection introduced the notion of violations of obligations and prohibitions  and showed how to formalize it in the proposed computational ontology. 

\begin{figure}[ht]
  \centering
  \includegraphics[width=0.99\linewidth]{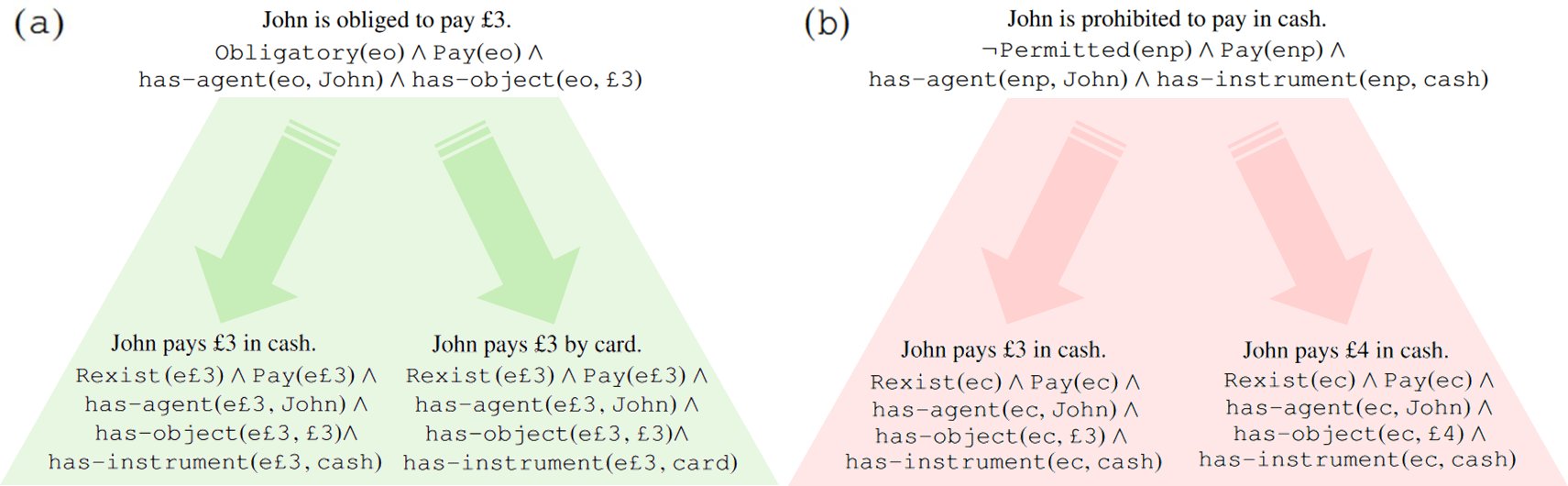}
  \caption{Compliance of obligations and violations of prohibitions}\label{ComplianceViolation}
\end{figure}

The rationale of the previous subsubsection may be depicted as in Figure \ref{ComplianceViolation}. Every obligation is associated with a ``green area'' representing the {\it set} of all really existing eventualities that instantiate the eventuality associated with the obligation: these are the eventualities that comply with the obligation. In Figure \ref{ComplianceViolation}.a, for instance, the obligation in (\ref{johnisobligedorprohibitedtopaid}.a) is complied with by any eventuality that really exists and that specifies the same thematic roles of the obligation (plus, possibly, other thematic roles, e.g., the instrument). On the other hand, every prohibition is associated with a ``red area'' representing the {\it set} of all really existing eventualities that instantiate the eventuality associated with the prohibition: these are the eventualities that violate the prohibition. In Figure \ref{ComplianceViolation}.b, for instance, the prohibition in (\ref{johnisobligedorprohibitedtopaid}.c) is violated by any eventuality that really exists and that specifies the same thematic roles of the prohibition (plus, possibly, other thematic roles, e.g., the object).

\noindent Now, in order to understand and formalize conflicts, we need to consider two eventualities that {\it both} hold for specific deontic modalities, e.g., for an obligation and a prohibition respectively. 

\begin{figure}[ht]
  \centering
  \includegraphics[width=0.95\linewidth]{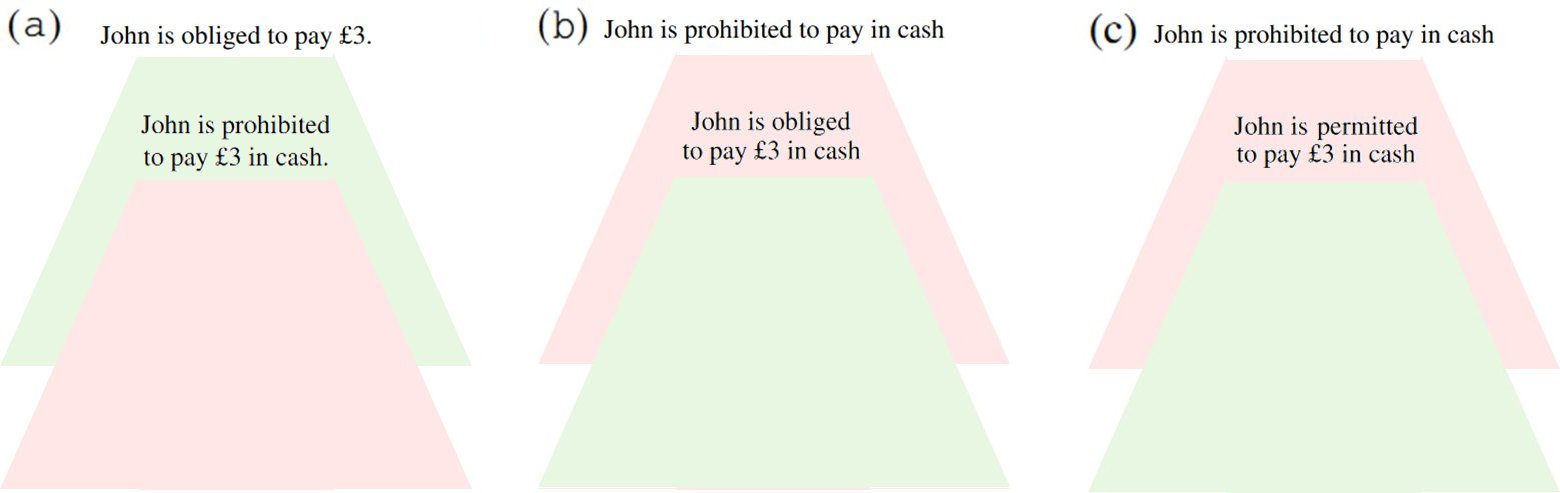}
  \caption{Compliance of obligations and violations of prohibitions}\label{conflictsObligationsProhibitionsPermissions}
\end{figure}

When a prohibited eventuality is more specific than an obligatory one, in the sense that it specifies the same action/state and the same thematic roles (with the same values) of the latter, it seems that there is no conflict: it is indeed possible to pick an eventuality in the green area but not in the red one, i.e., an eventuality that complies with the obligation without violating the prohibition. For example, in Figure \ref{conflictsObligationsProhibitionsPermissions}.a,  the fact that John pays £3 {\it by card} complies with his obligation of paying £3 without violating his prohibition of paying £3 {\it in cash}.

On the contrary, when an obligatory eventuality is more specific than a prohibited one, the red area {\it includes} the green one and so it is not possible to pick an eventuality in the green area that does not also belong to the red one. This is when a conflict occurs: any eventuality that complies with the obligation violates the prohibition. An example is shown in Figure \ref{conflictsObligationsProhibitionsPermissions}.b: the only way for John to comply with his obligation of paying £3 {\it in cash} is to violate his prohibition of using cash for his payments. Same considerations hold for permissions, although  permissions are not ``complied with'' or ``violated'' but rather ``applied'' or ``denied''. Apart from this different terminology, permissions appear to behave exactly as obligations with respect to the definition of conflicts. For instance, as exemplified in Figure \ref{conflictsObligationsProhibitionsPermissions}.c, if John executes a more specific action allowed by his permission of paying £3 in cash, then he will again violate his prohibition of paying in cash.

\noindent Since (\ref{DeonticTraditionalSchemeEntailmentsRepeated}.c), i.e., {\tt OB(p)}\hspace{-1pt}$\rightarrow$\hspace{1pt}{\tt PE(p)}, holds, both configurations of conflicts exemplified in Figure \ref{conflictsObligationsProhibitionsPermissions}.b and Figure \ref{conflictsObligationsProhibitionsPermissions}.c can be handled via a single SPARQL rule, shown in (\ref{SPARQLisinconflictwith}). This rule searches in the knowledge graph for the pattern {\tt PE}(\hspace{1pt}{\tt p}) $\wedge\hspace{2pt}\neg${\tt PE}({\tt q}\hspace{1pt}), in which {\tt p} denotes an eventuality more specific than the one denoted by {\tt q}; if this pattern is found, the rule does {\it not} assert that the two statements are in contradiction; instead, it asserts that they are {\it in conflict}. This is done by connecting the reifications of the two statements through the property {\tt is-in-conflict-with}.

\enumsentence{\label{SPARQLisinconflictwith}
\begin{minipage}[t]{550pt}\tt
[a :InferenceRule; :has-sparql-code """\\
\mbox{}\hspace{-10pt}CONSTRUCT\{?enr :is-in-conflict-with [a :true,:hold;\\
\mbox{}\hspace{30pt}rdf:subject\hspace{-2pt} ?e;\hspace{-4pt} rdf\hspace{-1pt}:\hspace{-1pt}predicate\hspace{-1pt} rdf\hspace{-1pt}:\hspace{-1pt}type;\hspace{-4pt} rdf\hspace{-1pt}:\hspace{-1pt}object\hspace{-3pt} :Permitted\hspace{-1pt}]\hspace{-1pt}\}\\
\mbox{}\hspace{-10pt}WHERE\{?enr\hspace{-1pt} a\hspace{-1pt} :false,:hold;\\
\mbox{}\hspace{-6pt}rdf:subject ?en; rdf:predicate rdf:type; rdf:object :Permitted.\\
\mbox{}\hspace{-6pt}?en a ?c. ?e\hspace{-1pt} a\hspace{-1pt} :Permitted,\hspace{-4pt} ?c. ?c a :Eventuality.\\
\mbox{}\hspace{-6pt}NOT\hspace{-2pt} EXISTS\hspace{-1pt}\{\hspace{-1pt}?tr\hspace{-2pt} a\hspace{-2pt} :\hspace{-1pt}ThematicRole.\hspace{-4pt} ?en\hspace{-1pt} ?tr\hspace{-2pt} ?vn.\hspace{-3pt} NOT\hspace{-1pt} EXISTS\hspace{-1pt}\{?e\hspace{-1pt} ?tr\hspace{-1pt} ?vp\}\}\\
\mbox{}\hspace{-6pt}NOT\hspace{-2pt} EXISTS\hspace{-1pt}\{\hspace{-1pt}?tr\hspace{-2pt} a\hspace{-2pt} :\hspace{-1pt}ThematicRole.\hspace{-4pt} ?en\hspace{-1pt} ?tr\hspace{-2pt} ?vn.\hspace{-3pt} ?e\hspace{-2pt} ?tr\hspace{-2pt} ?vp.\hspace{-3pt} FILTER\hspace{-1pt}(?vn!\hspace{-1pt}=?vp)\hspace{-2pt}\}\\
\mbox{}\hspace{-6pt}NOT\hspace{-3pt} EXISTS\hspace{-1pt}\{?enr :is-in-conflict-with ?te. ?te rdf:type :true,:hold;\\
\mbox{}\hspace{10pt}rdf:subject\hspace{-2pt} ?e;\hspace{-2pt} rdf:predicate\hspace{-2pt} rdf:type;\hspace{-2pt} rdf:object\hspace{-2pt} :Permitted\}\}"""]\hspace{-1pt}.
\end{minipage}
}
\vspace{5pt}

\noindent Let us now illustrate two examples; the GitHub repository contains these and further examples as well. Consider sentences (\ref{johnoptionalandnotpermittedtoleave}.a) and (\ref{johnoptionalandnotpermittedtoleave}.c), respectively formalized in RDF as in (\ref{johnoptionalandnotpermittedtoleave}.b) and (\ref{johnoptionalandnotpermittedtoleave}.d).

\enumsentence{\label{johnoptionalandnotpermittedtoleave}
\begin{enumerate}
    \renewcommand{\labelenumi}{\alph{enumi}.}
    \item It is optional for John to leave.
    
    \vspace{2pt}
    \item {\tt soa:elj a soa:Leave,:Optional; soa:has-agent soa:John; :not soa:enlj.}

    \vspace{5pt}
    \item John is not permitted to leave.

    \vspace{2pt}
    \item{\tt \hspace{-1pt}[a :false,:hold; rdf:subject\hspace{-1pt} soa:elj;\\ 
    \mbox{}\hspace{5pt}rdf:predicate\hspace{-1pt} rdf:type;\hspace{-3pt} rdf:object\hspace{-1pt} :Permitted]\hspace{-1pt}.}
\end{enumerate}
}

\noindent From (\ref{johnoptionalandnotpermittedtoleave}.b), the two SPARQL rules in (\ref{SPARQLrelatingOptionalAndObligatory1}) infer (\ref{johnoptionalandnotpermittedtoleavederivations}.a), i.e., that John is not obliged to leave nor to not leave, and from (\ref{johnoptionalandnotpermittedtoleavederivations}.a) the SPARQL rule in (\ref{SPARQLrelatingObligatoryAndPermitted2}) infers (\ref{johnoptionalandnotpermittedtoleavederivations}.b), i.e., that is both permitted to leave and to not leave. Finally, from (\ref{johnoptionalandnotpermittedtoleave}.d) and (\ref{johnoptionalandnotpermittedtoleavederivations}.b), the SPARQL rule in (\ref{SPARQLisinconflictwith}) infers (\ref{johnoptionalandnotpermittedtoleavederivations}.b): the fact that John is not permitted to leave is in conflict with the fact that he is permitted to leave.

\enumsentence{\label{johnoptionalandnotpermittedtoleavederivations}
\begin{enumerate}
    \renewcommand{\labelenumi}{\alph{enumi}.}
    \item
    {\tt \hspace{-1pt}[a :false,:hold; rdf:subject\hspace{-1pt} soa:elj;\\    \mbox{}\hspace{5pt}rdf:predicate\hspace{-1pt} rdf:type;\hspace{-3pt} rdf:object\hspace{-1pt} :Obligatory]\hspace{-1pt}.}\\ 
    {\tt \hspace{-1pt}[a :false,:hold; rdf:subject\hspace{-1pt} soa:enlj;\\ \mbox{}\hspace{5pt}rdf:predicate\hspace{-1pt} rdf:type;\hspace{-3pt} rdf:object\hspace{-1pt} :Obligatory]\hspace{-1pt}.}

    \vspace{4pt}
    \item {\tt elj a :Permitted. enlj a :Permitted.}

    \vspace{4pt}
    \item
    {\tt \hspace{-1pt}[a :false,:hold; rdf:subject\hspace{-1pt} soa:elj;\hspace{-3pt} rdf:predicate\hspace{-1pt} rdf:type;\\ \mbox{}\hspace{-4pt}rdf:object\hspace{-1pt} :Permitted] is-in-conflict-with [a :true,:hold;}\\ 
    \mbox{}\hspace{-4pt}{\tt rdf:subject\hspace{-1pt} soa:elj\hspace{-1pt};\hspace{-3pt} rdf:predicate\hspace{-1pt} rdf:type;\hspace{-3pt} rdf:object\hspace{-2pt} :Permitted]\hspace{-1pt}.}
\end{enumerate}
}

\noindent The second example that we illustrate is an example of partial conflicts; consider the partial conflict in (\ref{conflictingobligationsPartialOnJohn}), which is a simplified version of the one shown in (\ref{conflictingobligationsPartial}) above.

\enumsentence{\label{conflictingobligationsPartialOnJohn}
\begin{enumerate}
    \renewcommand{\labelenumi}{\alph{enumi}.}
    \item John is obliged to pay in cash.
    \vspace{5pt}
    \item John is obliged to pay by card.
\end{enumerate}
}

\noindent It has been explained in subsubsection \ref{representingAndInferringContradictions} that if the eventualities in 
(\ref{conflictingobligationsPartialOnJohn}.a-b) are represented as in (\ref{JohnPayInCashAndCard}) above, i.e., in terms of two individuals {\tt epjcash} and {\tt epjcard} that are assumed to instantiate the same abstract eventuality, and it is asserted that they both really exist, then a contradiction between the facts that they really exist is inferred.

On the other hand, if it is asserted that they are both obligatory, i.e.:

\enumsentence{\label{ExamplesConflictingThematicRolesFormalized2}
\hspace{10pt}{\tt soa:epjcash a :Obligatory.\hspace{3pt} soa:epjcard a :Obligatory.}
}

\noindent The SPARQL rules implementing the Deontic Traditional Scheme and the one in (\ref{SPARQLisinconflictwith}) infer two (symmetric) conflicts among the two obligations. In particular, the rule in (\ref{SPARQLrelatingObligatoryAndPermitted1}) infers that {\tt soa:epjcash} and {\tt soa:epjcard} are not permitted while the rule in 
(\ref{SPARQLruleFromOBtoPE1}) infers that they are permitted; then, the rule in (\ref{SPARQLisinconflictwith}) infers the two (symmetric) conflicts. The RDF representation of the two conflicts is exactly as the one in (\ref{johnoptionalandnotpermittedtoleavederivations}.c) but for the involved eventualities ({\tt epjcash} and {\tt epjcard} in place of {\tt elj}).

\subsection{Modelling norms}\label{ModellingNorms}

\noindent The third intuition that this paper assumes to be valid for modelling and reasoning with deontic modalities is that deontic modalities always have an agent. In the proposed computational ontology, this means that every eventuality that holds for at least one deontic modality will always occur as the subject of the {\tt has-agent} thematic role.

This assumption is not new in past literature in deontic logic, as explained in (\ref{intuitions}.c) above. Furthermore, regardless of this literature, it appears reasonable to make this assumption at least for norms taken from existing legislation: every obligation or prohibition from existing legislation must always specify {\it who} is the so-called ``bearer'' of that obligation or prohibition, i.e., who must comply with it. Same considerations hold for permissions, although these cannot be complied with or violated and nobody is sanctioned if they do not do what they are permitted to do.

Sometimes the bearers of the norms are {\it implicit}, although they can be easily deducted from the context of the norms, even for hypothetical norms such as the very first ones considered in this paper and repeated again in (\ref{conflictingobligations1repeated}) for reader's convenience:

\enumsentence{\label{conflictingobligations1repeated}
\begin{enumerate}
    \renewcommand{\labelenumi}{\alph{enumi}.}
    \item It is obligatory to leave the building.
    \vspace{5pt}
    \item It is obligatory to not leave the building.
\end{enumerate}
}

\noindent Of course, the bearer of (\ref{conflictingobligations1repeated}.a-b) is ``every human inside the building'', although this is not explicitly written. In other words, the two obligations do not apply to cats, dogs, or other objects inside the building, which are incapable to understand the two norms, nor to the humans {\it outside} the building. In general, norms from legislation always apply to entities that have decisional power, i.e., humans, but also {\it legal} persons such as companies or even robots.

For this reason, in standard legal theory \cite{Searle:95} norms are usually represented as if-then rules that specify, in the antecedent, the conditions that must hold for the bearer of the norms in order to make what is specified in the consequent as obligatory, permitted, etc. for the bearer themselves. 

In the proposed computational ontology, if-then rules are represented as SPARQL rules. Therefore, while ``John is obliged to leave'' is represented with a set of triples as shown above, (\ref{conflictingobligations1repeated}.a) and (\ref{conflictingobligations1repeated}.b), which apply to every human inside the building and not only to John, are represented with the SPARQL rules in (\ref{SPARQLeveryoneisobligedtoleavethebuilding}) and (\ref{SPARQLeveryoneisobligedtonotleavethebuilding}) respectively.

\enumsentence{\label{SPARQLeveryoneisobligedtoleavethebuilding}
\begin{minipage}[t]{550pt}\tt
[a :InferenceRule; :has-sparql-code """\\
\mbox{}\hspace{14pt}CONSTRUCT\{[a soa:Leave, :Obligatory; soa:has-agent ?u;\\
\mbox{}\hspace{80pt}soa:has-from-location soa:theBuilding]\}\\
\mbox{}\hspace{14pt}WHERE\{?e a soa:Be; soa:has-agent ?u;\\
\mbox{}\hspace{51pt}soa:has-inside-location soa:theBuilding.\hspace{-5pt} ?u a soa:Human\}"""]\hspace{-1pt}.
\end{minipage}
}

\enumsentence{\label{SPARQLeveryoneisobligedtonotleavethebuilding}
\begin{minipage}[t]{550pt}\tt[a :InferenceRule; :has-sparql-code """\\
\mbox{}\hspace{14pt}CONSTRUCT\{[a :Obligatory; :not [a soa:Leave; soa:has-agent ?u;\\
\mbox{}\hspace{80pt}soa:has-from-location soa:theBuilding]]\}\\
\mbox{}\hspace{14pt}WHERE\{?e a soa:Be; soa:has-agent ?u;\\
\mbox{}\hspace{51pt}soa:has-inside-location soa:theBuilding.\hspace{-5pt} ?u a soa:Human\}"""]\hspace{-1pt}.
\end{minipage}
}
\vspace{5pt}

\noindent Now, by encoding that John is a human inside the building, i.e.:

\enumsentence{\label{JohnHumanInsideTheBuilding}\tt 
\mbox{}\hspace{5pt}soa:John a soa:Human. soa:ebj a soa:Be; soa:has-agent soa:John;\\ 
\mbox{}\hspace{5pt}soa:has-inside-location soa:theBuilding.
}

\noindent The two SPARQL rules in (\ref{SPARQLeveryoneisobligedtoleavethebuilding}) and (\ref{SPARQLeveryoneisobligedtonotleavethebuilding}) infer that John is obliged to both leave and not leave the building, and a conflict is subsequently inferred as in the examples illustrated above. The same inferences are of course obtained for all individuals that satisfy the {\tt WHERE} clause of the two rules.

The last example shows that, in the proposed computational ontology, conflicts {\it cannot} be inferred between general statements such as (\ref{conflictingobligations1repeated}.a-b), because these correspond to SPARQL (if-then) rules. The general obligations must be instantiated on specific individuals such as John; then, the conflicts can be inferred. Therefore, in order to discover conflicts between general obligations, LegalTech applications should run {\it simulations} by generating synthetic datasets including individuals that satisfy the {\tt WHERE} clauses of the SPARQL rules, in all possible combinations. Of course, this is not ideal; it would be highly preferable to explicitly represent a {\it single} conflict between the two general obligations and without creating any synthetic dataset or simulation. This is not possible in the current setting; however, as it will be explained in section \ref{FutureWorks} below, by incorporating a solid formalization of sets and natural language quantification, such as the one that we envision in our future works, it will be indeed possible to do so.

Another more complex example is shown in (\ref{everyoneParkingSkettyIsObligedToPay3pounds}); this SPARQL rule formalizes the sentence in 
(\ref{conflictingobligations2}.a) above, repeated in (\ref{everyoneParkingSkettyIsObligedToPay3pounds}) for reader's convenience.

\enumsentence{\label{everyoneParkingSkettyIsObligedToPay3pounds}
\hspace{2pt}Whoever parks in a parking spot is obliged to pay £3 at the parking meter associated with that spot.\\[5pt]
\hspace{-2pt}\begin{minipage}[t]{550pt}\tt[a :InferenceRule; :has-sparql-code """\\
\mbox{}\hspace{12pt}CONSTRUCT\{[a soa:Pay,:Obligatory; soa:has-agent ?u;\\
\mbox{}\hspace{78pt}soa:has-object soa:3pounds; soa:has-recipient ?pm]\}\\
\mbox{}\hspace{12pt}WHERE\{?e a soa:Park; soa:has-agent ?u; soa:has-location ?p.\\
\mbox{}\hspace{22pt}?u a soa:Human.\hspace{-3pt} ?p a soa:parkingSpot.\hspace{-3pt} ?pm soa:associated-with ?p\\
\mbox{}\hspace{22pt}NOT EXISTS\{?r a soa:Pay,:Obligatory; soa:has-agent ?u;\\
\mbox{}\hspace{69pt}soa:has-object soa:3pounds; soa:has-recipient ?pm\}\}"""]\hspace{-2pt}.
\end{minipage}
}
\vspace{4pt}

\noindent Similarly to the previous example, if John or anyone else parks in a parking spot, (\ref{everyoneParkingSkettyIsObligedToPay3pounds}) will infer that this individual, i.e., the agent of the parking action, is obliged to pay £3 at the parking meter associated with that spot. Therefore, if the state of affairs also contains a (general) norm such as ``It is prohibited to pay at the parking meters associated with parking spots'', represented via the following SPARQL rule:

\enumsentence{\label{prohibitedParkingAtParkingMeters}
\hspace{-2pt}\begin{minipage}[t]{550pt}\tt[a :InferenceRule; :has-sparql-code """\\
\mbox{}\hspace{12pt}CONSTRUCT\{[a :false,:hold; rdf:subject [a soa:Pay;\\
\mbox{}\hspace{78pt}soa:has-agent ?u; soa:has-recipient ?pm];\\
\mbox{}\hspace{78pt}rdf:predicate rdf:type; rdf:object :Permitted]\}\\
\mbox{}\hspace{12pt}WHERE\{?u a soa:Human.\hspace{-2pt} ?p a soa:parkingSpot.\hspace{-2pt} ?pm soa:associated-with ?p.\\
\mbox{}\hspace{22pt}NOT\hspace{-1pt} EXISTS\{?r a :false,:hold;\hspace{-2pt} rdf:subject ?rp;\hspace{-2pt} rdf:predicate rdf:type;\\
\mbox{}\hspace{88pt}rdf:object :Permitted. ?rp a soa:Pay;\\
\mbox{}\hspace{88pt}soa:has-agent ?u; soa:has-recipient ?pm\}\}"""]\hspace{-2pt}.
\end{minipage}
}
\vspace{2pt}

\noindent A conflict is inferred between the fact that John is not permitted to pay at the parking meter associated with the parking spot where he parked and the fact that he is obliged, thereby permitted, to pay £3 at that parking meter.

The last example of this subsection is the (general) partial conflict shown above in (\ref{conflictingobligationsPartial}) and repeated again in (\ref{conflictingobligationsPartialRepeated}) for reader's convenience.

\enumsentence{\label{conflictingobligationsPartialRepeated}
\begin{enumerate}
    \renewcommand{\labelenumi}{\alph{enumi}.}
    \item It is obligatory to pay in cash.
    \vspace{5pt}
    \item It is obligatory to pay by card.
\end{enumerate}
}

\noindent In (\ref{conflictingobligationsPartialOnJohn}), we already investigated the simplified version of (\ref{conflictingobligationsPartialRepeated}) applied to John; now, we will generalize that example fit to make the very same inferences for {\it any} human, and not only for John. The generalization is achieved by replacing the RDF triples that hold for John with the two SPARQL rules in (\ref{itisobligatorytopayincash}), which trigger for any human.

\enumsentence{\label{itisobligatorytopayincash}
\begin{minipage}[t]{550pt}\tt[a :InferenceRule; :has-sparql-code """\\
\mbox{}\hspace{12pt}CONSTRUCT\{[a soa:Pay,:Obligatory; soa:has-agent ?u;\\
\mbox{}\hspace{78pt}soa:has-instrument soa:cash]\}\\
\mbox{}\hspace{13pt}WHERE\{?u a soa:Human. NOT\hspace{-1pt} EXISTS\{?r a soa:Pay,:Obligatory;\\
\mbox{}\hspace{50pt}soa:has-agent ?u; soa:has-instrument soa:cash\}\}"""]\hspace{-2pt}.
\end{minipage}\\[8pt]
\begin{minipage}[t]{550pt}\tt[a :InferenceRule; :has-sparql-code """\\
\mbox{}\hspace{12pt}CONSTRUCT\{[a soa:Pay,:Obligatory; soa:has-agent ?u;\\
\mbox{}\hspace{78pt}soa:has-instrument soa:card]\}\\
\mbox{}\hspace{13pt}WHERE\{?u a soa:Human. NOT\hspace{-1pt} EXISTS\{?r a soa:Pay,:Obligatory;\\
\mbox{}\hspace{50pt}soa:has-agent ?u; soa:has-instrument soa:card\}\}"""]\hspace{-2pt}.
\end{minipage}
}

\noindent The GitHub repository includes examples about all inferences discussed in this subsection.

\vspace{-5pt}
\section{Conjunction, disjunction, and (material) implication of deontic statements}\label{ConjunctionDisjunctionImplicationOfDeonticStatements}
\vspace{-2pt}

\noindent This section extends the content of the previous one, which contains the core part of the research presented in this paper, by integrating therein SPARQL rules for properly dealing with conjunctions and disjunctions of eventualities.

This section specifically aims at providing an answer to the Smith argument and its variants discussed in \cite{Goble:13} as well as in subsection \ref{conflictTolerantDeonticLogicGoble} above (Desideratum \#3). As explained in that subsection, among the revisionist strategies reviewed in \cite{Goble:13}, only one, the deontic logic {\tt BDL}, as well as a slight variant of it, are capable to achieve all desiderata, including the proper representation of the Smith argument and its variants. The axioms of {\tt BDL} constrain the conjunction, disjunction, and implication of the deontic operator {\tt OB}, which denotes obligatoriness.

Concerning conjunction, the two axioms in (\ref{axiomatizationBDLconjunction}) hold in {\tt BDL}: if a conjunction of statements is obligatory, then also each statement is (axiom ({\tt M})) and if two statements are obligatory, then also its conjunction is (axiom ({\tt C})).

\enumsentence{\label{axiomatizationBDLconjunction}
\begin{enumerate}
    \renewcommand{\labelenumi}{}
    
    \item\noindent ({\tt M})\hspace{23pt}
    {\tt OB}(\hspace{1pt}{\tt p}\hspace{1pt}$\wedge$\hspace{1pt}{\tt q}\hspace{1pt}) $\rightarrow$ {\tt OB}(\hspace{1pt}{\tt p})
    
    \vspace{2pt}
    \item\noindent ({\tt C})\hspace{23pt}
    ({\tt OB}(\hspace{1pt}{\tt p}\hspace{1pt}) $\wedge$ {\tt OB}({\tt q}\hspace{1pt})) $\rightarrow$ {\tt OB}(\hspace{1pt}{\tt p}\hspace{1pt}$\wedge$\hspace{1pt}{\tt q}\hspace{1pt})
\end{enumerate}
}

\noindent It has been shown in (\ref{GordonHobbsNotAndOrAxioms}.b) above that a bi-implication corresponding to ({\tt M}) and ({\tt C}) holds in \cite{Gordon-Hobbs:17} for the {\tt Rexist} modality, thus the two rules in (\ref{SPARQLand1}) and (\ref{SPARQLand2}) have been added to the proposed computational ontology. 

In light of the equivalences in natural language exemplified in (\ref{MandCforObPeOp}.a-c), it sounds rather intuitive to also add corresponding SPARQL rules for both the class {\tt Obligatory}, to parallel the two axioms in (\ref{axiomatizationBDLconjunction}), as well as for the classes {\tt Permitted} and {\tt Optional}. In (\ref{MandCforObPeOp}.a-c), ``iff'' means that the implication also holds in the opposite direction (i.e., that it is actually a {\it bi}-implication).

\enumsentence{\label{MandCforObPeOp}
\begin{minipage}[t]{350pt}\vspace{-10pt}\begin{enumerate}
    \renewcommand{\labelenumi}
    {\alph{enumi}.}

    \item\noindent Iff ``John is obliged to eat and drink'' then ``John is obliged to eat'' and ``John is obliged to drink''.
    \vspace{5pt}
    
    \item\noindent Iff ``John is permitted to eat and drink'' then ``John is permitted to eat'' and ``John is permitted to drink''.
    \vspace{5pt}
    
    \item\noindent Iff ``it is optional for John to eat and drink'' then ``it is optional for John to eat'' and ``it is optional for John to drink''.
\end{enumerate}
\end{minipage}
}
\vspace{2pt}

\noindent Therefore, the SPARQL rules in (\ref{SPARQLand1}) and (\ref{SPARQLand2}) above have  been extended to cover the other three deontic modalities as well. However, this paper omits to show their extended version; the reader may find it in the GitHub repository.

On the other hand, it is easy to see that the bi-implications holding for {\tt Rexist} with respect to negation and disjunction do {\it not} also hold for the three deontic modalities. 

With respect to negation, it has been already mentioned above in subsection \ref{GordonHobbsNegationDisjunctionConjunction} that if two eventualities are related through the {\tt not} predicate, the fact that one of them is obligatory does {\it not} entail that the other one is not obligatory. For example, the fact that John is obliged to leave is {\it not} equivalent to the fact that he is not obliged to stay (i.e., to not leave). From the implications of the Deontic Traditional Scheme, this might be now also seen formally: the implication only holds in one direction. For instance, with respect to the {\tt Obligatory} modality, it holds:

\enumsentence{\label{obligatoryAndNot}
\hspace{5pt}({\tt Obligatory}({\tt p}) $\wedge$ {\tt not}({\tt p}, {\tt q})) $\rightarrow$ $\neg${\tt Permitted}({\tt q}) $\rightarrow$ $\neg$\hspace{1pt}{\tt Obligatory}({\tt q})\\[3pt]
\mbox{}\hspace{5pt}($\neg$\hspace{1pt}{\tt Obligatory}({\tt q}\hspace{1pt}) $\wedge$ {\tt not}({\tt p}, {\tt q}\hspace{1pt})) $\rightarrow$ {\tt Permitted}(\hspace{1pt}{\tt p}) $\not\rightarrow$ $\neg$\hspace{1pt}{\tt Obligatory}(\hspace{1pt}{\tt p})
}
\vspace{2pt}

\noindent Similar considerations hold for disjunction; neither {\tt BDL} nor any axiomatization for {\tt OB} proposed in the literature stipulates that 
{\tt OB}({\tt p}$\vee${\tt q})\hspace{1pt}$\leftrightarrow$\hspace{1pt}{\tt OB}({\tt p})\hspace{1pt}$\vee$\hspace{1pt}{\tt OB}({\tt q}); indeed, it is easy to see that the Deontic Traditional Scheme does {\it not} entail this equivalence:

\enumsentence{\label{obligatoryAndVee}
{\tt OB}({\tt p}\hspace{1pt}$\vee$\hspace{1pt}{\tt q}) $\rightarrow$ {\tt OB}($\neg\neg$\hspace{1pt}({\tt p}\hspace{1pt}$\vee$\hspace{1pt}{\tt q})) $\rightarrow$ {\tt OB}($\neg$\hspace{1pt}($\neg$\hspace{1pt}{\tt p}\hspace{1pt}$\wedge$\hspace{1pt}$\neg$\hspace{1pt}{\tt q})) $\rightarrow$ $\neg${\tt PE}($\neg$\hspace{1pt}{\tt p}\hspace{1pt}$\wedge$\hspace{1pt}$\neg$\hspace{1pt}{\tt q})$\rightarrow$\\[3pt]
$\neg$\hspace{1pt}({\tt PE}($\neg$\hspace{1pt}{\tt p})\hspace{1pt}$\wedge$\hspace{1pt}{\tt PE}($\neg$\hspace{1pt}{\tt q})) $\rightarrow$ ($\neg$\hspace{1pt}{\tt PE}($\neg$\hspace{1pt}{\tt p})\hspace{1pt}$\vee$$\neg$\hspace{1pt}{\tt PE}($\neg$\hspace{1pt}{\tt q})) $\rightarrow$ ($\neg$\hspace{1pt}{\tt OB}($\neg$\hspace{1pt}{\tt p})\hspace{1pt}$\vee$$\neg$\hspace{1pt}{\tt OB}($\neg$\hspace{1pt}{\tt q})) $\rightarrow$\\[3pt]
({\tt PE}({\tt p})\hspace{1pt}$\vee$\hspace{1pt}{\tt PE}({\tt q}))
$\not\rightarrow$ ({\tt OB}({\tt p})\hspace{1pt}$\vee$\hspace{1pt}{\tt OB}({\tt q}))
}
\vspace{2pt}

\noindent Since the Deontic Traditional Scheme does not entail the bi-implication {\tt OB}({\tt p}$\vee${\tt p})\hspace{1pt}$\leftrightarrow$\hspace{1pt}{\tt OB}({\tt p})\hspace{1pt}$\vee$\hspace{1pt}{\tt OB}({\tt p})\hspace{1pt}, in order to capture the intuitions of the Smith argument the axiom {\tt DDS} must be added to {\tt BDL}. {\tt DDS} implements Disjunctive Syllogism for the propositional formulae occurring {\it within} the operator {\tt OB}:

\enumsentence{\label{axiomatizationBDLdisjunction}
\begin{enumerate}
    \renewcommand{\labelenumi}{}
    \item\noindent ({\tt DDS})\hspace{11pt}
    ({\tt OB}(\hspace{1pt}{\tt p}\hspace{1pt}$\vee$\hspace{1pt}{\tt q}\hspace{1pt}) $\wedge$ {\tt OB}(\hspace{1pt}$\neg$\hspace{1pt}{\tt q}\hspace{1pt})) $\rightarrow$ {\tt OB}(\hspace{1pt}{\tt p})
\end{enumerate}
}

\noindent No SPARQL rule introduced so far parallels the axiom in (\ref{axiomatizationBDLdisjunction}). The rule shown above in (\ref{DisjunctiveSyllogismAtStatementLevel}) implements Disjunctive Syllogism {\it at the level of the statements}; this rule parallels 
(({\tt A}$\vee${\tt B})\hspace{1pt}$\wedge$\hspace{0pt}$\neg${\tt B})\hspace{1pt}$\rightarrow$\hspace{1pt}{\tt A} for any {\tt A} or {\tt B}, and so it enables the implication 
(({\tt OB}({\tt p})$\vee${\tt OB}({\tt q}))\hspace{1pt}$\wedge$\hspace{0pt}$\neg${\tt OB}({\tt q}))\hspace{1pt}$\rightarrow$\hspace{1pt}{\tt OB}({\tt p}) but not the one in  (\ref{axiomatizationBDLdisjunction}). 

Therefore, a SPARQL rule implementing Disjunctive Syllogism at the level of the eventualities for the class {\tt Obligatory}, which parallels the axiom {\tt DDS}, must be also added to the proposed computational ontology:

\enumsentence{\label{DisjunctiveSyllogismOnObligatory}
\begin{minipage}[t]{450pt}\tt
\mbox{}\hspace{0pt}[a :InferenceRule; :has-sparql-code\hspace{-3pt} """\\
\mbox{}\hspace{16pt}CONSTRUCT\{?e2 a :Obligatory\}\\
\mbox{}\hspace{16pt}WHERE\{?eo a :Obligatory. ?en1 a :Obligatory.\\
\mbox{}\hspace{51pt}\{?eo :or1 ?e1; :or2 ?e2\}UNION\{?eo :or1 ?e2; :or2 ?e1\}\\
\mbox{}\hspace{51pt}\{?e1 :not ?en1\}UNION\{?en1 :not ?e1\}\}"""]\hspace{-1pt}.
\end{minipage}
}
\vspace{3pt}

\noindent It is now easy to see that the rule in (\ref{DisjunctiveSyllogismOnObligatory}) enables the inferences of the Smith argument. Let's consider the simplified version of the Smith argument in (\ref{SmithArgumentSimplified}), which allows for more compact RDF representations than the original version shown above in (\ref{SmithArgument}).

\enumsentence{\label{SmithArgumentSimplified}
\noindent{\it From}:\hspace{2pt} Smith ought to eat or drink.\\[4pt]
\noindent\mbox{}\hspace{2pt}{\it And}:\hspace{2pt} Smith ought not to eat.\\[4pt]
\noindent\mbox{}\hspace{2pt}{\it It is intuitive to conclude that:}\hspace{2pt} Smith ought to drink.
}
\vspace{2pt}

\noindent The RDF representation of the {\it From:} and the {\it And:} clauses in (\ref{SmithArgumentSimplified}) is the following:

\enumsentence{\label{SmithOughtToEatOrDrinkAndOughtToNotEat}\tt 
\mbox{}\hspace{2pt}soa:eso a :Obligatory; :or1 soa:ese; :or2 soa:esd.\\
\mbox{}\hspace{2.5pt}soa:ese a soa:Eat; soa:has-agent soa:Smith.\\
\mbox{}\hspace{2.5pt}soa:esd a soa:Drink; soa:has-agent soa:Smith.\\ 
\mbox{}\hspace{2.5pt}soa:ese :not soa:ense. soa:ense a :Obligatory.
}
\vspace{2pt}

\noindent From (\ref{SmithOughtToEatOrDrinkAndOughtToNotEat}), the rule in (\ref{DisjunctiveSyllogismOnObligatory}) infers the triple ``{\tt soa:esd a :Obligatory}'', i.e., that Smith is obliged to drink.

Note, on the other hand, that Disjunctive Syllogism at the level of the eventualities does not hold for the class {\tt Permitted}. However, for the class {\tt Permitted} the following bi-implications hold:

\enumsentence{\label{permissionAndVee}
{\tt PE}({\tt p}\hspace{1pt}$\vee$\hspace{1pt}{\tt q}) $\leftrightarrow$ {\tt PE}($\neg\neg$\hspace{1pt}({\tt p}\hspace{1pt}$\vee$\hspace{1pt}{\tt q})) $\leftrightarrow$ {\tt PE}($\neg$\hspace{1pt}($\neg$\hspace{1pt}{\tt p}\hspace{1pt}$\wedge$\hspace{1pt}$\neg$\hspace{1pt}{\tt q})) $\leftrightarrow$ $\neg${\tt OB}($\neg$\hspace{1pt}{\tt p}\hspace{1pt}$\wedge$\hspace{1pt}$\neg$\hspace{1pt}{\tt q})$\leftrightarrow$\\[3pt]
$\neg$\hspace{1pt}({\tt OB}($\neg$\hspace{1pt}{\tt p})\hspace{1pt}$\wedge$\hspace{1pt}{\tt OB}($\neg$\hspace{1pt}{\tt q})) $\leftrightarrow$
\mbox{}\hspace{2pt}($\neg$\hspace{1pt}{\tt OB}($\neg$\hspace{1pt}{\tt p})\hspace{1pt}$\vee$$\neg$\hspace{1pt}{\tt OB}($\neg$\hspace{1pt}{\tt q})) $\leftrightarrow$ ({\tt PE}({\tt p})\hspace{1pt}$\vee$\hspace{1pt}{\tt PE}({\tt q}))
}
\vspace{2pt}

\noindent Therefore, the SPARQL rule implementing the following bi-implication is also added to the proposed computational ontology. (\ref{GordonHobbsNotAndOrAxiomsPermitted}) parallels the bi-implication for the {\tt Rexist} modality shown above in (\ref{GordonHobbsNotAndOrAxioms}.c).

\enumsentence{\label{GordonHobbsNotAndOrAxiomsPermitted}
\mbox{}$\forall_{{\tt eo},\hspace{2pt}{\tt e1},\hspace{2pt} {\tt e2}}$[ {\tt or}\hspace{1pt}({\tt eo}, {\tt e1}, {\tt e2}) $\rightarrow$\\[3pt]
\mbox{}\hspace{46pt}({\tt Permitted}({\tt eo}) $\leftrightarrow$ ({\tt Permitted}({\tt e1}) $\vee$ {\tt Permitted}({\tt e2}))) ]
}
\vspace{2pt}

\noindent It is easy to see that the same considerations hold for optionality, therefore a SPARQL rule implementing the counterpart of (\ref{GordonHobbsNotAndOrAxiomsPermitted}) for the class {\tt Optional}, of which this paper omits further details, is added to the ontology.

Given (\ref{permissionAndVee}), it is easy to see that the Deontic Traditional Scheme does not entail Disjunctive Syllogism at the level of the eventualities for the class {\tt Permitted}:

\enumsentence{\label{noDSwithPermission}
\mbox{}({\tt PE}(\hspace{1pt}{\tt p}\hspace{1pt}$\vee$\hspace{1pt}{\tt q}\hspace{1pt}) $\wedge$ {\tt PE}(\hspace{1pt}$\neg$\hspace{1pt}{\tt q}\hspace{1pt})) $\leftrightarrow$ (({\tt PE}({\tt p})\hspace{1pt}$\vee$\hspace{1pt}{\tt PE}({\tt q})) $\wedge$ {\tt PE}(\hspace{1pt}$\neg$\hspace{1pt}{\tt q}\hspace{1pt})) $\leftrightarrow$\\[3pt]
\noindent\mbox{}\hspace{3pt}(({\tt PE}(\hspace{1pt}{\tt p}\hspace{1pt}) $\wedge$ {\tt PE}(\hspace{1pt}{\tt q}\hspace{1pt})) $\vee$ {\tt OP}(\hspace{1pt}{\tt q}\hspace{1pt})) $\not\rightarrow$ {\tt PE}(\hspace{1pt}{\tt p}\hspace{1pt})
}
\vspace{2pt}

\noindent In addition, it seems that neither our intuitions about the interplay between permissions and disjunction suggest that we should add a SPARQL rule that parallels (\ref{DisjunctiveSyllogismOnObligatory}) also for the class {\tt Permitted}. Consider:

\enumsentence{\label{SmithArgumentSimplifiedPermitted}
\noindent{\it From}:\hspace{2pt} Smith is permitted to eat or drink.\\[4pt]
\noindent\mbox{}\hspace{2pt}{\it And}:\hspace{2pt} Smith is permitted not to eat.\\[4pt]
\noindent\mbox{}\hspace{2pt}{\it It is} \underline{\textbf{NOT}} {\it intuitive to conclude that:}\hspace{2pt} Smith is permitted to drink.
}
\vspace{2pt}

\noindent In (\ref{SmithArgumentSimplifiedPermitted}), the sentence in the ``{\it From:}'' clause states that Smith is permitted to either eat or drink (or both), but we do not know which ones of the two. However, the sentence in the ``{\it And:}'' clause, which states that Smith is also permitted to not eat, i.e., to starve, does not inform us which one of the two disjunctive permissions in the ``{\it From:}'' clause holds or does not hold. Therefore, we cannot conclude that Smith is permitted to drink.

The Smith argument and all its variants discussed in \cite{Goble:13}, called the ``Jones'', the ``Roberts'', and the ``Thomas'' arguments, have been formalized, commented, and made available in the GitHub repository. The SPARQL rules shown above in this section enable the proper inferences in all these arguments.

Before concluding the section, let us also spend a couple of words about implication. So far, this section discussed the counterparts in the proposed computational ontology of all  axioms defined in {\tt BDL} except one: the axiom ({\tt RBE}), which is a reduced version of the axiom ({\tt NM}) from Standard Deontic Logic fit to avoid deontic explosion in {\tt BDL}. Both ({\tt RBE}) and ({\tt NM}) are repeated again in (\ref{RBEandNM}) for reader's convenience:

\enumsentence{\label{RBEandNM}
\begin{enumerate}
    \renewcommand{\labelenumi}{}

    \item ({\tt RBE})\hspace{9pt}
    (\hspace{1pt}{\tt p}\hspace{2pt}$\leftrightarrow_{\tt\hspace{-0.5pt} A}$\hspace{2pt}{\tt q}\hspace{1pt}) $\rightarrow$ ({\tt OB}(\hspace{1pt}{\tt p}\hspace{1pt})\hspace{2pt}$\leftrightarrow$\hspace{2pt}{\tt OB}(\hspace{1pt}{\tt q}\hspace{1pt}))
    
    \vspace{3pt}
    \item\mbox{}\hspace{3pt}({\tt NM})\hspace{17pt}$\Box$\hspace{1pt}(\hspace{1pt}{\tt p}\hspace{1pt}$\rightarrow$\hspace{1pt}{\tt q}\hspace{1pt}) $\rightarrow$ ({\tt OB}(\hspace{1pt}{\tt p}) $\rightarrow$ {\tt OB}({\tt q}\hspace{1pt}))
\end{enumerate}
}

\noindent The proposed computational ontology does not currently include SPARQL rules that implement the axiom ({\tt NM}) nor, more generally, entailments among obligatory (or permitted, or optional) eventualities. Note, on the other hand, that, contrary to {\tt BDL}, here there is no need to implement a restricted version of ({\tt NM}) since deontic explosion is already avoided by avoiding Disjunctive Introduction, as explained above in subsubsection \ref{ParaconsistentDeonticLogic}.

Still, all examples shown in this paper, which are superior in number, expressivity, and variety to the ones discussed in \cite{Goble:13}, do not need SPARQL rules implement entailments among obligatory/permitted/optional eventualities. 

Indeed, a closer look reveals that ({\tt NM}) and its variants were primarily asserted in the deontic logics discussed in \cite{Goble:13} for a {\it technical} reason, and not because of some intuitions about the notion of obligatoriness: ({\tt NM}) is needed in these deontic logics in order to enable classical logic inferences {\it within} the second-order operator {\tt OB}. For instance, the assertion {\tt OB}(\hspace{-2pt}{\tt(A$\rightarrow$B)$\wedge$\hspace{1pt}{\tt A}}\hspace{0.8pt}) does {\it not} entail {\tt OB}({\tt B}\hspace{1pt}) without the axiom ({\tt NM}): nothing in the definition of {\tt OB} stipulates that Modus Ponens applies to the formulae in its scope. On the other hand, by taking {\tt p} as (({\tt A}$\rightarrow${\tt B})$\wedge$\hspace{1pt}{\tt A}) and {\tt q} as {\tt B}, since Modus Ponens is a tautology in classical logic, the axiom ({\tt NM}) stipulates that: 

\begin{center}
    $\Box$\hspace{1pt}(\hspace{1pt}(({\tt A}$\rightarrow${\tt B})\hspace{1pt}$\wedge$\hspace{1pt}{\tt A}) $\rightarrow$ {\tt B})\hspace{1pt} $\rightarrow$ ({\tt OB}(\hspace{-2pt}{\tt(A$\rightarrow$B)$\wedge$\hspace{1pt}{\tt A}}\hspace{1pt}) $\rightarrow$ {\tt OB}({\tt B}\hspace{1pt})),
    i.e., that {\tt OB}(\hspace{-2pt}{\tt(A$\rightarrow$B)$\wedge$\hspace{1pt}{\tt A}}\hspace{0.8pt}) entails {\tt OB}(\hspace{1pt}{\tt B}\hspace{1pt})
\end{center}

\noindent Therefore, the reason why the proposed computational ontology does not need SPARQL rules that parallel ({\tt NM}), at least not to deal with the examples considered in this paper, is that the ontology's formalization is not grounded on second-order operators such as {\tt OB}, in which scope the entailments from classical logic much be enabled for the logical framework to achieve the desired inferences.

Conversely, the proposed computational ontology is grounded on the distinction between the level of the eventualities and the level of the statements, which in a sense respectively correspond to the ``inside'' and the ``outside'' of the {\tt OB} second-order operator. Both levels are implemented in RDF. Then, special SPARQL rules connect and constrain the assertions in the two levels depending on negation, conjunction, disjunction, as well as the modalities.

On the other hand, adding further SPARQL rules that implement entailments among deontic statements is not so simple as it appears at first glance. In some cases, an axiom such as ({\tt NM}) seems to be reasonable and intuitive; for example, if it is true that to reach the supermarket it is necessary to cross the bridge (which, in propositional symbol, might be formalized as $\Box$({\tt S}$\rightarrow${\tt B})), then we may conclude that ``John is obliged to go to the supermarked'' entails ``John is obliged to cross the brige'', i.e., {\tt OB}({\tt S})$\rightarrow${\tt OB}({\tt B}). Nevertheless, it is likewise easy to find cases in which the same conclusions do not sound so reasonable and intuitive. For example, if ``John drives the car'' entails ``The car is moving'', i.e., the latter is a necessary condition to state that the former holds true, it does not seem to be so intuitive to conclude that ``John is obliged to drive'' entails ``The car is obliged to move'', {\it unless John's car is an autonomous vehicle, i.e., a robot, that is obliged to execute all what John is}.

These considerations are not new in the literature in formal logic. It is widely assumed that {\it material} implication, i.e., when {\tt A}$\rightarrow${\tt B} is taken as equivalent to 
$\neg$\hspace{1pt}{\tt A}$\vee$\hspace{1pt}{\tt B}\hspace{1pt}, is unable to represent the proper meaning of several statements and it leads to some paradoxes. The example of John driving is car is therefore just another evidence of the overall material implication's inadequacy for representing meaning.

Suitable alternatives to material implication are William Parry's logic of Analytic Implication \cite{Parry:89} as well as Kit Fine's truthmaker semantics, counterfactual conditionals and non-classical consequence relations \cite{Faroldi-vanDePutte:23}. In these accounts, an implication {\tt A}$\rightarrow${\tt B} is valid only if the consequent {\tt B} is somewhat related to the antecedent {\tt A}. The relation between consequent and antecedent is enforced by stipulating some formal constraints; for instance, Parry’s Analytic Implication requires all propositional variables occurring in {\tt B} to also occur in {\tt A}. What the proposed computational ontology and these account have in common is that, while they accept Disjunctive Syllogism, they avoid Disjunctive Introduction in its general form (cf. discussion in subsubsection \ref{ParaconsistentDeonticLogic} above). For instance, in Parry’s Analytic Implication, 
{\tt A}$\rightarrow$\hspace{0.5pt}({\tt A}$\vee${\tt B}) is valid only if all propositional variables occurring in {\tt B} also occur in {\tt A}.

In light of this, in the future we will consider incorporating insights from \cite{Parry:89} and \cite{Faroldi-vanDePutte:23} in the proposed computational ontology, namely to implement them into new RDF resources and SPARQL rules. Nevertheless, it is clear that lot of further research is needed to this end; for example, how can we formalize  the relation between John and his autonomous car in the example considered above?

\section{Deontic modalities and contextual constraints}\label{DeonticModalitiesAndContextualConstraints}

\noindent This section also extends the content of section 
\ref{DeonticModalitiesInRDFsAndSPARQL}, which is the core section of this paper, by integrating therein additional SPARQL rules in light of a specific comparison with the work in \cite{Goble:13}. 

As explained above in (\ref{intuitions}.b), \cite{Goble:13} defines conflicts among obligations as 
situations in which ``an agent ought to do a number of things, each of which is possible for the agent, but it is impossible for the agent to do them all'', formalized as ``{\tt OB}({\tt A}) $\wedge$ {\tt OB}({\tt B}) $\wedge\hspace{2pt}\neg\hspace{0.5pt}\Diamond$({\tt A}$\wedge${\tt B})''. This section argues that this is not the right definition of conflicts but it is rather a specific case of the general interplay between obligations and constraints holding in the context.

Consider again the sentences shown above in (\ref{conflictingobligations2}.b) and (\ref{conflictingobligations3}), repeated again in (\ref{conflictingobligations2repeated}.a-b) for reader's convenience.

\enumsentence{\label{conflictingobligations2repeated}
\begin{enumerate}
    \renewcommand{\labelenumi}{\alph{enumi}.}
    \item The parking meter in Sketty only accepts cash.
    \vspace{5pt}
    \item It is prohibited to pay by cash.
\end{enumerate}
}

\noindent (\ref{conflictingobligations2repeated}.a) denotes a {\it constraint} holding in the state of affairs: in Sketty it is {\it not possible} to not pay in cash, i.e., it is {\it necessary} to pay in cash. In standard propositional modal logic, this might be represented as 
$\neg\hspace{0.5pt}\Diamond$($\neg$\hspace{0.5pt}{\tt C}). (\ref{conflictingobligations2repeated}.b) states that it is prohibited to pay in cash, i.e., that it is obligatory to not pay in cash. In standard propositional deontic logic, this might be represented as 
{\tt OB}($\neg$\hspace{0.5pt}{\tt C}). Therefore, whenever 
{\tt OB}({\tt A}\hspace{1pt}$\wedge$\hspace{1pt}{\tt B})\hspace{1pt}$\leftrightarrow$\hspace{0.5pt}{\tt OB}({\tt A})\hspace{1pt}$\wedge$\hspace{0.5pt}{\tt OB}({\tt B}) holds, e.g., in {\tt BDL} but also in the proposed computational ontology, the definition used in \cite{Goble:13} is just a particular case of what has been exemplified above in (\ref{conflictingobligations2repeated}): when {\tt A}\hspace{1pt}$\wedge$\hspace{1pt}{\tt B}\hspace{1.5pt}$\leftrightarrow$$\neg$\hspace{0.5pt}{\tt C}.

Having said that, this section will show how the proposed computational ontology incorporates the notions of possibility and necessity, i.e., how it models the interplay between deontic modalities and contextual constraints.

First of all, we observe that contextual constraints appear to only apply to the {\it thematic roles} of an eventuality. In other words, they restrict {\it the way} in which some actions or states might take place. For instance, (\ref{conflictingobligations2repeated}.a) constrains the {\it instrument} of the payments in Sketty: this can only be cash.

Two new classes are then introduced at the level of the statements: {\tt necessary} and {\tt possible}:

\enumsentence{\label{necessarypossibleRDFsClasses}\tt
\mbox{}\hspace{2pt}:necessary a rdfs:Class; rdfs:subClassOf :statement.\\
\mbox{}\hspace{5pt}:possible a rdfs:Class; rdfs:subClassOf :statement.
}

\noindent By using these classes, the fact that it is {\it necessary} for the instrument of the paying action {\tt ep} to be cash is encoded as in (\ref{necessarypossibleExample}.a) while the fact that this is {\it possible} is encoded as in (\ref{necessarypossibleExample}.b):

\enumsentence{\label{necessarypossibleExample}
\begin{enumerate}
    \renewcommand{\labelenumi}{\alph{enumi}.}
    \item \tt[a :necessary,:hold; rdf:subject soa:ep;\\ 
    \mbox{}\hspace{6pt}rdf:predicate soa:has-instrument; rdf:object soa:cash]
    \vspace{4pt}
    \item \tt[a :possible,:hold; rdf:subject soa:ep;\\ 
    \mbox{}\hspace{6pt}rdf:predicate soa:has-instrument; rdf:object soa:cash]
\end{enumerate}
}

\noindent In Hobbs's, the triples in (\ref{necessarypossibleExample}.a) might be represented as $\Box${\tt soa:has-instrument}({\tt ep}, {\tt cash}) while those in (\ref{necessarypossibleExample}.b) as $\Diamond${\tt soa:has-instrument}({\tt ep}, {\tt cash}). Since RDF does not define operators corresponding to $\Box$ and $\Diamond$, similarly to what has been done with the standard negation $\neg$, the classes in (\ref{necessarypossibleRDFsClasses}) have been introduced.

As it is well-known, the equivalence $\Box${\tt A}$\leftrightarrow$$\neg\Diamond\neg${\tt A} holds, meaning that if something is necessary then its contrary is not possible and vice versa. However, the SPARQL rules implementing this equivalence are omitted because they are not needed for the examples shown in this section; similarly to the distributivity laws on conjunction and disjunction or De Morgan's laws, their implementation is left to the reader as an exercize.

(\ref{conflictingobligations2repeated}.a) can be now represented via the SPARQL rule in (\ref{inskettyitisnecessarytopaycash}), which states that for every payment at the parking meter associated with the parking spot in Sketty it is necessary for the instrument of this payment to be cash.

\enumsentence{\label{inskettyitisnecessarytopaycash}
\hspace{-2pt}\begin{minipage}[t]{550pt}\tt[a :InferenceRule; :has-sparql-code """\\
\mbox{}\hspace{12pt}CONSTRUCT\{[a :necessary, :hold; rdf:subject ?ep;\\
\mbox{}\hspace{68pt}rdf:predicate soa:has-instrument; rdf:object soa:cash]\}\\
\mbox{}\hspace{12pt}WHERE\{?ep a soa:Pay; soa:has-recipient ?pm.\\
\mbox{}\hspace{20pt}?pm soa:associated-with soa:psSketty.\\
\mbox{}\hspace{22pt}NOT EXISTS\{?r a :necessary, :hold; rdf:subject ?ep;\\
\mbox{}\hspace{40pt}rdf:predicate soa:has-instrument; rdf:object soa:cash\}\}"""]\hspace{-2pt}.
\end{minipage}
}
\vspace{2pt}

\noindent Now, by encoding that John paid at the parking meter associated with the parking spot in Sketty, i.e., by asserting the RDF triples in (\ref{JohnPaidInSketty}), it should be possible to infer that John paid {\it in cash}. However, this inference is enabled only by adding to the proposed computational ontology the SPARQL rule in 
(\ref{IfNecessaryItIsTrue}), which asserts as true every statement that is necessary; (\ref{IfNecessaryItIsTrue}) corresponds to the well-known modal logic rule $\Box$\hspace{0.5pt}{\tt A}$\rightarrow${\tt A}, which imposes reflexivity on the accessibility relation in standard Kripke possible-world semantics.

\enumsentence{\label{JohnPaidInSketty}\tt
\mbox{}soa:John a soa:Human. soa:epsj\hspace{-1pt} a\hspace{-1pt} soa:Pay,:Rexist;\\
\mbox{}\hspace{2pt}soa:has-agent\hspace{-1pt} soa:John; soa:has-recipient\hspace{-1pt}\\
\mbox{}\hspace{2pt}soa:pmSketty\hspace{-1pt}. soa:psSketty a soa:parkingSpot.\\
\mbox{}\hspace{2pt}soa:pmSketty soa:associated-with soa:psSketty.
}

\enumsentence{\label{IfNecessaryItIsTrue}
\hspace{-2pt}\begin{minipage}[t]{550pt}\tt[a :InferenceRule; :has-sparql-code """\\
\mbox{}\hspace{14pt}CONSTRUCT\{?s ?p ?o\}\\
\mbox{}\hspace{14pt}WHERE\{?r a :necessary, :hold; rdf:subject ?s;\\
\mbox{}\hspace{50pt}rdf:predicate ?p; rdf:object ?o\}"""]\hspace{-2pt}.
\end{minipage}
}
\vspace{2pt}

\noindent Let us now complete the example in (\ref{conflictingobligations2repeated}); the prohibition in (\ref{conflictingobligations2repeated}.b) is represented as follows:

\enumsentence{\label{itisprohibitedtopayincash}
\hspace{-2pt}\begin{minipage}[t]{550pt}\tt[a :InferenceRule; :has-sparql-code """\\
\mbox{}\hspace{12pt}CONSTRUCT\{[a :false,:hold; rdf:subject [a soa:Pay;\\
\mbox{}\hspace{78pt}soa:has-agent ?u; soa:has-instrument soa:cash];\\
\mbox{}\hspace{78pt}rdf:predicate rdf:type; rdf:object :Permitted]\}\\
\mbox{}\hspace{12pt}WHERE\{?u a soa:Human. \hspace{-4pt}NOT\hspace{-1pt} EXISTS\{?r a :false,:hold;\hspace{-2pt} rdf:subject\hspace{-2pt} ?rp;\\
\mbox{}\hspace{34pt}\hspace{-2pt} rdf:predicate rdf:type; rdf:object :Permitted.\hspace{-4pt} ?rp a soa:Pay;\\
\mbox{}\hspace{38pt}soa:has-agent ?u; soa:has-instrument soa:cash\}\}"""]\hspace{-2pt}.
\end{minipage}
}
\vspace{2pt}

\noindent From the triples in (\ref{JohnPaidInSketty}), the SPARQL rules in (\ref{inskettyitisnecessarytopaycash}), (\ref{IfNecessaryItIsTrue}), 
(\ref{itisprohibitedtopayincash}), and, finally, the rule shown above in (\ref{SPARQLisviolatedbyNotPermittedRexist}), infer the triples in (\ref{JohnPaidInCashAndHeIsProhibitedToPayInCash}.a-c): the rule in (\ref{inskettyitisnecessarytopaycash}) infers that it is necessary for the instrument of {\tt soa:epsj} to be cash, i.e., (\ref{JohnPaidInCashAndHeIsProhibitedToPayInCash}.a); from (\ref{JohnPaidInCashAndHeIsProhibitedToPayInCash}.a), the rule in (\ref{IfNecessaryItIsTrue}) infers that the instrument of {\tt soa:epsj} is cash, i.e., (\ref{JohnPaidInCashAndHeIsProhibitedToPayInCash}.b). In parallel, 
from the triples in (\ref{JohnPaidInSketty})
the rule in 
(\ref{itisprohibitedtopayincash}) infers that John is prohibited to pay in cash. Finally, since in (\ref{JohnPaidInSketty}) it holds that the payment {\tt soa:epsj} really exists and its instrument is cash,
the rule in (\ref{SPARQLisviolatedbyNotPermittedRexist}) infers that the fact that {\tt soa:epsj} really exists violates John's prohibition, i.e., (\ref{JohnPaidInCashAndHeIsProhibitedToPayInCash}.c).

\enumsentence{\label{JohnPaidInCashAndHeIsProhibitedToPayInCash}
\begin{minipage}[t]{400pt}\tt
\begin{enumerate}
    \renewcommand{\labelenumi}{\alph{enumi}.}
    \item {\tt\hspace{-2pt}[a :necessary, :hold; rdf:subject soa:epsj;
    rdf:predicate\\
    \mbox{}\hspace{4pt}soa:has-instrument; rdf:object soa:cash]}
    \vspace{3pt}
    \item\tt soa:epsj soa:has-instrument soa:cash.
    \vspace{3pt}
    \item\tt\hspace{-2pt}[a :false,:hold; rdf:subject [a soa:Pay; soa:has-agent soa:John;\\
    \mbox{}\hspace{4pt}soa:has-instrument soa:cash]; rdf:predicate rdf:type; rdf:object\\
    \mbox{}\hspace{4pt}:Permitted] is-violated-by [a :true,hold; rdf:subject soa:epsj;\\ \mbox{}\hspace{4pt}rdf:predicate rdf:type; rdf:object :Rexist].
\end{enumerate}
\end{minipage}
}

\noindent The example just discussed is available on the GitHub repository; the reader may easily verify that by changing the modality of {\tt soa:epsj} from {\tt Rexist} to {\tt Obligatory} a by re-executing the example, a {\it conflict} is inferred in place of the violation: John would be obliged to pay in cash while, at the same time, being prohibited to do so.

Therefore, regardless of the modality ({\tt Rexist} rather than {\tt Obligatory}), the fact that it is prohibited to pay in cash is incompatible with the fact that (in Sketty) it is necessary to do so. This incompatibility might be seen as further ``abnormality'' that we would like LegalTech applications to automatically find out and notify.

In order to do so, a new RDF property {\tt is-necessarily-violated-by} is added to the proposed computational ontology, together with the following SPARQL rule, which states that if it is necessary for the thematic role of a given eventuality to have a specific value, and there is another more specific eventuality that features that value on that specific thematic role and that is prohibited, then the fact that the latter is prohibited {\tt is-necessarily-violated-by} the fact that the thematic role is necessary in the former.

\enumsentence{\label{SPARQLisNecessarilyViolatedbyNotPermittedRexist}
\begin{minipage}[t]{550pt}\tt
[a :InferenceRule; :has-sparql-code """\\
\mbox{}\hspace{10pt}CONSTRUCT\{?rep :is-necessarily-violated-by ?ren\}\\
\mbox{}\hspace{10pt}WHERE\{?trn a :ThematicRole. ?en ?trn ?vn. ?ren a :necessary,:hold;\\
\mbox{}\hspace{20pt}rdf:subject ?en; rdf:predicate ?trn; rdf:object ?vn.\\
\mbox{}\hspace{20pt}?rep a :false,:hold; rdf:subject ?ep; rdf:predicate rdf:type;\\
\mbox{}\hspace{20pt}rdf:object :Permitted. ?en a ?c. ?ep a ?c. ?c a :Eventuality.\\
\mbox{}\hspace{20pt}NOT\hspace{-2pt} EXISTS\hspace{-1pt}\{\hspace{-1pt}?tr\hspace{-2pt} a\hspace{-2pt} :\hspace{-1pt}ThematicRole.\hspace{-4pt} ?en\hspace{-1pt} ?tr\hspace{-2pt} ?vn.\hspace{-3pt} NOT\hspace{-1pt} EXISTS\hspace{-1pt}\{?ep\hspace{-1pt} ?tr\hspace{-1pt} ?vp\}\}\\
\mbox{}\hspace{20pt}NOT\hspace{-2pt} EXISTS\hspace{-1pt}\{\hspace{-1pt}?tr\hspace{-2pt} a\hspace{-2pt} :\hspace{-1pt}ThematicRole.\hspace{-4pt} ?en\hspace{-1pt} ?tr\hspace{-2pt} ?vn.\hspace{-3pt} ?ep\hspace{-2pt} ?tr\hspace{-2pt} ?vp.\\
\mbox{}\hspace{20pt}FILTER(?vn!=?vp)\}\}"""]\hspace{-1pt}.
\end{minipage}
}
\vspace{6pt}

\noindent The SPARQL rule in (\ref{SPARQLisNecessarilyViolatedbyNotPermittedRexist}) adds the following RDF triples to  (\ref{JohnPaidInCashAndHeIsProhibitedToPayInCash}.c): the fact that John is prohibited to pay in cash 
{\tt is-necessarily-violated-by}
by the fact that it is necessary for the instrument of {\tt soa:epsj} to be cash.

\enumsentence{\label{violatedAndNecessarilyViolated}
\tt[a :false,:hold; rdf:subject [a soa:Pay; soa:has-agent soa:John;\\
    \mbox{}\hspace{4pt}soa:has-instrument soa:cash]; rdf:predicate rdf:type; rdf:object\\
    \mbox{}\hspace{4pt}:Permitted]\hspace{-1pt} is-necessarily-violated-by\hspace{-1pt} [a\hspace{-1pt} :necessary,hold;\hspace{-1pt} rdf:subject\\ \mbox{}\hspace{4pt}soa:epsj; rdf:predicate soa:has-instrument; rdf:object soa:cash].
}

\noindent A symmetric SPARQL rule might be added to infer that, whenever 
it is {\it not possible} for a certain thematic role to have a certain value but for a given eventuality this is {\it obligatory}, then the fact that the latter is obligatory {\tt is-necessarily-violated-by} the fact that it is not possible to have that value on that thematic role. Similarly, we might introduce a new RDF property {\tt is-nullified-by}
and a SPARQL rule to infer that, whenever 
it is {\it not possible} for a certain thematic role to have a certain value but for a given eventuality this is {\it permitted}, then the permission {\tt is-nullified-by} the fact that it is actually impossible to do what is permitted.

Developing an exhaustive list of RDF properties that explicitly represent abnormalities between deontic statements and/or really existing eventualities and/or contextual constraints is much beyond the scope of this paper.

Only a final one is worth commenting, before concluding this section. Suppose that the SPARQL rules shown in (\ref{payCashNotCardAndCardNotCashSPARQL}) and (\ref{inskettyitisnecessarytopaycash}) above hold in the state of affairs and that John declares the following:

\enumsentence{\label{JohnPaysByCardAtSkettyParkingMeter}
\hspace{2pt}John pays {\it by card} at the parking meter in Sketty.\\[5pt]
\mbox{}\hspace{2pt}\begin{minipage}[t]{550pt}\tt soa:epscj a soa:Pay,:Rexist; soa:has-agent soa:John;\\
soa:has-recipient soa:pmSketty; 
soa:has-instrument soa:card.
\end{minipage}
}
\vspace{2pt}

\noindent The SPARQL rule in (\ref{inskettyitisnecessarytopaycash}) infers that John pays {\it in cash} at the parking meter in Sketty (since it is {\it necessary} to do so), and then the rules in (\ref{payCashNotCardAndCardNotCashSPARQL}) infers that John did not pay neither in cash nor by card at that parking meter:

\enumsentence{\label{JohnPaysByCashButNotByCashNorByCardAtSkettyParkingMeter}\tt
\mbox{}soa:epscj soa:has-instrument soa:cash. [a :false,:hold;\\
\mbox{}\hspace{2pt}rdf:subject soa:epscj; rdf:predicate soa:has-instrument;\\ 
\mbox{}\hspace{2pt}rdf:object soa:cash]. [a :false,:hold; rdf:subject soa:epscj;\\ 
\mbox{}\hspace{2pt}rdf:predicate soa:has-instrument; rdf:object soa:cash].
}
\vspace{2pt}

\noindent The triples in (\ref{JohnPaysByCardAtSkettyParkingMeter}) and (\ref{JohnPaysByCashButNotByCashNorByCardAtSkettyParkingMeter}) are of course contradictory. However, the single SPARQL rule added so far to the proposed computational ontology to infer contradictions (shown above in (\ref{SPARQLcontradictionRexistNotRexist})) is unable to detect so because it only searches for eventualities that really exist and that, simultaneously, do not. 

We therefore add the following SPARQL rule that searches for eventualities having a specific value for one of their thematic roles and for which, simultaneously, the knowledge graph asserts that that thematic role does not have that value; the rule infers that the fact that the eventuality has that value on that thematic role {\tt is-in-contradiction-with} the fact that it does not have it.

\enumsentence{\label{SPARQLcontradictionRexistNotRexistThematicRole}
\begin{minipage}[t]{550pt}\tt
[a :InferenceRule; :has-sparql-code """\\
\mbox{}\hspace{13pt}CONSTRUCT\{\hspace{-1pt}[a :true,:hold;\\
\mbox{}\hspace{78pt}rdf:subject ?e; rdf:predicate ?tr; rdf:object ?tv]\\
\mbox{}\hspace{73pt}:is-in-contradiction-with ?r\}\\
\mbox{}\hspace{14pt}WHERE\{?e\hspace{-1pt} ?tr\hspace{-1pt} ?tv. ?tr a :ThematicRole.\hspace{-2pt} ?r\hspace{-1pt} a\hspace{-1pt} :false,\hspace{-1pt}:hold;\\ 
\mbox{}\hspace{50pt}rdf:subject\hspace{-1pt} ?e; rdf:predicate ?tr; rdf:object ?tv.\\
\mbox{}\hspace{51.5pt}NOT\hspace{-3pt} EXISTS\hspace{-1pt}\{\{?t :is-in-contradiction-with ?r\} UNION\\
\mbox{}\hspace{58pt}\{?r :is-in-contradiction-with ?t\} ?t a :true,:hold;\\
\mbox{}\hspace{60pt}rdf:subject\hspace{-2pt} ?e; rdf:predicate ?tr; rdf:object ?tv\}\hspace{-1pt}\}\hspace{-1pt}"\hspace{-1pt}"\hspace{-1pt}"\hspace{-1pt}]\hspace{-1pt}.
\end{minipage}
}
\vspace{2pt}

\noindent The SPARQL rule in (\ref{SPARQLcontradictionRexistNotRexistThematicRole}) properly infers that 
the triples in (\ref{JohnPaysByCardAtSkettyParkingMeter}) and (\ref{JohnPaysByCashButNotByCashNorByCardAtSkettyParkingMeter}) are contradictory. 

Suppose now that a LegalTech application based on the proposed computational ontology just notified us that the triples in (\ref{JohnPaysByCardAtSkettyParkingMeter}) and (\ref{JohnPaysByCashButNotByCashNorByCardAtSkettyParkingMeter}) are contradictory. What would we conclude? That either {\it John lied} when declaring (\ref{JohnPaysByCardAtSkettyParkingMeter}) or that {\it there is an error in the database}, namely that there is written there that in Sketty it is necessary to pay by cash but this is not true: indeed payments are also possible by card.

We want to conclude this section by arguing that, thanks to reification, it is in principle possible to replicate and automate these very same inferences. Although this requires much further work, it should be clear to the reader that it is possible to add further SPARQL rules that, after the property {\tt is-in-contradiction-with} is inferred from the triples in (\ref{JohnPaysByCardAtSkettyParkingMeter}) and (\ref{JohnPaysByCashButNotByCashNorByCardAtSkettyParkingMeter}), in turn infer the following triples:

\enumsentence{\label{JohnLiedOrErrorInDB}\tt
\mbox{}\hspace{-2pt}[a :Rexist;\\
\mbox{}\hspace{2pt}:or1\hspace{-2pt} [a\hspace{-2pt} :Lie,:Rexist;\hspace{-2pt} soa:has-agent soa:John;\hspace{-2pt} \mbox{soa:has-theme:\hspace{2pt}soa:epscj]\hspace{-1pt};}\\
\mbox{}\hspace{2pt}:or2\hspace{-2pt} [a\hspace{-2pt} :true,:hold;\\ \mbox{}\hspace{20pt}rdf:subject\hspace{-2pt} 
[a :necessary,:hold;
rdf:subject soa:epscj;\\
\mbox{}\hspace{76pt}rdf:predicate soa:has-instrument; rdf:object soa:cash];\\
\mbox{}\hspace{20pt}rdf:predicate rdf:type; rdf:object :Error]
}
\vspace{2pt}

\noindent The triples in (\ref{JohnLiedOrErrorInDB}) represent that John is lying about the eventuality that he declared in (\ref{JohnPaysByCardAtSkettyParkingMeter}) or the fact that it is necessary for the instrument of that eventuality to be cash is an error.

What has just been discussed is what we meant in the Introduction by ``explicit representation of fallacies, violations, mistakes, etc., (henceforth referred under the
general term “abnormalities”) fit to {\it reason} about them'', which has been also recently advocated in \cite{Steen-Benzmuller:24} as one of the main research directions for the future of AI. Reification and, therefore, RDF appear to provide a straightforward and effective way to achieve so.

\section{Future works: (1) sets and NL quantification and (2) temporal management}\label{FutureWorks}

\noindent Several future works has been already pointed out here and there all over the sections above. For instance, at the end of section \ref{BackgroundLegalTechOnRDF}, it has been mentioned the need of re-implementing the lightweight reasoner used in this paper in DLV2 or Vadalog, to both achieve efficiency and to represent defeasible norms; at the end of section \ref{ConjunctionDisjunctionImplicationOfDeonticStatements}, it has been pointed out that the interplay between deontic modalities and (non-material) implication is trickier than what it seems at first glance and so it deserve further investigations; at the end of the previous section it has been discussed how it would be possible to reason about abnormalities towards the identification of the {\it causes} that led to such abnormalities; etc.

All these gaps are due to the fact that this paper is, to the best of our knowledge, the first substantial attempt to merge into a unified framework contributions from three different research strands, which have been investigated almost independently in past literature: (1) LegalTech solutions compatible with RDF  (2) Natural Language Semantics via reification and (3) conflict-tolerant accounts in formal Deontic Logic.

Therefore, lot of further research is still needed to make the proposed computational ontology deployable within existing applications. In this section, we want to elaborate a bit more about two (additional) future works in particular, which we consider as our next steps: integrating in the proposed computational ontology (1) a proper representation of sets and natural language quantification and (2) temporal management.

\subsection{Sets and NL quantification}

\noindent Everywhere in this paper but in subsection \ref{ModellingNorms} all examples involved a {\it single} fictional individual, in most cases called ``John'': ``John leaves'', ``John is obliged to pay £3'', ``John is prohibited to pay in cash'', etc. Conversely, subsection \ref{ModellingNorms} discusses how to model {\it norms}, which do not usually apply to single individuals but rather to a contextually-relevant {\it set} of individuals, called the ``bearers'' of the norms.

In line with standard literature in legal theory, in subsection \ref{ModellingNorms} norms have been represented as SPARQL if-then rules, in which the ``if'' part, i.e., the {\tt WHERE} clause, specifies the conditions that the individuals of the state of affairs must satisfy in order to belong to the set of bearers. The triples specified in the {\tt CONSTRUCT} clause, instead, are those that will be asserted on every individual bearer identified by the {\tt WHERE} clause.

However, as explained in subsection \ref{ModellingNorms}, in order to identify conflicts among norms represented as if-then rules, it would be necessary to run {\it simulations} with synthetic datasets that include fictional individuals satisfying the {\tt WHERE} clauses of the SPARQL rules, in all possible ways. This is not ideal: it would be preferable to set up an inference schema that treats sets as if they were single individuals.

This problem is not new in the literature, and it is just a specific case of the general problem of uniformly representing and relating sets and members of sets in formal semantics. The solution adopted in \cite{Gordon-Hobbs:17}, imported from \cite{McCarthy:77}, is to introduce special individuals called {\it typical elements}, which are reifications of sets. Typical elements are syntactically/formally encoded as single specific individuals like John, but it is stipulated that every assertion on a typical element entails corresponding assertions on each member of the set that the typical element reifies.

In our future works, we intend to model norms not as if-then rules, but rather as sets of triples asserted on the typical elements reifying the norm's set of bearers and any other set involved in the norm' meaning. Thus, it will be possible to infer conflicts among norms without running simulations with synthetic datasets created on purpose. On the other hand, for every known individual belonging to a reified set of bearers, further SPARQL rules will check whether or not the individual complies with the obligations asserted on the typical element reifying that set.

Although the advocated solution sounds reasonable, its formalization in RDF and SPARQL is not so trivial in the general case, but it rather requires lot of care and further investigations.

As discussed in \cite{Gordon-Hobbs:17}, typical elements are intimately related with the distinction between abstract eventualities and their instantiations as well as with the distinction between types and tokens\footnote{See \url{https://plato.stanford.edu/entries/types-tokens}}, which has been widely investigated in philosophy, linguistics, and formal semantics. Suppose for instance that John is obliged to pay £3 in cash and that he has six coins of £1 each in his pocket. Then, there are 6!/(3!*(6-3)!)=20 possible subsets of the six coins that John can use to pay £3. From the point of view of the obligation, which denotes an abstract eventuality, each of the 20 combinations is fine. Once John selects one of them to pay, i.e., an eventuality is asserted on the {\tt Rexist} modality to comply with the obligation, the set of coins in his pocket changes accordingly, and this in turn affects the set of possible combinations he can make his future payments with. In light of this, it is clear that while the {\tt Rexist} modality holds for typical elements reifying sets of {\it tokens}, deontic modalities hold for typical elements reifying sets of {\it types}, which may be represented as families of sets of tokens. It is then necessary to define a specific {\it arithmetic} on types and tokens and implement it in terms of SPARQL rules, in order to derive the right inferences.

Further complexities arise when thematic roles involve generalized quantifiers\footnote{See \url{https://plato.stanford.edu/entries/generalized-quantifiers}}. For instance, how could we represent and reason with the meaning of (\ref{conflictingobligationsGQs}.a-b), in order to infer that they conflict of one another?

\enumsentence{\label{conflictingobligationsGQs}
\begin{enumerate}
    \renewcommand{\labelenumi}{\alph{enumi}.}
    \item John is obliged to pay at least £10.
    \vspace{5pt}
    \item John is prohibited to pay more than £5.
\end{enumerate}
}

\noindent Generalized quantifiers may also engender ambiguities when two or more of them occur in the sentence, with each ambiguous interpretation featuring different compliance checking inferences. Consider:

\enumsentence{\label{ambiguousObligationsGQs}
\begin{enumerate}
    \renewcommand{\labelenumi}{\alph{enumi}.}
    \item The men are obliged to wear a red t-shirt.
    \vspace{5pt}
    \item The men are obliged to lift a table.
    \vspace{5pt}
    \item The men are obliged to pay a £3000 sanction.
\end{enumerate}
}

\noindent The preferred interpretation of (\ref{ambiguousObligationsGQs}.a) is the {\it distributive} reading in which every man is obliged to wear a different red t-shirt. Conversely, the preferred interpretation of (\ref{ambiguousObligationsGQs}.b) is the {\it collective} reading in which the men, altogether, must lift a single table. Finally, the preferred interpretation of (\ref{ambiguousObligationsGQs}.c) is the {\it cumulative} reading in which the men must share a £3000 sanction among them, i.e., in which each of them will pay an amount of money lower than £3000 but such that the sum of all the amounts paid is equal to £3000. From these 
examples, especially (\ref{ambiguousObligationsGQs}.a) and (\ref{ambiguousObligationsGQs}.c), it should be clear that the SPARQL rules to check compliance seen above must be evolved in order to infer whether {\it sets} of really existing eventualities comply or not with the obligations.

To conclude, in order to evolve the proposed computational ontology as explained in this subsection, it is necessary to incorporate therein a solid account of sets and natural language quantification. In our future works, we plan in particular to incorporate in the proposed computational ontology insights from %\cite{Robaldo:10},
\cite{Robaldo:11} and \cite{Robaldo-etal:14}, in which sets are also reified into terms.

\subsection{Temporal management}

\noindent It is well-known that temporal information is crucial for automated compliance checking and legal reasoning as a whole. Not only because it contributes to identify the contextually relevant sets of bearers that the deontic statements are about, but also because every deontic statement is also, explicitly or implicitly, associated with a temporal validity as well as with temporal constraints within which the obligations must be complied with.

This is the case for deontic statements coming from existing legislation, i.e., the ones our research activity as a whole aims at representing and processing. Every act from legislation is associated with a date in which it enters into force. Then, the norms in the act might be subsequently amended and the amendments will be also associated with a date from which they will enter into force. Still, the previous version of the norm will continue to apply for facts that took place in the temporal interval between the two dates.

In light of this, it is clear that the proposed computational ontology cannot be deployed yet within LegalTech applications checking compliance with respect to existing legislation until mechanisms for temporal management will be incorporated therein. While all examples discussed above involve eventualities that take place at the instant ``now'', further SPARQL rules ought to be added to properly process deontic statements that span over specific temporal {\it intervals}. As a simple example, consider the following sentences:

\enumsentence{\label{obligationsTemporalManagement}
\begin{enumerate}
    \renewcommand{\labelenumi}{\alph{enumi}.}
    \item It is prohibited to enter the park from 3pm until 5pm.
    \vspace{5pt}
    \item John was in the park from 4pm until 6pm.
\end{enumerate}
}

\noindent From (\ref{obligationsTemporalManagement}.a-b), it is possible to infer that John violated the obligation in (\ref{obligationsTemporalManagement}.a) {\it but only from 4pm until 5pm}. In order to implement this inference in the proposed computational ontology, it is necessary to add SPARQL rules able to infer that the two intervals mentioned in (\ref{obligationsTemporalManagement}.a-b) {\it overlap} and then able to create a {\it new} interval out of their temporal boundaries. A solution to achieve so is to add SPARQL rules that encode the well-known Allen's relation \cite{Allen:84} and that create, in their {\tt CONSTRUCT} clause, the new relevant intervals in which the deontic statements are violated, complied with, conflicting of one another other, etc.

Similar considerations hold about the intervals between the instant in which an obligation starts being in force and the instants in which the really existing eventualities complying with that obligation take place. For instance, if John receives a fine because he did not pay for his parking, he will have a specific amount of time, e.g., one month, to pay the fine, otherwise he will receive another fine. In the literature, the payment of a fine is usually represented as a further obligation (see \cite{Governatori-Rotolo:19}): if John violates his obligation of paying the parking, he is {\it obliged} to pay a fine within a certain amount of time. The obligation of paying the fine will replace the obligation of paying the parking. If John will pay the fine within the established temporal interval, the obligation of paying it will be compensated by the payment. Otherwise, another obligation of paying an increased fine will replace the previous obligation.

Based on these considerations, 
\cite{Governatori:15} presents an approach in which obligations are subcategorized according to the time in which the eventualities complying with them really exist and to how the temporal interval in which obligations are in force changes after they are complied with.

In our future works, we plan to incorporate the same subcategorization in the proposed computational ontology while representing time via the Time Ontology\footnote{\url{https://www.w3.org/TR/owl-time}}. The vocabulary of the Time Ontology include RDF resources to represent instants, intervals, and Allen's relations among intervals. We will introduce specific thematic roles that associate eventualities with instants or intervals as well as SPARQL rules to carry out the desired inferences on temporal data. The Time Ontology does not natively enable these inferences because it is based on OWL, but OWL does not provide constructs for temporal managements, as already explained in section \ref{BackgroundLegalTechOnRDF} above.

\section{Conclusions}\label{Conclusions}

\noindent This paper proposed a conflict-tolerant deontic logic to handle irresolvable conflicts suitable for the Semantic Web, because it has been axiomatized in RDFs and SPARQL. We believe the proposed computational ontology to be novel for two main reasons.

First, to the best of our knowledge, it is the first substantial attempt to propose a vocabulary of RDF resources and associated inference rules to deal with {\it all} deontic modalities identified in the Deontic Traditional Scheme, i.e., obligations, permissions, optionality, and their negations, as well as with various types of irresolvable conflicts, violations, and the interplay between deontic modalities and contextual constraints.

Other approaches in this direction have been proposed in past literature in LegalTech, but (1) their coverage is much smaller than the one of the proposed computational ontology and (2) they are either based on formats that are not W3C standards, e.g., SWRL, or they are based on OWL, which has a limited expressivity, unable to represent the inferences discussed here nor other inferences needed in legal reasoning. 

Although the formalization of norms could appear to be a niche research topic for some, we believe it to be crucial for enabling further developments in many disciplines as well as connected industrial use cases. Well-functioning judicial systems are a crucial determinant for economic performance. Since legislation is at the basis of
and regulates our everyday life and societies, many categories of Big Data (e.g.,
medical records in eHealth, financial data, etc.) must comply with and are thus highly
dependent on specific norms. Matching and annotating Big Data with legislative
information will produce even more and richer Big Data. In light of this, although a lot of further work still needs to be done for making the proposed computational ontology deployable within industrial LegalTech applications, as explained all over the paper and in particular in the previous section, we deem our work to be a necessary step in the general research and innovation in Big Data processing that, sooner or later, someone had to make.

The second reason why we believe
the proposed computational ontology to be novel is that, in our view, it also represents a substantial step forward in the {\it theoretical} research in (conflict-tolerant) deontic logics. 

State-of-the-art approaches in formal Deontic Logic feature two main limits, which significantly hinder their investigation beyond theoretical boundaries. First, {\it most deontic logics proposed in the literature are based on propositional logic}, which is not expressive enough to distinguish terms from predicates, eventualities from their thematic roles, etc. Secondly, {\it most deontic logics proposed in the literature focuses on formalizing obligatoriness} while they kind of tacitly assume that the other deontic modalities can be reconciled with obligatoriness through the inferences of the Deontic Traditional Scheme. On the contrary, as already stated above, this paper formalizes {\it all} deontic modalities from the Deontic Traditional Scheme altogether. None of them is considered as ``more important'' than the others; however, note that if we really want to identify one of them as the ``main one'', this should be permission, not obligatoriness, at least from the point of view of inferring conflicts among deontic statements: the SPARQL rule to infer conflicts, shown above in (\ref{SPARQLisinconflictwith}), looks for pairs of eventualities such that one of them is permitted, the other one is not, and the former is more specific than the latter. While an obligation entails a permission, a permission does {\it not} entail an obligation but rather a not-obligation, from which it is impossible to infer conflicts with other deontic statements. Therefore, the most general rule to infer all categories of conflicts identified in the work of Hans Kelsen, must be based on the notion of permission rather than obligation.

The conflict-tolerant deontic logics reviewed in \cite{Goble:13}, which we consider as the most complete survey on the topic, are exemplifying of these limitations. Furthermore, we do not agree with the definition of conflicts used in \cite{Goble:13}, i.e., situations in which an agent ought to do a number of
things, each of which is possible for the agent, but it is impossible for the agent to do them all. In line with Kelsen's work, in this paper a conflict of deontic statements is instead defined as a situation in which two or more deontic statements hold in the context but complying with one of them entails violating (or not permitting) another one. In other words, our definition is grounded on the notion of {\it violation} rather than on the notions of possibility and necessity. The notion of violation has been neglected in past literature in deontic logic while this paper took it as the starting point to design the SPARQL rule in (\ref{SPARQLisinconflictwith}).

Conversely, in our view the notions of possibility and necessity mostly concern {\it the thematic roles} of the eventualities, in that contextual constraints usually restrict {\it possible values} on these thematic roles, due to the physical limits occurring in the state of affairs (e.g., the fact that some parking meters only accept cash). This level of details is not addressed in state-of-the-art deontic logics because, as observed earlier, they are based on propositional logic and so they are not fine-grained enough to represent and reason with thematic roles and their possible values.

The main architectural choice that we implemented in the proposed computational ontology is the distinction between {\it two levels of representation}, which we called ``the level of the eventualities'' and ``the level of the statements'', in line with the philosophical and psycholinguistic studies of Jerry R. Hobbs. {\it Deontic modalities belong to the level of the eventualities}, also in line with past studies in deontic logic claiming that a proper truth-conditional logic of norms is impossible in that norms do not carry truth
values (cf. J{\o}rgensen's dilemma \cite{Jorgensen:37}). On top of the level of the eventualities, the level of the statements allows for the modelling of standard boolean connectives (\mbox{$\neg$\hspace{1pt},} \mbox{$\wedge$\hspace{1pt}}, and $\vee$\hspace{1pt}) as well as of the operators for necessity and possibility (\hspace{1pt}$\Box$ and $\Diamond$). This 2-level architecture is alternative to standard approaches in deontic logics, in which deontic operators are usually defined on top of standard propositional modal logic, which have been historically proposed before. In our approach, it is the other way round.

In our formalization, the level of the eventualities as well as its connections with the level of the statements have been implemented via {\it reification}, a well-known mechanism, massively used in Hobbs's, to associate abstract entities with first-order individuals. RDF also provides constructs to reify the triples, which we used to ascribe them their truth values, thus reconciling negation, disjunction, necessity, and possibility with the Open World Assumption. To the best of our knowledge, this paper also represents the first attempt to implement these operators from classical modal logic in RDF, while retaining the Open World Assumption.

Finally, it is precisely thanks to reification that the proposed computational ontology manages to achieve another crucial advantage, which is perhaps its {\it main} advantage. After reifying the statements, it is possible to introduce special RDF properties to notify contradictions, conflicts, violations, and other {\it abnormalities}, so that it is in turn possible to {\it reason} about them. For example, as it has been discussed at the end of section \ref{DeonticModalitiesAndContextualConstraints} 
above that, by explicitly encoding that the fact that John paid by card in Sketty is in contradiction with the fact that in Sketty only cash is accepted, it is possible to infer that either John is lying or that there is an error in the database. This is what humans would likely infer when standard reasoners would report them such a contradiction. 

Nevertheless, standard automated reasoners, e.g., HermiT, typically {\it stop} their execution whenever a contradiction is inferred. In other words, these reasoners are capable to only work with {\it consistent} knowledge. As argued in \cite{Steen-Benzmuller:24}, this is no longer acceptable in modern Artificial Intelligence systems, which should be instead capable of processing abnormal statements, including contradictory ones. Same considerations hold for conflicts or violations: although they are consistent formulae, they should be flagged as abnormal, i.e., they should be distinguished from other consistent formulae, in order to enable further inferences from them. Therefore, conflict-tolerant deontic logics that do not distinguish conflicts from other consistent formulae appear to be limited also from this point of view.

\bibliographystyle{fullname}
\bibliography{fullname-doc}

\end{document}